\documentclass{article}

\usepackage{times}
\usepackage{graphicx}
\usepackage{subfigure} 

\usepackage[round]{natbib}
\usepackage{algorithm}
\usepackage{algorithmic}

\usepackage{amsmath}
\usepackage{amssymb}
\usepackage{authblk}
\usepackage{lscape}
\usepackage[toc,page]{appendix}

\usepackage{xcolor}

\newcommand{\vect}[1]{\boldsymbol{\mathbf{#1}}}

\usepackage[hidelinks]{hyperref}

\title{A Hippocampus Model for Online \\ One-Shot Storage of Pattern Sequences}

\author{Jan Melchior}
\author{Mehdi Bayati}
\author{Amir Azizi}
\author{\hspace{5cm}Sen Cheng} 
\author{Laurenz Wiskott}
\affil{\normalsize Institut f\"ur Neuroinformatik, Ruhr Universit\"at Bochum, 44780 Bochum, Germany
\texttt{<forename>.<surname>@ruhr-uni-bochum.de} }

\begin{document}
\date{}
\maketitle

\begin{abstract}  
We present a computational model based on the CRISP theory (Content Representation, Intrinsic Sequences, and Pattern completion) of the hippocampus that allows to continuously store pattern sequences online in a one-shot fashion.
Rather than storing a sequence in CA3, CA3 provides a pre-trained sequence that is hetero-associated with the input sequence, which allows the system to perform one-shot learning.
Plasticity on a short time scale therefore only happens in the incoming and outgoing connections of  CA3.
Stored sequences can later be recalled from a single cue pattern.
We identify the pattern separation performed by subregion DG to be necessary for storing sequences that contain correlated patterns.
A design principle of the model is that we use a single learning rule named \textit{Hebbian-descent} to train all parts of the system. \textit{Hebbian-descent} has an inherent forgetting mechanism that allows the system to continuously memorize new patterns while forgetting early stored ones.
The model shows a plausible behavior when noisy and new patterns are presented and has a rather high capacity of about 40\% in terms of the number of neurons in CA3.
One notable property of our model is that it is capable of `boot-strapping' (improving) itself without external input in a process we refer to as `dreaming'.
Besides artificially generated input sequences we also show that the model works with sequences of encoded handwritten digits or natural images.
To our knowledge this is the first model of the hippocampus that allows to store correlated pattern sequences online in a one-shot fashion without a consolidation process, which can instantaneously be recalled later.
\end{abstract} 

\section{Introduction}\label{sec:introduction}

Since the discovery of place cells in the rodent hippocampal formation by~\citet{o1971hippocampus}, honored with the Nobel prize in 2014, the hippocampus has become one of the most extensively studied regions in the brain.
Like in no other brain region the response of neurons in the hippocampal formation of rodents show a direct correlation to the temporo-spatial location of the rodent in the environment, which makes them an excellent object of study for how rodents orient and navigate~\citep{o1978hippocampus}.
On the other hand the hippocampus is also heavily involved in forming new memories especially those of episodic nature~\citep{tulving1972episodic}, making it an excellent object of study for memory formation in the human brain~\citep{bailey2013medial, swallow2011changes, tulving1998episodic}.
A prominent example is patient H.M. (Henry Gustav Molaison), whose hippocampus was almost completely removed in an attempt to cure his epileptic seizures as a consequence of which he could not form new episodic memories anymore~\citep{scoville1957loss, vargha1997differential}. 

The anatomical structure of the hippocampal formation can roughly be divided into the three major subregions Dentate Gyrus (DG), Cornu Ammonis area 1 (CA1), and Cornu Ammonis area 3 (CA3), which are connected in a feed-forward manner~\citep{amaral1990chapter}. 
The hippocampus receives input from and sends output back to the Entorhinal Cortex (EC), which itself receives input and sends output back to various brain regions in the cortex.
DG consists of a large number of sparsely active granule cells~\citep{leutgeb2007pattern}, whereas CA1 and CA3 consist of pyramidal cells where the latter is also famous for its recurrent collaterals~\citep{ishizuka1990organization, li1994hippocampal, rolls2007attractor}.
For a detailed introduction to the anatomical structure and physiological properties of the hippocampal formation see~\citep{andersen2007hippocampus}.
Among the aforementioned subregions, CA3 has been studied most intensively and is suggested, due to its recurrent collaterals, to function as an auto-associative memory that is capable of retrieving memories based on incomplete or corrupted cues~\citep{mcnaughton1987hippocampal, marr1991simple, treves1994computational, o1994hippocampal, guzman2016synaptic}.
This has become known as the standard framework~\citep{nadel1997memory} over the last decades, claiming that plasticity in CA3 is crucial.
Some experiments have provided evidence that CA3 is required for successful cue-based retrieval of memories~\citep{gold2005role}.
However, experiments have shown that the hippocampus is involved in learning of patterns in temporal order referred to as episodic memories~\citep{tulving1972episodic, cheng2013crisp}, which the standard framework does not naturally account for.
The recently proposed CRISP (Content Representation, Intrinsic Sequences, and Pattern completion) theory~\citep{cheng2013crisp} explicitly accounts for the temporal aspects of episodic memories in the hippocampus.
In this theory, episodic memories have been suggested to be represented by neuronal sequences in EC, which are hetero-associated with temporal intrinsic sequences in CA3. 
Later on, the network can retrieve entire sequences based on a single cue pattern.
Notice that the idea of associating an input sequence with an intrinsic sequence has previously been proposed by~\citep{Lisman-1999, verduzco2012model}
CRISP, like the standard framework can be mapped to the anatomical structure of the hippocampus and it has lately been shown to be capable of successfully storing and retrieving sequences of memories~\citep{Bayati2018}.
It also provides an explanation for the phenomenon of pre-play/re-play observed in the hippocampal formation, where stored sequences are recalled without external input~\citep{AziziWiskottCheng2013b}. 
In contrast to the standard framework, however, CRISP does not claim that plasticity in CA3 is required.

Undeniably, the brain must be able to instantaneously store new memories. 
Surprisingly, this online one-shot learning aspect (\emph{i.e.} storing one pattern at a time through a single update) is ignored in most computational studies regarding the hippocampus and off\-line learning (\emph{i.e.} storing all patterns at the same time) is used instead. 
This is probably due to the fact that online storage of patterns in artificial neural networks is tremendously more difficult than off\-line storage, as online storage has to avoid catastrophic forgetting~\citep{McCloskey1989} and at the same time allow to integrate new information while forgetting old information.
Online storage of static patterns in auto associative memories~\citep[e.g.][]{miyata2012properties} and off\-line storage of memory sequences in the hippocampus~\citep[e.g.][]{de2018heteroassociative} have been studied in separation.
We are not aware of any study that considers the online learning aspect and the storage of sequences in the hippocampal formation at the same time.

In this work we explicitly address the two aspects by proposing a simple hippocampal model for one-shot storage of pattern sequences that is based on the CRISP theory. 
It is related to the work by~\citet{Bayati2018}, but that study analyzed off\-line storage and retrieval of memory sequences.
Furthermore, we focus on the analysis of the hippocampal subregions EC, DG, and CA3 instead of EC, CA3, and CA1.
While both studies, assign the role of intrinsic sequence generation to CA3, we additionally assign the role of pattern separation to DG, which is consistent with the standard framework but differs from the role of `context reset' CRISP assigns to DG. 
However, `context reset' can be understood as a separation of sequences, so that the roles the two frameworks assign to DG are rather similar.
Subregion CA1 is not yet included in the model and postponed to future work.

As the primary objective is to understand what is really necessary for successful online storage of pattern sequences, we use rate-based neurons and postpone the more realistic spiking networks to future work.
Plasticity is implemented using a biologically plausible learning rule named Hebbian-decent~\citep{Melchior2019a}. 
It has been shown that this learning rule is superior to Hebb's rule and the covariance rule as it is not only capable of online learning but also provably stable, can deal much better with correlated input patterns, it has an inherent forgetting mechanism, and profits from seeing patterns several times.
We show that using this learning rule the model is indeed capable of one-shot storage of pattern sequences that can later be recalled from a single cue. 
Furthermore, the model shows a reasonable rate of forgetting allowing it to continuously store new patterns without interference.
We show decorrelation in DG is necessary for hetero-associating input sequences of correlated patterns and can be achieved in a very simple way.
We show that hetero-association with intrinsic sequences enables the model to successfully perform one-shot storage of pattern sequences.
Our model has the notable property to `bootstrap' (improve) itself based on a `dreaming' process that does not require external input.
Finally, we show that the required average activity in CA3 increases with decreasing model size, so that only large models function properly when using the average activity recorded from rat hippocampus.

\section{Model Architecture}\label{sec:model_architecture}

\begin{figure}[htbp!]
\begin{center}
\centerline{\includegraphics[scale=0.33, trim=0 0 50 0, clip]{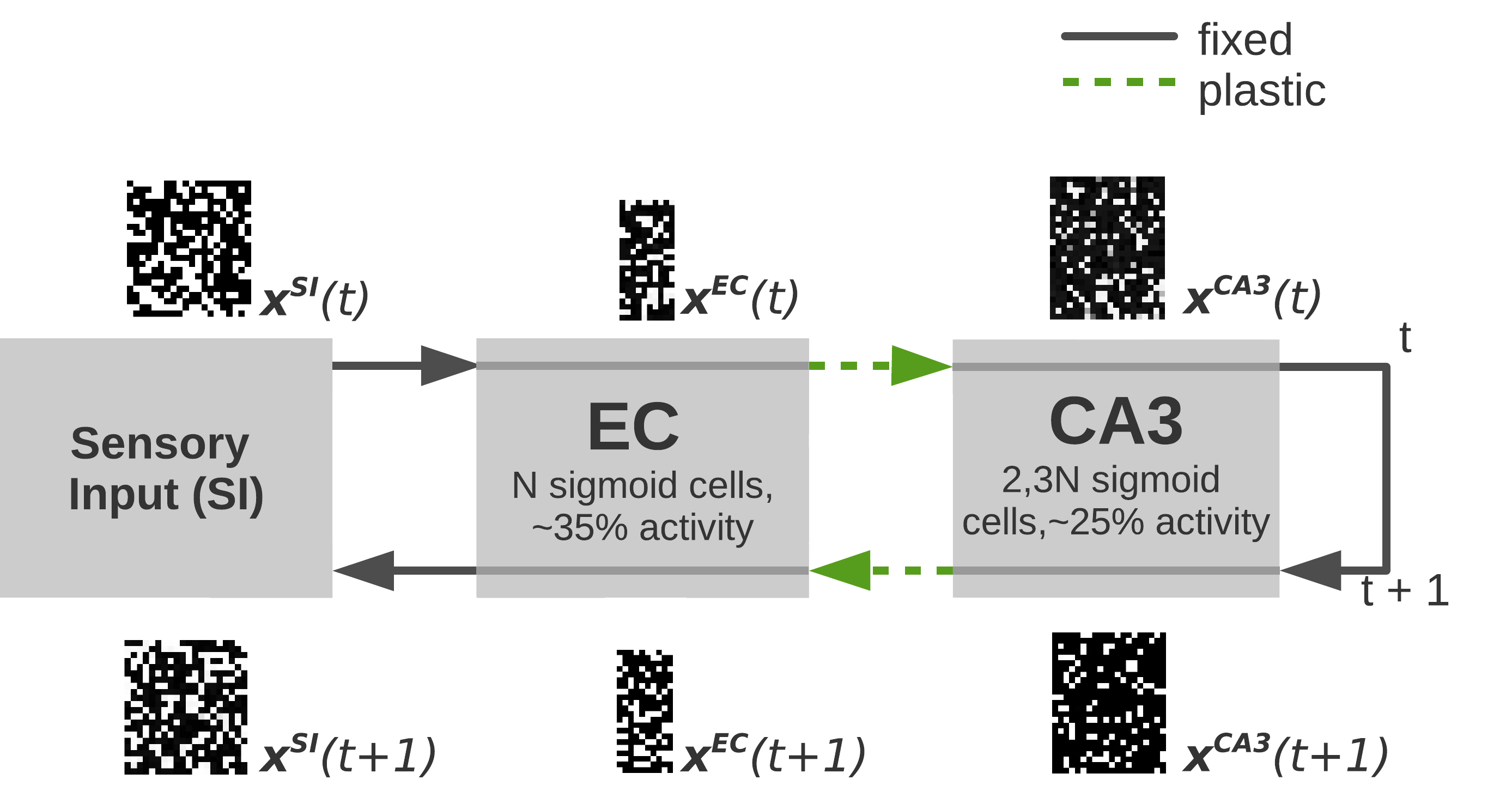}}
\caption{Illustration of \emph{Model-A}, which does not include DG. Here $N$ denotes the size of the model in terms of the number of neurons in EC.
Solid gray arrows denote non-plastic pathways while dashed green arrows denote plastic pathways. 
For illustration purpose, example patterns for the corresponding subregions are shown.
}
\label{fig:model_EC_CA3}
\end{center}
\end{figure}

The model of the hippocampus we consider in this work includes in its basic form only the hippocampal subregion CA3 and EC. 
We refer to this model as \emph{Model-A} illustrated in Figure~\ref{fig:model_EC_CA3}. 
This is our minimal model to store a sequence of input patterns and retrieve them in correct order given only a single cue pattern. 
In general we use dashed green arrows to represent plastic pathways and solid gray arrows to represent pathways that are fixed, \emph{i.e} not plastic on the fast time scale of learning we consider in this work. 
Plasticity in \emph{Model-A} therefore happens in the pathways EC~$\rightarrow$~CA3 and CA3~$\rightarrow$~EC, where the former pathway hetero-associates a sequence of binary patterns $\vect{x}^{EC}(1), \vect{x}^{EC}(2), \vect{x}^{EC}(3), \cdots$ with a sequence of intrinsic binary patterns $\vect{x}^{CA3}(1), \vect{x}^{CA3}(2), \vect{x}^{CA3}(3), \cdots$ one pattern-pair at time $t$ and the latter pathway performs the reverse hetero-association.
The patterns in EC are either generated artificially for analysis purpose or, more realistically, sensory input is transformed via the non-plastic but pre-trained pathway SI~$\rightarrow$~EC to binary patterns in EC, as shown in Figure~\ref{fig:model_EC_CA3}.
The non-plastic recurrent connections in CA3 are pre-trained to provide an intrinsic sequence of random patterns, which is described in more detail along with data generation, pre-training, and pattern storage/retrieval in the following sections.

\begin{figure}[htbp!]
\begin{center}
\centerline{\includegraphics[scale=0.33]{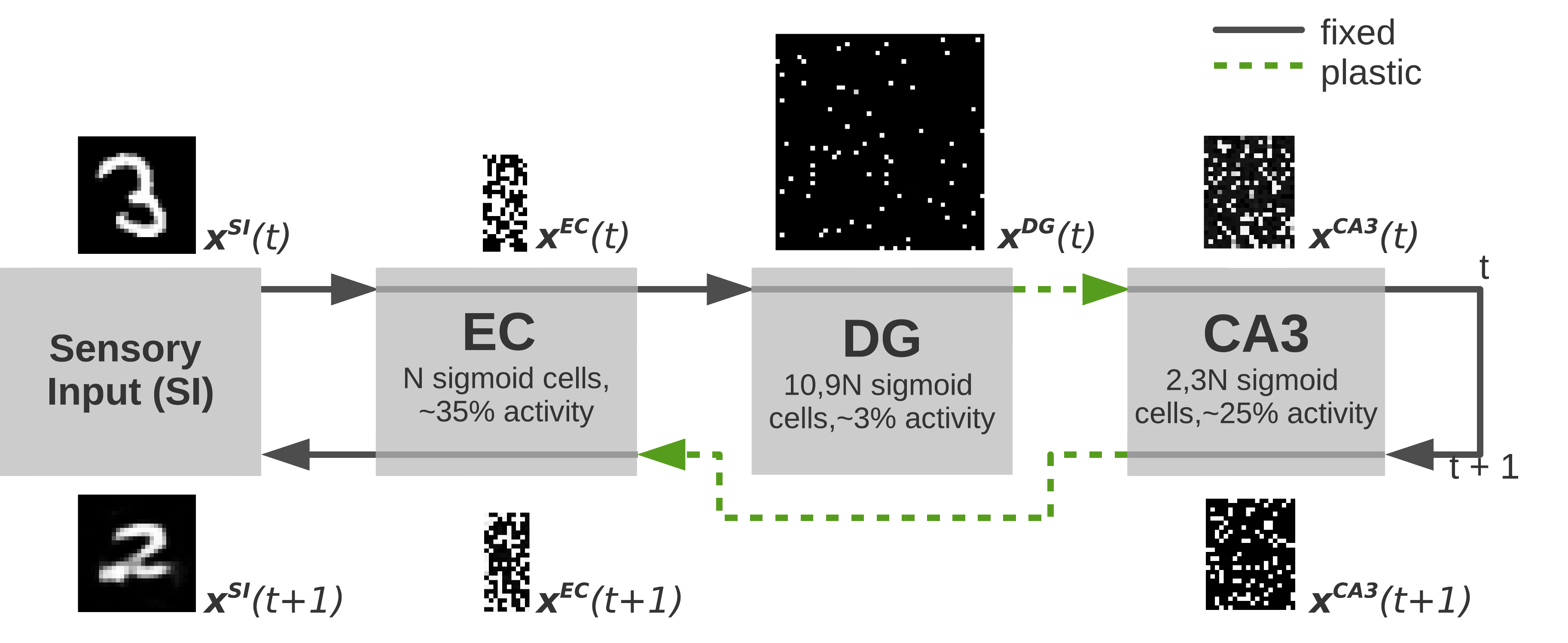}}
\caption{Illustration of \emph{Model-B}, which is the same as \emph{Model-A} except that it includes DG to transform pattern $\vect{x}^{EC}$ to a sparse representation $\vect{x}^{DG}$ via non-plastic pathway EC~$\rightarrow$~DG before being propagated to CA3 via plastic pathway DG~$\rightarrow$~CA3.
}
\label{fig:model_EC_DG_CA3}
\end{center}
\end{figure}

The basic model can be complemented by DG, which we then refer to as \emph{Model-B} shown in Figure~\ref{fig:model_EC_DG_CA3}.
According to the standard framework, DG is believed to perform pattern separation that allows to associate very similar patterns in EC with different patterns in CA3. 
As shown in the experiments this can be achieved by a generic and pre-trained DG that is independent of the input statistics. 
Notice, that pathway EC~$\rightarrow$~CA3 and EC~$\rightarrow$~DG~$\rightarrow$~CA3 do both exist in the hippocampal formation but we use the two models to analyze their role separately.

\subsection{Storage and Retrieval}

\begin{figure}[htbp!]
\begin{center}
\subfigure[Storage]{
\includegraphics[scale=0.27, trim=19 13 0 10, clip]{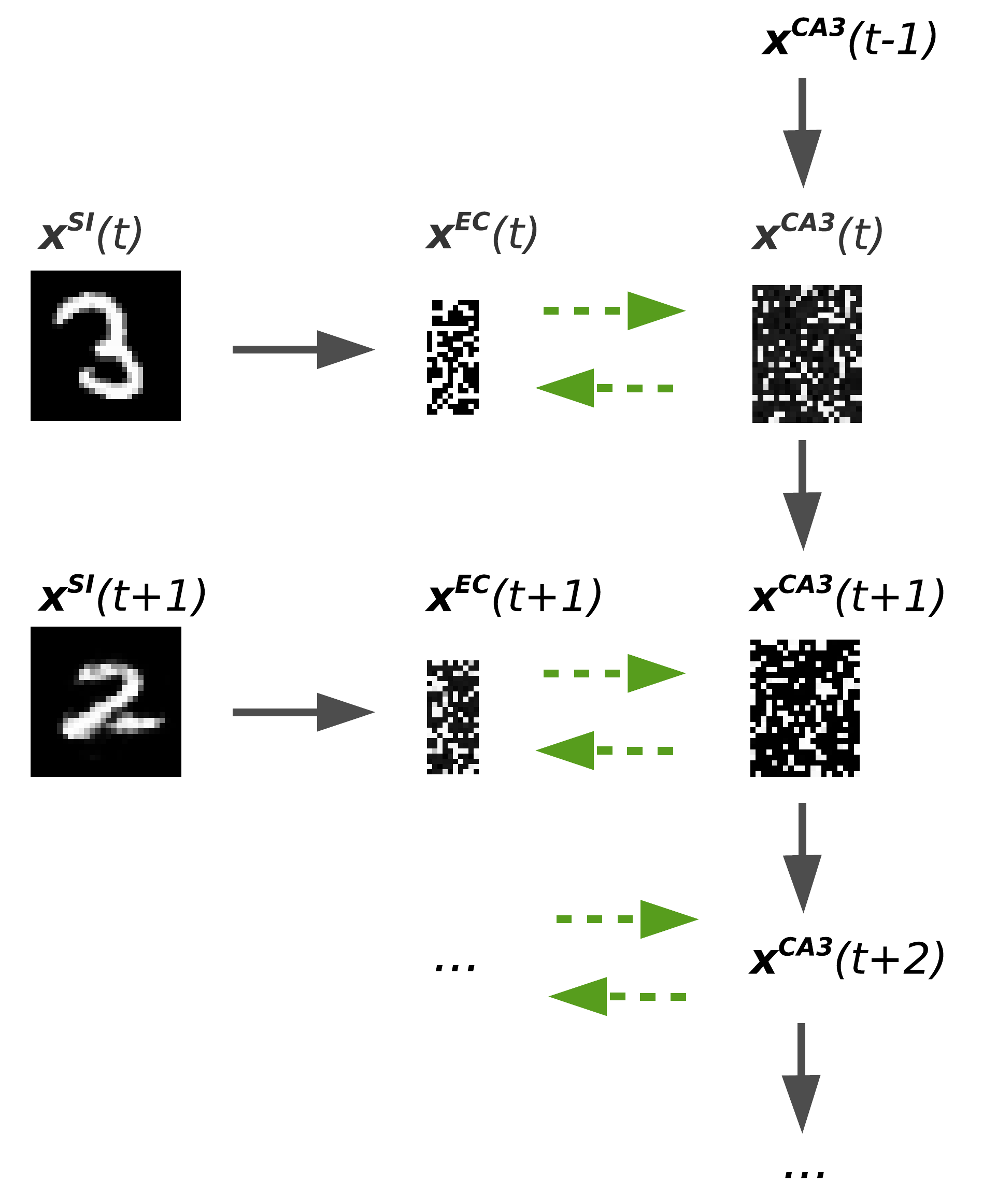}\label{fig:hippo_storage}}
\hfill
\subfigure[Retrieval]{
\includegraphics[scale=0.27, trim=0 12 20 10, clip]{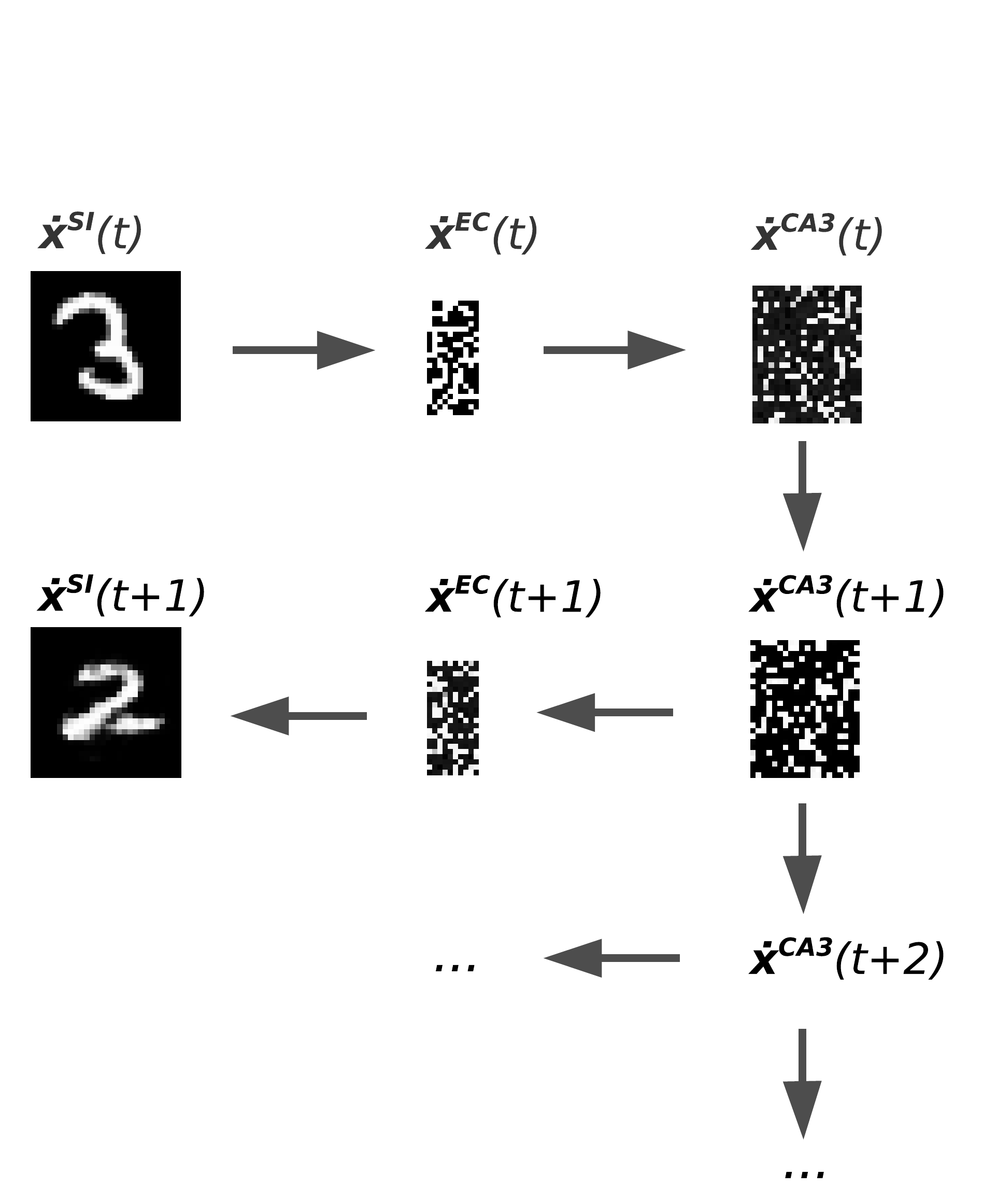}\label{fig:hippo_retrieval}}
\caption{Illustration of (a) storage and (b) retrieval of a pattern sequence in \emph{Model-A}, which works similarly for \emph{Model-B}. 
Solid gray arrows denote forward propagation of activities while dashed green arrows denote hetero-association in the corresponding pathway. 
Notice that the very first pattern $\vect{x}^{CA3}(t-1)$ is picked randomly from the intrinsic sequence and that $\dot{\vect{x}}$ denotes a retrieved potential corrupted pattern.
}
\label{fig:hippo_storage_and_retrieval}
\end{center}
\end{figure}

In this work we consider storage and retrieval as two separate modes that are illustrated in Figure~\ref{fig:hippo_storage_and_retrieval}. 
During storage the current input pattern $\vect{x}^{SI}(t)$ is transformed via pathway SI~$\rightarrow$~EC to pattern $\vect{x}^{EC}(t)$. 
At the same time the intrinsic dynamics in CA3 generates an intrinsic pattern $\vect{x}^{CA3}(t)$ from its predecessor $\vect{x}^{CA3}(t-1)$. 
Notice that the CA3 dynamics is cyclic such that each pattern has a unique predecessor and that the very first pattern $\vect{x}^{CA3}(t-1)$ is picked randomly from the intrinsic sequence.
Pattern $\vect{x}^{EC}(t)$ and $\vect{x}^{CA3}(t)$ are then hetero-associated in both directions through pathways EC~$\rightarrow$~CA3 and CA3~$\rightarrow$~EC. 
In the next time step  $\vect{x}^{CA3}(t)$ triggers the next intrinsic pattern $\vect{x}^{CA3}(t+1)$, which will be hetero-associated in both directions with pattern $\vect{x}^{EC}(t+1)$. This process is repeated for the next pattern pair and so on. 
We assume that the hippocampal theta rhythm suppresses the input from pathway EC~$\rightarrow$~CA3 such that the CA3 states only depend on the intrinsic dynamics during storage~\citep{buzsaki2002theta,hasselmo2002proposed,colgin2016rhythms}.

As illustrated in Figure~\ref{fig:hippo_retrieval}, retrieval is initiated by providing a potentially corrupted cue pattern $\dot{\vect{x}}^{SI}(t)$, which is propagated via pathway SI~$\rightarrow$~EC~$\rightarrow$~CA3 while the intrinsic dynamics in CA3 is suppressed~\citep{hasselmo2002proposed,colgin2016rhythms},
so that a pattern $\dot{\vect{x}}^{CA3}(t)$ is triggered that should ideally be equal to $\vect{x}^{CA3}(t)$.
After that, the pathway EC~$\rightarrow$~CA3 is suppressed, so that the CA3 dynamics re-plays the intrinsic sequence starting from pattern $\dot{\vect{x}}^{CA3}(t)$. 
Each intrinsic pattern is then back-projected via EC to SI to visualize the input $\dot{\vect{x}}^{SI}(t+1)$.   
This implements the main idea of CRISP~\citep{cheng2013crisp}, \emph{i.e.}\ CA3 severs as an intrinsic sequence generator and plasticity happens mainly in the feed-forward connections of the hippocampal circuit. As shown in the experiments, this is a key requirement for recalling entire sequences from a single cue. 

\subsection{Model Assumptions and Simplifications}\label{sec:model_assumptions_and_simplifications}

For simplicity we do not distinguish between the deep and superficial layers in EC, supported by experimental evidence that the layers in EC act in unity~\citep{kloosterman2000functional}.
Notice that this does not mean (unless explicitly mentioned in the experiments) that the output activities in EC project back to the current input in EC, as the input and output pathways through EC are considered at different time steps in any case.

In our experiments we have seen little to no effect on the quality of the retrieval of patterns in EC when the hippocampal subregion CA1 was also incorporated into the model.

We have chosen the size of the subregions in terms of the number of neurons according to anatomical data~\citep[see][Fig.~1]{NeherChengEtAl-2015}, where we use $N$ to denote the size of the model. 
The number of neurons of all subregions is given relative to $N$ such that EC is $1.1 N$, CA3 is $2.5 N$, and DG is $12 N$, which allows us to scale the model comfortably with $N$.
Notice that $N = 100.000$ would lead to a model of original size of the rat hippocampus.
As this is computationally too expensive, we consider models of size $N =200$ and $N = 1000$, which are scaled down by a factor of 500 and 100, respectively, compared to the actual size of the rat hippocampus.

We have chosen higher average activity, \emph{i.e.}\ the proportion of cells that are active on average at any given time, compared to recordings from the rat hippocampus~\citep[see][Tab. 1]{NeherChengEtAl-2015}. 
That is 20\% instead of 3.2\% for CA3 and 3\% instead of 0.78\% for DG.
We argue that this requirement in our models is due to the fact that the model is 100 to 500 times smaller than the rat hippocampus.
If we would choose $N=100$ with an average activity 3.2\% there would only be eight neurons active in CA3 on average. 
But it is clear that independent of the actual pattern size one needs a certain number of active neurons for robust pattern association. 
As shown in the experiments one can choose sparser activities as the model gets larger without reducing the capacity, so that a model of original size will most likely work with average activities recorded from the rat hippocampus. 
Furthermore, we show that the required average activities also depend on the type of input data.

We consider an all-to-all connectivity between the subregions although the number of connections is much lower according to anatomical recordings from the rat hippocampus~\citep[see][Fig~1.]{NeherChengEtAl-2015}.
In our experience the choice of a reduced fixed network connectivity, \emph{e.g.} random, locally, etc. has a significant impact on the network performance especially for smaller models and might thus lead to wrong conclusions when analyzing the entire system. 
We thus prefer to use an all-to-all connectivity between the subregions and let the network determine the optimal connectivity on its own. 
The distribution of the weight values after learning follow a Gaussian-distribution where most of the values are close to zero and thus imply a sparser connectivity in any case (results not shown).

Notice, that we store one very long sequence of length $N$ rather than several shorter sequences in the model. 
However, this is only to simplify the analysis, one could equivalently store a number of short intrinsic sequences in CA3~\citep[see also][]{Bayati2018}.

\subsection{Artificial Neuron Model}

In this work we consider rate-based models in which each neuron is represented by a centered artificial neuron~\citep{MontavonMueller-2012, Melchior2016}
\begin{eqnarray}
	h_j &=& \phi \left(a_j\right) = \phi \left(\sum_i^{N_{in}} \left( x_i-\mu_i \right) w_{ij}  +b_j \right),\label{eqn:neuron_element}
\end{eqnarray}
with pre-synaptic input $x_i$, membrane potential $a_j$, output activity $h_j$, bias $b_j$, weight $w_{ij}$, offset value $\mu_i$ (usually the target activity of the layer, see Figure~\ref{fig:model_EC_DG_CA3}), and activation function $\phi \left( \cdot \right)$.
In this work we use the Sigmoid and Step activation functions, but others could be used instead, which in our experience often leads to similar results.
By centering, \emph{i.e.}\ subtracting the mean from the inputs, the pre-synaptic activations become mean free. 
It has been shown~\citep{Melchior2019a} that centering is a necessary requirement for one-shot storage of patterns in artificial neural networks independently of the used learning rule. Even in batch learning, although biologically not plausible, centering often leads to better results especially in case of unsupervised learning~\citep{Melchior2016}.

\subsection{Learning Rule}

We use Hebbian-descent~\citep{Melchior2019a} as a learning rule for all subregions, which is a stable, and biologically plausible learning rule for hetero-association as well as auto-association.
Hebbian-descent performs significantly better than Hebb's rule and the covariance rule in general and significantly better than gradient descent in online learning. 
Furthermore, Hebbian-descent provides an automatic mechanism of forgetting, see~\citep{Melchior2019a} for a detailed comparison of all methods.
For hetero-associating an input activity $x_i$ with a desired output activity $t_j$ the Hebbian-descent updates for weight and bias are given by
\begin{eqnarray}
\delta w_{ij} &=& -\eta (x_i- \mu_i)(h_j - t_j),\label{eqn:hetero_update_hippo_w}\\
\delta b_{j} &=& -\eta (h_j -  t_j),\label{eqn:hetero_update_hippo_b}
\end{eqnarray}
with learning rate $\eta$.
Hebbian-descent therefore measures the correlation between the network's centered input and the difference between the network's produced output and desired output.
In case of auto-association the updates for weight and bias are given by
\begin{eqnarray}
\delta w_{ij} &=& -\eta  (h_j - \lambda_j )(z_i - x_i) ,\label{eqn:auto_update_hippo_w}\\
\delta c_{i} &=& -\eta (z_i -  x_i),\label{eqn:auto_update_hippo_c}\\
\delta b_{j} &=& -\eta (h_j -  \tilde \lambda_j), \,\,\,\,\,\,\,\,\,\,\,\,\,\,\,\,\,\,\,\,\,\,\,\text{(optional)}\label{eqn:auto_update_hippo_b}
\end{eqnarray}
with learning rate $\eta$, hidden offset $\lambda_j$ (\emph{i.e.} the mean of the output unit $j$),  desired average hidden activity $\tilde \lambda_j$, and reconstructed input 
\begin{equation}
z_i = \phi \left(\sum_j^M \left(h_j-\lambda_j \right) w_{ij}  + c_i \right),\label{eqn:neuron_element_decoder} 
\end{equation}
with hidden bias $c_j$. 
Notice, that analogously to contrastive Hebbian learning~\citep{RumelhartMcClellandEtAl-1986} this update rule is only local if we introduce time such that $\vect x$ and $\vect z$ are the post synaptic activities of the same neurons at two different time steps.

\subsection{Training of the Individual Subregions}
The training of the specific subregions is described in the following subsections and the detailed calculations of the network are given in Appendix~\ref{appendix:calculation_flow_models}.

\subsubsection{Pre-Training CA3 as an Intrinsic Sequence Generator}\label{sec:pretrain_CA3_description}

A stable and robust CA3 network is crucial for the performance of the entire system, as it fully determines the sequential order of the patterns.
If the intrinsic dynamics does not return the correct successive pattern  $\vect{x}^{CA3}(t+1)$ given $\vect{x}^{CA3}(t)$, the model is not be able to retrieve the correct sequence of patterns in CA3 and consequently also not in EC. 
Patterns of the intrinsic sequence in CA3 should thus not be too similar as they could easily interfere with each other. 
Hence, they should ideally be orthogonal to each other or at least have a large pairwise Euclidean distance, which has been analyzed in detail by~\citet{Bayati2018}.
We therefore pre-trained CA3 on a cyclic sequence of $N$ independently sampled binary random patterns $\vect{x}^{CA3}(1), \vect{x}^{CA3}(2), \cdots , \vect{x}^{CA3}(N)$, where each random pattern has an average activity of exactly 20\% (if no other average activity is explicitly mentioned in the experiments).
We used the hetero-associative Hebbian-descent update given by Equations~\eqref{eqn:hetero_update_hippo_w} and~\eqref{eqn:hetero_update_hippo_b}, where in each update step the input corresponds to one of the patterns $\vect{x}^{CA3}(t)$ and the desired output corresponds to its successive pattern $\vect{x}^{CA3}(t+1)$.
Pre-training of CA3 is performed for 100 epochs, \emph{i.e.} each pattern is seen 100 times by the network, which is necessary to learn the associations between all patterns perfectly.
We used a mini batch size of 10, \emph{i.e.} the update is the average over 10 samples, a learning rate of 1.0, and in each epoch we randomly flipped 10\% of the input values to induce more robustness with respect to noise in the input. 
Notice that only for pre-training we use multiple epochs, and online learning otherwise.
According to the chosen average activity of 20\% the offset values are set to $\mu_i = 0.2$.
Theoretically we can store any set of $2.5N$ patterns perfectly in our model, \emph{i.e.} when each neuron in CA3 is active for exactly one pattern\footnote{Notice that depending on statistics of input and CA3 patterns this value might be larger, but it is the guaranteed theoretical bound for any set of patterns.}.
We therefore define the theoretical capacity of our model to be $2.5N$. 
Due to the difficulties in online learning the empirical capacity is only about $N$ and therefore $\frac{100N}{2.5N}=40\%$ of the theoretical capacity.

\subsubsection{Pre-Training DG as a Pattern Separator}

Our DG model is generic, which means that it is independent of the actual input statistics and is the same for all datasets.
This is achieved by training the connections EC~$\rightarrow$~DG using the Hebbian-descent update given by Equations~(\ref{eqn:auto_update_hippo_w}\,--\,\ref{eqn:auto_update_hippo_b}) on binary random data. 
It can formally be seen as training an auto encoder with tied weights and sigmoid non-linearity. 
The hidden representation, which corresponds to DG, is regularized to have a hidden activity of approximately 3\%, which is achieved by setting hidden offsets $\lambda_j$ as well as the desired activities $\tilde \lambda_j$   to 0.03 if not mentioned otherwise. 
The network is trained on 4000 random patterns with the EC activity of 35\%, so that $\mu_i$ is set to 0.35.
We used a single epoch and a mini batch size of 10 resulting in 400 updates using a very large learning rate of 100.
After training, the encoder part of the auto encoder (\emph{i.e.}\ EC~$\rightarrow$~DG) is used to transform any pattern with an activity of 35\% to a 10.9 times larger DG pattern with an activity of approximately 3\%. 
See~\citep{Melchior2019a} for a detailed illustration of auto encoders trained with Hebbian-descent.
Notice that a network trained in this way is much more robust with respect to noise compared to a randomly initialized network.
A similar effect has been observed for recurrent networks~\citep{Bayati2018}.

\subsubsection{Pre-Training the Transformation of the Sensory Input}

When sensory input data is provided we train the pathways SI~$\rightarrow$~EC and EC~$\rightarrow$~SI on the entire training data using the auto-associative Hebbian-descent update given by Equation~(\ref{eqn:auto_update_hippo_w}\,--\,\ref{eqn:auto_update_hippo_b}). 
This can again be formally seen as training an auto encoder with tied weights and step function as output non-linearity to transform the potentially real-valued input data into a binary representation. 
The visible offset $\mu_i$ is set to the mean of the dataset and by setting the hidden offset $\lambda_j$ as well as the desired activities $\tilde \lambda_j$ to 0.35, we ensure that the activity in EC is approximately 35\%.
We used a mini-batch size of 100, a learning rate of 0.01, and a momentum (Fraction of the historical update added to the current update) of 0.9 for 10 epochs on the entire training data. 
After training, the encoder part of the network (\emph{i.e.}\ Equation~\ref{eqn:neuron_element}) transforms an input data-point $\vect x^{SI}(t)$ to a binary representation $\vect{x}^{EC}(t)$ with an activity of approximately 35\%.
The decoder part of the network  (\emph{i.e.}\ Equation~\ref{eqn:neuron_element_decoder}), which transforms the binary pattern in EC back to the input SI, can then be used to visualize the retrieved patterns in the input domain.

\subsubsection{One-Shot Storage in Plastic Pathways}

For training the hetero-associative pathways EC~$\rightarrow$~CA3, and CA3~$\rightarrow$~EC for \linebreak[4]{\emph{Model-A}}, and DG~$\rightarrow$~CA3, and CA3~$\rightarrow$~EC for \emph{Model-B} we use the hetero-associative Hebbian-descent update given by Equation~\eqref{eqn:hetero_update_hippo_w} and~\eqref{eqn:hetero_update_hippo_b}.
Here we use the online version to perform one-shot learning, meaning that each pattern pair is only seen once by the network and there is only one update step per pattern pair.
Thus there is only one comparison of pre- and post synaptic activity per pattern.
For the pathway EC~$\rightarrow$~CA3 for example, $\vect{x}^{EC}(t)$ is the input and $\vect{x}^{CA3}(t)$ is the desired output pattern for which a single update step is performed before the next pair of patterns is presented. 
The input offsets $\mu_i$ are initialized to the average activity of the corresponding subregion, \emph{i.e.}\ 0.35 for the pathway EC~$\rightarrow$~CA3, 0.2 for CA3~$\rightarrow$~EC, and 0.03 for DG~$\rightarrow$~CA3.
Although a constant learning rate of 0.1 works fairly well, a learning rate that takes the network size into account performs better in our experience. 
We found empirically that the simple learning rate 
\begin{equation}
\eta = \frac{20}{\text{N}},
\end{equation}
works well in practice for networks of size $N = 20$ to $2000$.
As Hebbian-descent controls the gradual forgetting of patterns automatically, we do not need a weight decay for stability as required when using Hebb's rule for example.

\subsubsection{Re-Training of Plastic Pathways (Dreaming)}\label{sec:retraining_of_plastic_pathways_dreaming}

A noteable property of our model is that it is capable of `bootstrapping' itself by re-training the pathway EC~$\rightarrow$~CA3 without external input.
Starting from a randomly selected intrinsic pattern $\vect{x}^{CA3}(t)$, the model recalls the corresponding EC pattern $\vect{\tilde{x}}^{EC}(t)$. 
The mapping (EC~$\rightarrow$~CA3) is then strengthened using hetero-associated Hebbian-descent with $\vect{\tilde{x}}^{EC}(t)$ as input and $\vect{x}^{CA3}(t)$ as target values.
The intrinsic pattern triggers the next intrinsic pattern and the whole process repeats.
For this retraining we recall the entire sequence 10 times in sequential order such that each pattern is seen another 10 times. 
However, randomly recalling the same number of intrinsic patterns works similarly well.
Although the quality of the recalled patterns is not perfect, in particular for older memories, it helps to improve the associations stored in the pathway EC~$\rightarrow$~CA3, as our experiments show. 
Re-training can even decorrelate stored patterns, so that \emph{Model-A} can recall a sequence of correlated input patterns in SI/EC after re-training, as also shown in the experiments.
As memories are recalled to stabilize the model's recall mechanism in an off\-line phase of the model (no external input), we call this process `dreaming'.
Notice, that `dreaming' would not be possible  with Hebb's rule or the covariance rule as both update rules do not improve by seeing patterns several times~\citep{Melchior2019a}.

\section{Methods}\label{sec:methods}

The following sections describe the methods and the input datasets the model is trained and evaluated on.

\subsection{Input Sequences}\label{sec:input_sequences}

Patterns in EC are either generated artificially or represent real world data that has been transformed to a binary representation using the pathway SI~$\rightarrow$~EC.

The artificial dataset {\bf{\emph{RAND}}} contains uncorrelated binary random patterns.
According to the average activity in EC, 35\% randomly selected pixels in each pattern take the value one and the remaining 65\% of the pixels are set to zero.

The second artificial dataset {\bf{\emph{RAND-CORR}}} contains temporally correlated binary random patterns that are generated as follows: 
In a first step an initial pattern is generated where 35\% of the pixels are set to one and the rest is set to zero. 
In a second step we select 10\% of all pixels, half of them with value one and half of them with value zero, and flip their values.
The second step is now repeated on the generated pattern to form the next pattern and so on. 
While the average correlation of the patterns in the dataset is close to zero each pattern has a high correlation of 0.8 with its direct predecessor.

The {\bf{\emph{MNIST}}}~\citep{LeCun1998} dataset consists of 70,000 gray scale images of handwritten digits divided into training and test set of 60,000 and 10,000 patterns, respectively. The images have a size of $28 \times 28$ pixels, where all pixel values are normalized to lie in a range of $[0,1]$. The dataset is not binary, but the values tend to be close to zero or one. Each pattern is assigned to one out of ten classes representing the digits 0 to 9. 

The {\bf{\emph{CIFAR}}}~\citep{Krizhevsky-2009a} dataset consists of 60,000 color images of various objects divided into training, validation, and test set with 40,000, 10,000, and 10,000 patterns, respectively. The images have a size of $32 \times 32$ pixels, which are converted to gray scale and normalized per pixel position to zero mean and unit variance. 
Each pattern is assigned to one out of ten classes representing trucks, cats, or dogs for example. 

An input sequence for the models is generated by randomly selecting $N$ patterns of the training data of one of the datasets.
For \emph{MNIST} and \emph{CIFAR}, the pathway SI~$\rightarrow$~EC is trained on the entire training data.

\subsection{Evaluation}

It is common to use correlation as an intuitive performance measure in computational neuroscience. 
And in staying consistent with related studies~\citep{NeherChengEtAl-2015, Bayati2018} we use the Pearson correlation coefficient to measure the performance between retrieved pattern $\dot{\vect x}$ and the ground truth pattern $\vect x$ in each subregion
\begin{equation}
Corr(\dot{\vect x}, \vect x) :=
\frac{\langle(\dot{\vect x}-\langle \dot{\vect x}  \rangle)(\vect x-\langle \vect x  \rangle)\rangle}{\sqrt{ \langle(\dot{\vect x}-\langle \dot{\vect x} \rangle)^2\rangle}\sqrt{\langle(\vect x-\langle \vect x  \rangle)^2\rangle}},
\end{equation}
where $D$ is the dimensionality of the patterns and $\langle \vect{x}\rangle_d = \frac{1}{D}\sum_{d=1}^{D} x_d$.
Using the absolute error or squared error as a performance measure leads to qualitatively the same results.
For non-artificial datasets we additionally show the reconstructed patterns to visualize the performance of the network.

\section{Results}

The following experiments analyze the performance of \emph{Model-A} and \emph{Model-B} on different datasets.
All plots show the correlation between retrieved and ground truth patterns in a particular subregion in solid blue.
Additionally a corresponding trendline is given in dashed orange as well as a baseline in dashed-dotted green. The latter is the correlation between the retrieved patterns $\dot{\vect{x}}$ and the mean of the entire sequence of ground truth patterns $\langle \vect{x}(t) \rangle_t$. 
The baseline thus reflects the trivial solution if the network had just learned to output the mean pattern independently of the input.
Each plot shows the result for one simulation, but each simulation was repeated with different initializations and input sequences to verify that the qualitative outcome stays the same and therefore does not depend on a particular initialization.
In the following we refer to pathway EC~$\rightarrow$~CA3 in \emph{Model-A} and pathway EC~$\rightarrow$~DG~$\rightarrow$~CA3 in \emph{Model-B} as encoder, as they encode the EC patterns into an intrinsic pattern.
Similarly we refer to pathway CA3~$\rightarrow$~EC as decoder and CA3~$\rightarrow$~CA3 as intrinsic dynamics, in both models.

\subsection{Storing a Sequence of Uncorrelated Patterns in \emph{Model-A}}

In a first experiment we store a sequence of 1000 uncorrelated binary random patterns (dataset \emph{RAND}) in \emph{Model-A} with size N=1000.

\subsubsection{One-Shot Encoding and Decoding Performance}\label{one_shot_encoding_and Decoding_performance}

\begin{figure}[htbp!]
\begin{center}
\subfigure[Encoder]{
\includegraphics[scale=0.4, trim=10 5 32 40, clip]{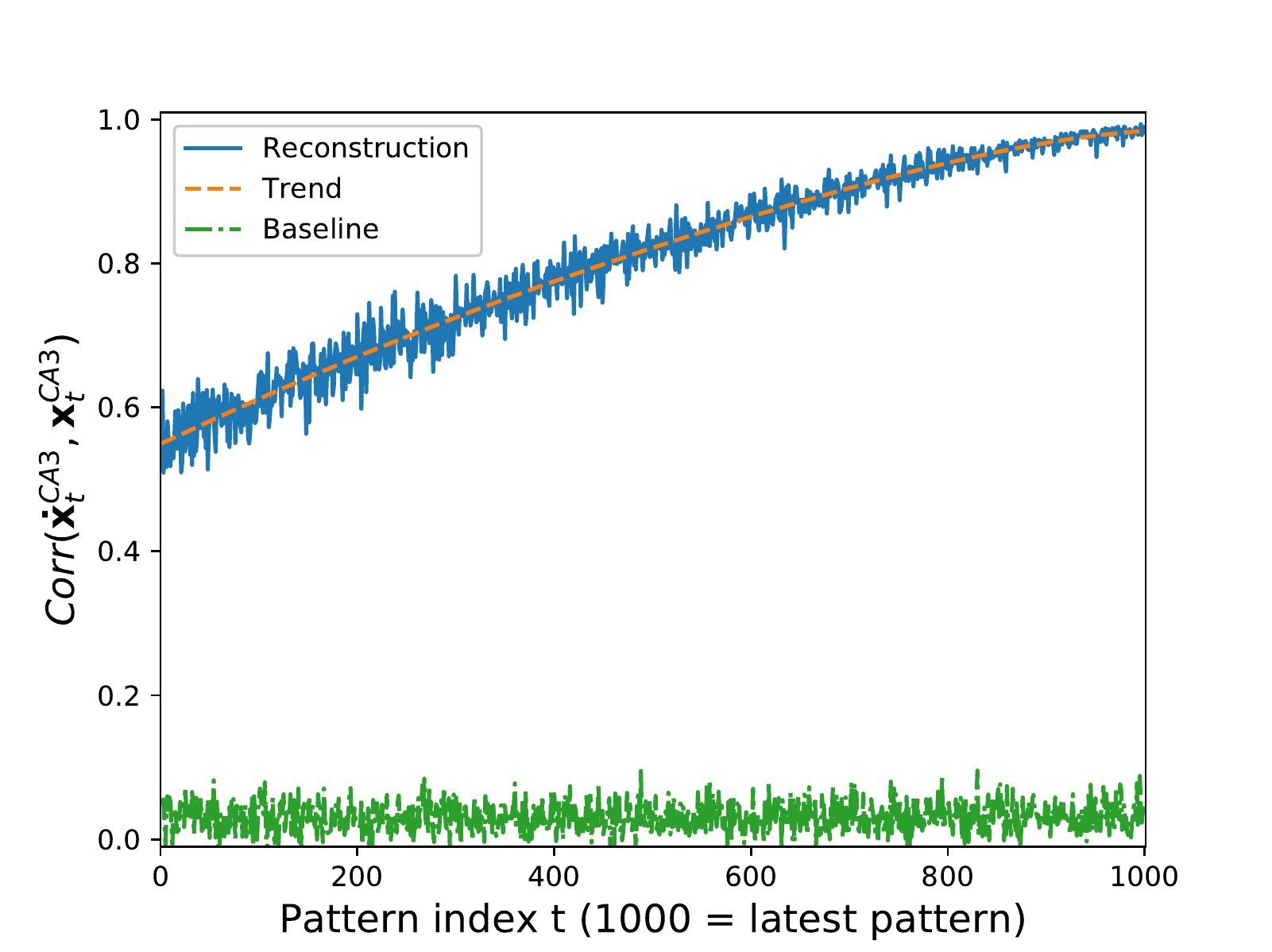}\label{fig:model_A_uncorr_encoder}}
\subfigure[Decoder]{
\includegraphics[scale=0.4, trim=10 5 32 40, clip]{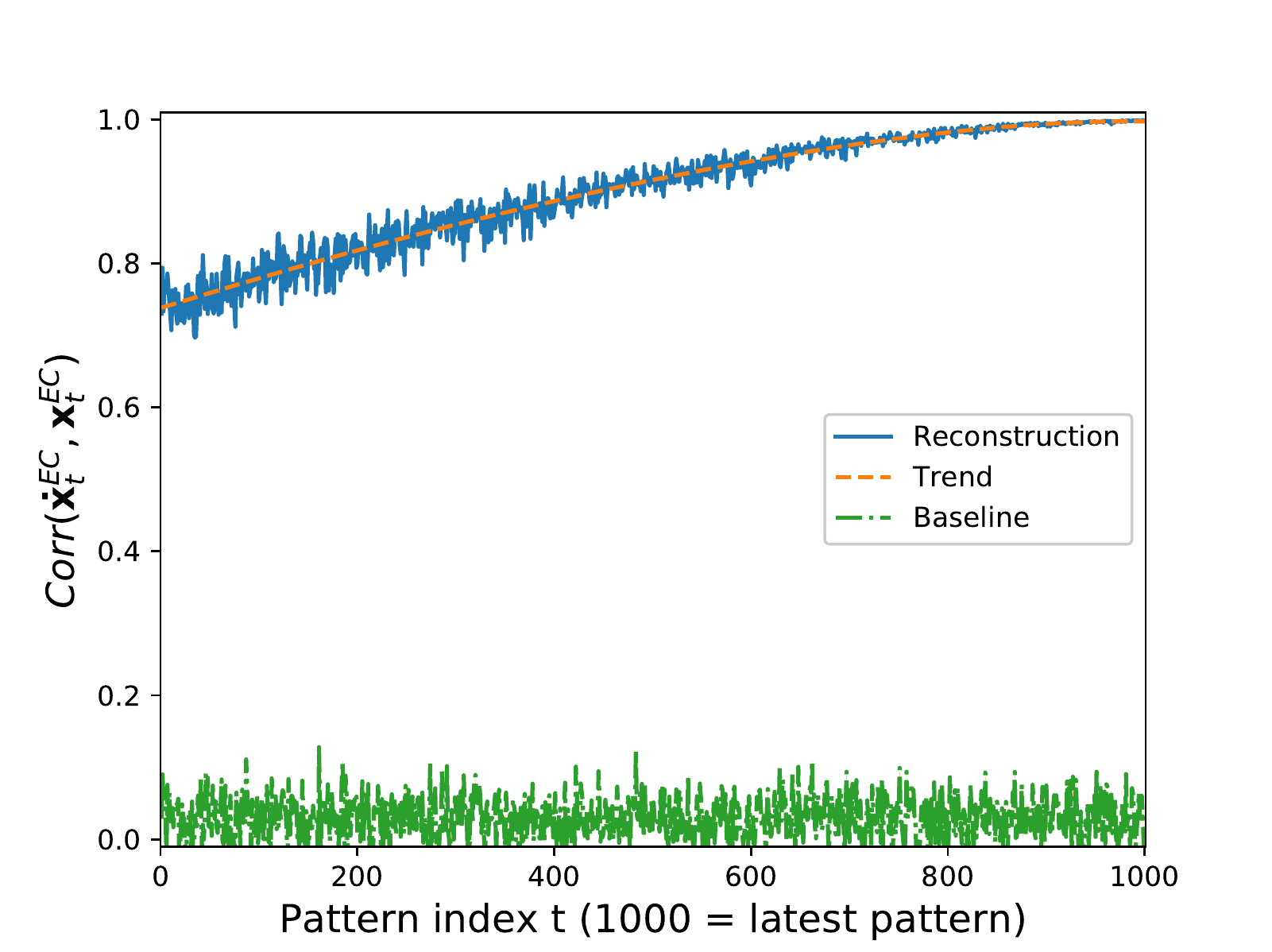}\label{fig:model_A_uncorr_decoder}}
\caption{Encoding and decoding performance of \emph{Model-A} on the \emph{RAND} dataset for (a)~encoder ({\tiny $\vect{\vect{x}}^{EC}(t)\rightarrow \dot{\vect{x}}^{CA3}(t)$}) and (b) decoder ({\tiny $\vect{\vect{x}}^{CA3}(t)\rightarrow \dot{\vect{x}}^{EC}(t)$}). For clarity the intrinsic dynamics has not been used yet and in all plots the input pattern is the ground truth.
The baseline denotes the correlation between retrieved patterns and the mean pattern of the entire sequence of ground truth patterns (\emph{i.e.} $Corr${\tiny  $\left(\dot{\vect{x}}^{CA3}(t),\langle \vect{\vect{x}}^{CA3}(t) \rangle_t \right)$} and $Corr$ {\tiny   $\left(\dot{\vect{x}}^{EC}(t),\langle \vect{\vect{x}}^{EC}(t) \rangle_t \right)$}, respectively). 
Notice, that all plots shows the result for a single simulation, but each simulation was repeated with different initializations and input sequences to verify that the qualitative outcome does not depend on a particular initialization.
}
\label{fig:model_A_uncorr_encoder_decoder}
\end{center}
\end{figure}

\begin{figure}[htbp!]
\begin{center}
\includegraphics[scale=0.23, trim=0 0 0 100, clip]{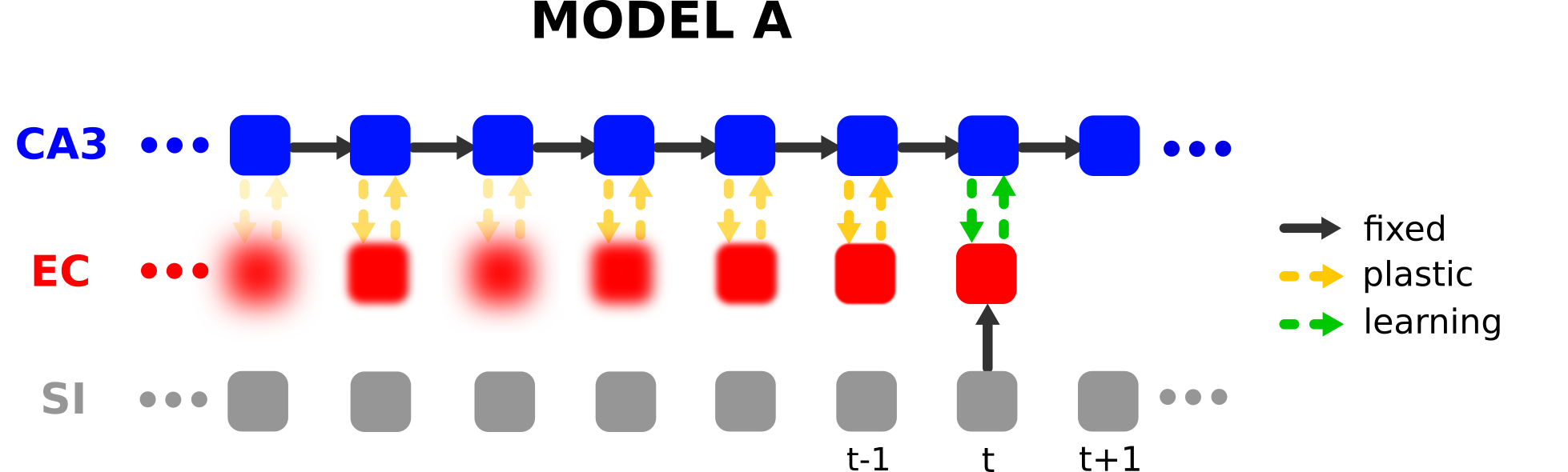}
\caption{Illustration of hetero-association and forgetting over time in \emph{Model-A}. At time step $t$ the EC pattern, inferred from the SI pattern, is hetero-associated in both directions (indicated by the green dashed arrows) with the corresponding CA3 pattern. The learned associations weaken over time (indicated by the increasing transparency of the arrows), leading to a degraded reconstruction (forgetting) in EC (indicated by the increasing blur).
}
\label{fig:schema_ModelA}
\end{center}
\end{figure}

The separate performance of encoder and decoder are shown in Figure~\ref{fig:model_A_uncorr_encoder} and~\ref{fig:model_A_uncorr_decoder}, respectively.
Both plots show 
a correlation of approximately one for the pattern that has been stored last (index 1000). 
The correlation degrades gradually for patterns stored earlier (smaller pattern index), which indicates forgetting.
The results in Figure~\ref{fig:model_A_uncorr_encoder_decoder} are consistent with the experimental observations that forgetting follows a power law~\citep{wixted1990analyzing}, which in this case is almost linear. 
For clarity we illustrated in Figure~\ref{fig:schema_ModelA} how the EC patterns are hetero-associated with the corresponding CA3 patterns in \emph{Model-A} and how these associations weaken over time.
The speed of forgetting is larger for the encoder than for the decoder, as it is easier to hetero-associate high (CA3) with low (EC) dimensional patterns than \emph{vice versa}.
Larger input patterns, at least in the uncorrelated case, provide more information/dimensions that can be associated with the output, so that for the same number of patterns the performance in EC for larger dimensional input is better than for lower dimensional input.

\begin{figure}[htbp!]
\begin{center}
\subfigure[Encoder + decoder]{
\includegraphics[scale=0.4, trim=10 5 32 40, clip]{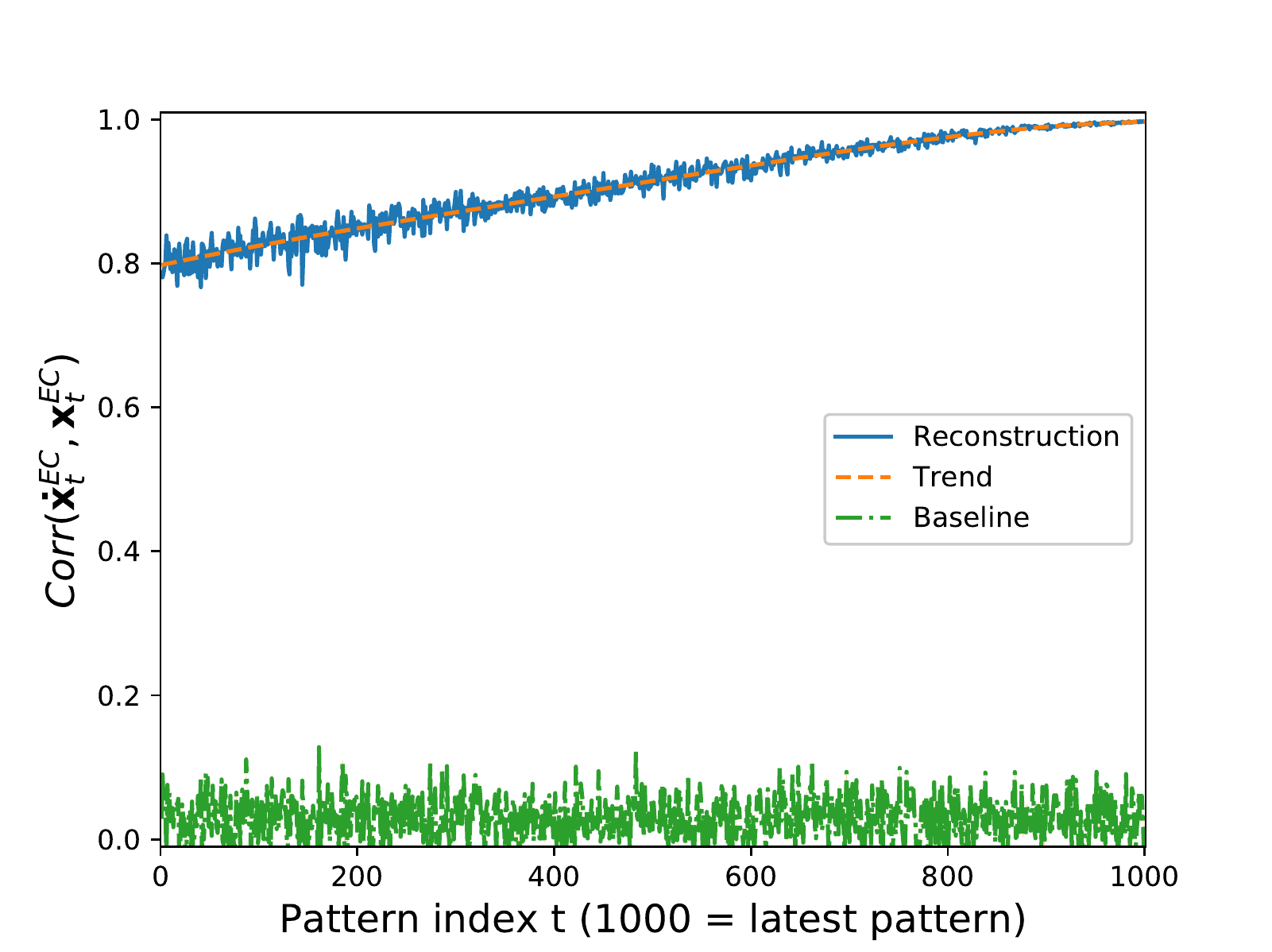}\label{fig:model_A_uncorr_encode_decode}}
\subfigure[Intrinsic recall for 1 transition]{
\includegraphics[scale=0.4, trim=10 5 32 40, clip]{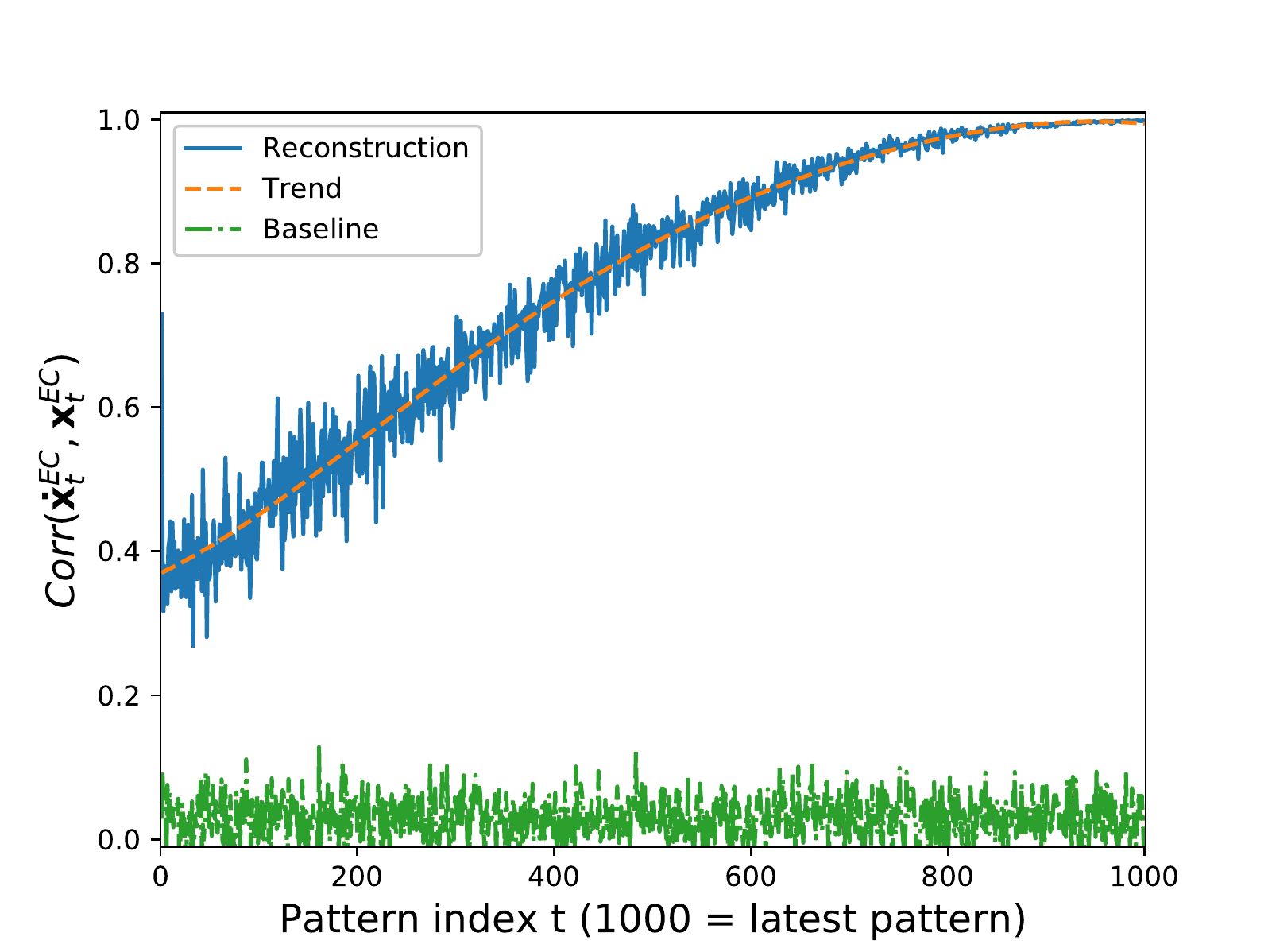}\label{fig:model_A_uncorr_full_loop_1_step}}
\subfigure[Intrinsic recall for 5 transitions]{
\includegraphics[scale=0.4, trim=10 5 32 40, clip]{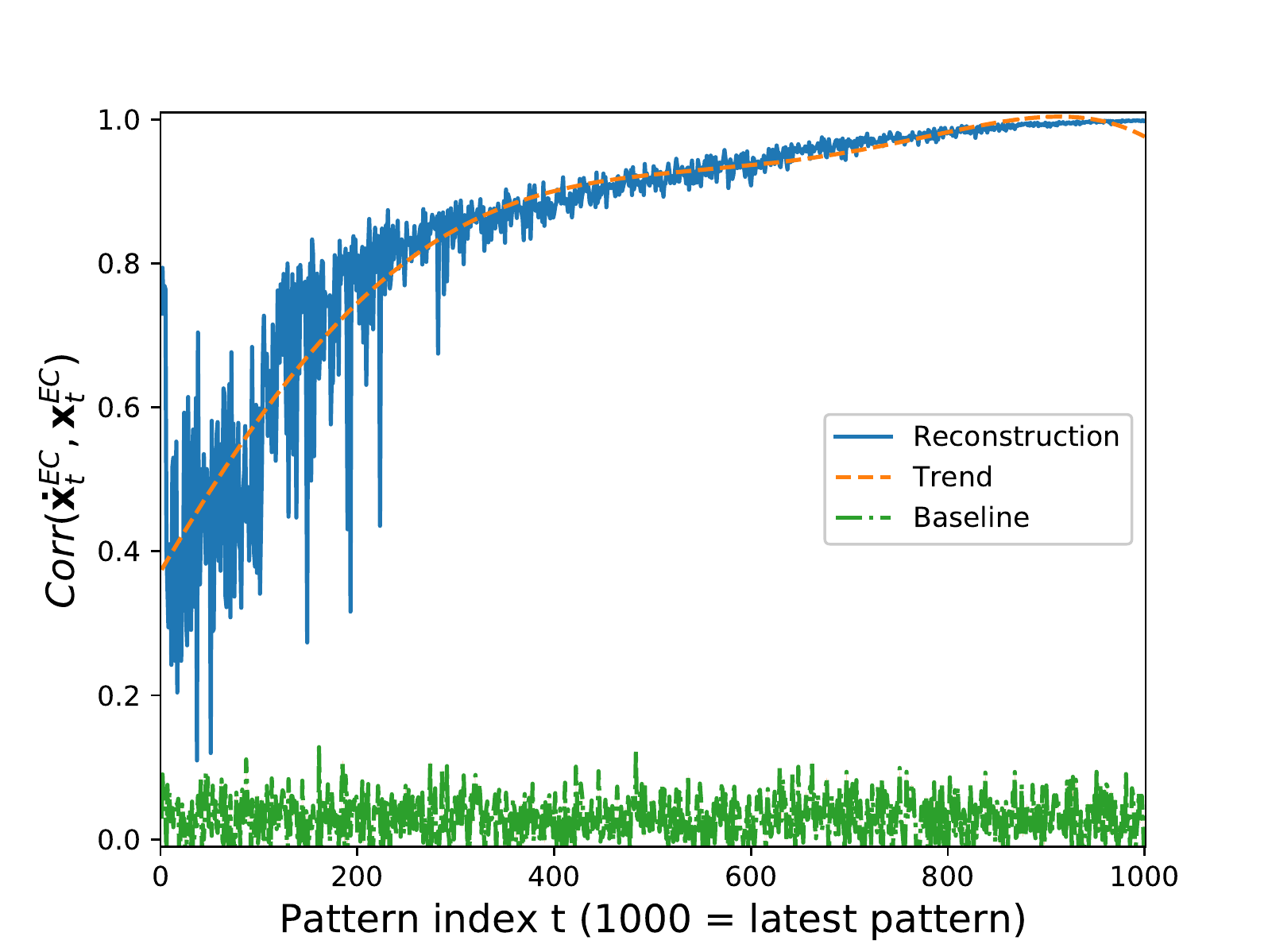}\label{fig:model_A_uncorr_intrinsic_5_steps}}
\subfigure[Intrinsic recall for 1000 transitions]{
\includegraphics[scale=0.4, trim=10 5 32 40, clip]{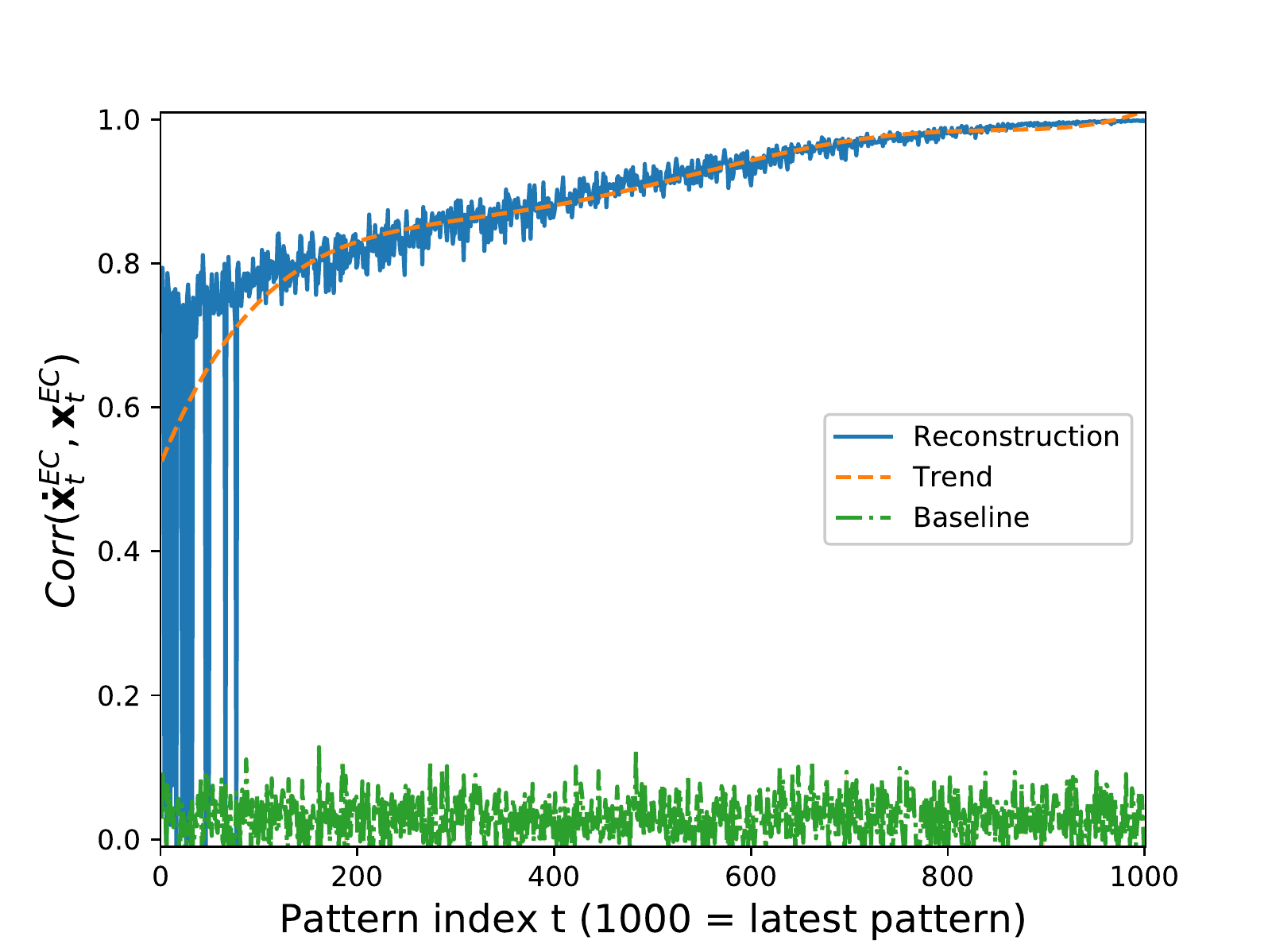}\label{fig:model_A_uncorr_intrinsic_all}}
\caption{Recall performance of \emph{Model-A} on the \emph{RAND} dataset.
Notice that the intrinsic sequence in CA3 is cyclic, so that the successive pattern for {\tiny $\vect{x}^{CA3}(T)$} is {\tiny $\vect{x}^{CA3}(1)$}.
(a) each pattern is encoded and directly decoded without intrinsic dynamics ({\tiny $\vect{\vect{x}}^{EC}(t)\rightarrow \dot{\vect{x}}^{CA3}(t)\rightarrow \dot{\vect{x}}^{EC}(t)$}), (b) each pattern is encoded followed by one intrinsic transition, and the corresponding pattern is decoded ({\tiny $\vect{\vect{x}}^{EC}(t-1)\rightarrow \dot{\vect{x}}^{CA3}(t-1)\rightarrow \dot{\vect{x}}^{CA3}(t)\rightarrow \dot{\vect{x}}^{EC}(t)$}), (c)  each pattern is encoded, the intrinsic transition is iterated five times, and the corresponding pattern is decoded ({\tiny $\vect{\vect{x}}^{EC}(t-5)\rightarrow \dot{\vect{x}}^{CA3}(t-5)\rightarrow \dot{\vect{x}}^{CA3}(t-4)\rightarrow \dot{\vect{x}}^{CA3}(t-3)\rightarrow \cdots \rightarrow \dot{\vect{x}}^{CA3}(t)\rightarrow \dot{\vect{x}}^{EC}(t)$}), and (d) each pattern is encoded, the intrinsic transition is looped fully through ($T$ transitions) arriving at pattern $t$ again, and the corresponding pattern is decoded ({\tiny $\vect{\vect{x}}^{EC}(t)\rightarrow \dot{\vect{x}}^{CA3}(t)\rightarrow \dot{\vect{x}}^{CA3}(t+1) \rightarrow \dot{\vect{x}}^{CA3}(t+2) \rightarrow \cdots \rightarrow \dot{\vect{x}}^{CA3}(T) \rightarrow \dot{\vect{x}}^{CA3}(t+T-1) \rightarrow \dot{\vect{x}}^{CA3}(t~+~T~=~t) \rightarrow \dot{\vect{x}}^{EC}(t)$}). 
}
\label{fig:model_A_uncorr_all}
\end{center}
\end{figure}

Figure~\ref{fig:model_A_uncorr_encode_decode} shows the performance of Model-A, combining the encoder and decoder. The overall performance in EC is slightly higher (by about 0.0089 on average) than that of Figure~\ref{fig:model_A_uncorr_decoder}, but with a higher correlation for early patterns and a lower correlation for late patterns.
The better performance of the decoder when providing imperfect EC patterns compared to the perfect EC pattern is counterintuitive.
However, it seems to be a property of the correlation coefficient, which compares observation and ground truth on a multiplicative base.
If we measure the performance with absolute or mean squared error, which compares observation and ground truth on a additive base, the performance is lower for the imperfect patterns as one would expect.

\subsubsection{Recall Performance}\label{Intrinsic_Recall_Performance_MODEL_A}

\begin{figure}[htbp!]
\begin{center}
\subfigure[Encoder + dynamics (1 transition)]{
\includegraphics[scale=0.4, trim=10 5 32 40, clip]{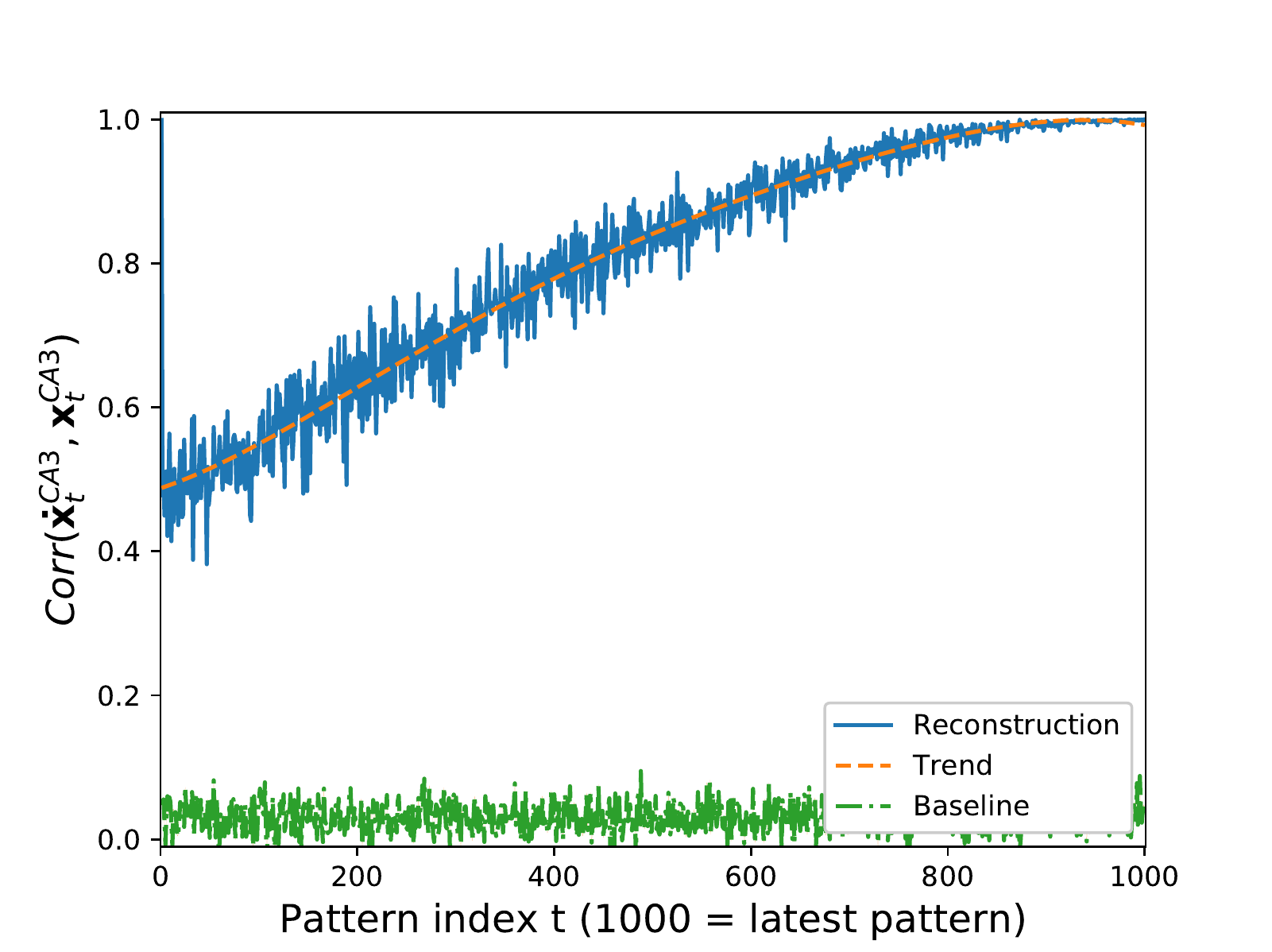}\label{fig:model_A_uncorr_encoder_dynamic_one_step}}
\subfigure[Encoder + dynamics (5 transitions)]{
\includegraphics[scale=0.4, trim=10 5 32 40, clip]{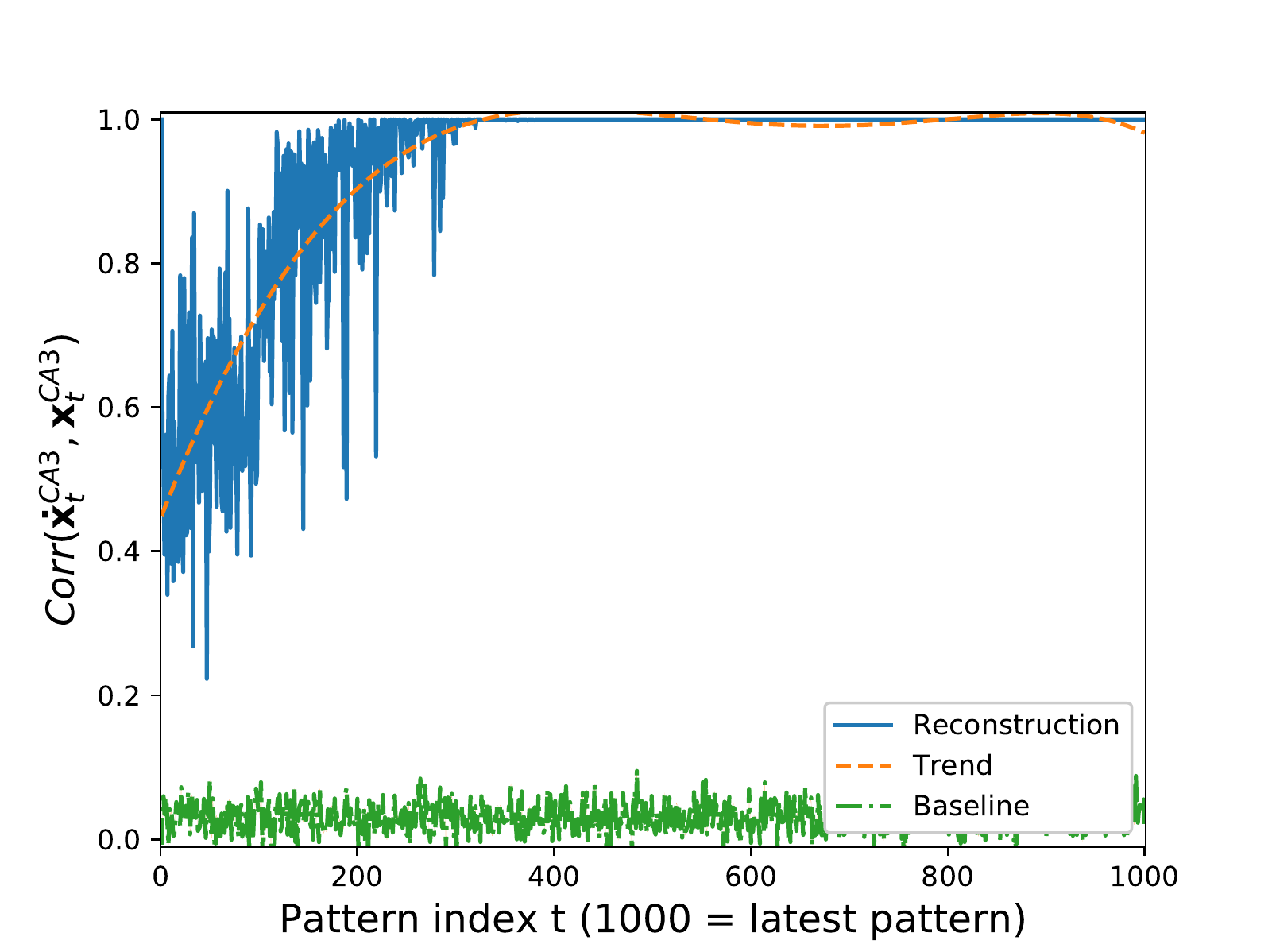}\label{fig:model_A_uncorr_encoder_dynamic_five_step}}
\caption{Intrinsic performance in CA3 of \emph{Model-A} on the \emph{RAND} dataset, where each pattern is encoded and the intrinsic transition is iterated (a) once 
({\tiny $\vect{\vect{x}}^{EC}(t-1)\rightarrow \dot{\vect{x}}^{CA3}(t-1)\rightarrow \dot{\vect{x}}^{CA3}(t)$})
and (b) five times ({\tiny $\vect{\vect{x}}^{EC}(t-5)\rightarrow \dot{\vect{x}}^{CA3}(t-5)\rightarrow \dot{\vect{x}}^{CA3}(t-4) \cdots \dot{\vect{x}}^{CA3}(t)$}).
}
\label{fig:model_A_encoder_intrinsic}
\end{center}
\end{figure}

\begin{figure}[htbp!]
\begin{center}
\includegraphics[scale=0.3, trim=0 0 0 0, clip]{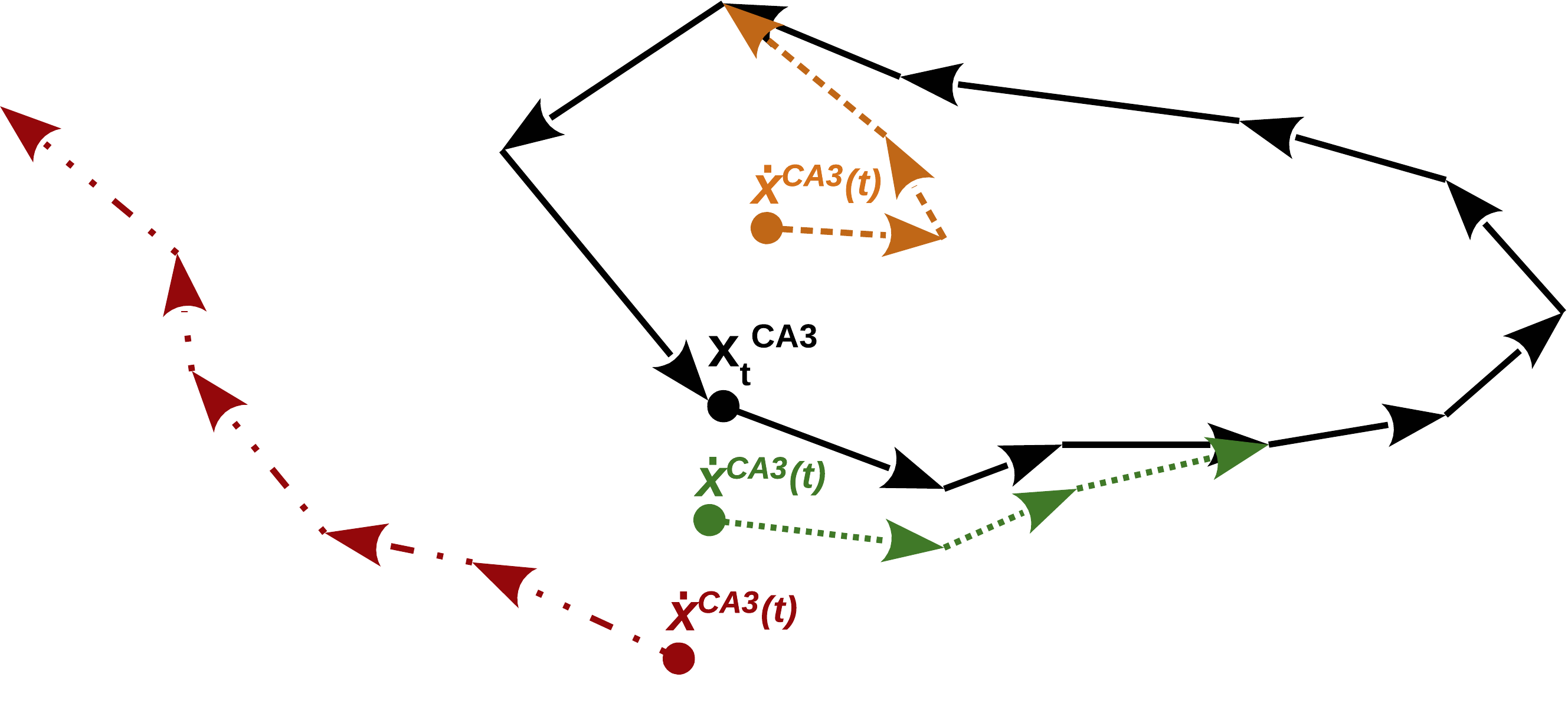}
\caption{Illustration of the sequence relaxation in CA3 with corrupted cues $\dot{\vect{x}}^{CA3}(t)$. The solid black arrows represent the transitions between the patterns of the intrinsic sequence.  If a corrupted cue is given to CA3 the system will 1.~(green dotted arrows) converge within a short `sequence completion phase' to the right position in the intrinsic sequence if the cue is similar enough to the ground truth pattern, 2.~(orange dashed arrows) converge to a wrong position in the intrinsic sequence if the cue is more similar to another pattern in the intrinsic sequence, or 3.~(red dashed-dotted arrows) converge to a spurious intrinsic sequence if the cue is similar to a pattern within a spurious sequence.}
\label{fig:error_illustration}
\end{center}
\end{figure}

If one adds a single intrinsic transition between encoding and decoding, the correlation decreases significantly for early patterns as shown in~\ref{fig:model_A_uncorr_full_loop_1_step}.
However, if more intrinsic transitions are added the correlation increases again as shown for five transitions in Figure~\ref{fig:model_A_uncorr_intrinsic_5_steps} and 1000 transitions in Figure~\ref{fig:model_A_uncorr_intrinsic_all}. 
After full intrinsic recall (1000 transitions) the performance in EC is almost equivalent to that of the decoder (Figure~\ref{fig:model_A_uncorr_decoder}) except for some very early patterns.
The CA3 network is thus able, after a significant degradation in the first transition, to denoise or complete the pattern within a couple of intrinsic transitions in most cases.
This is illustrated in Figure~\ref{fig:model_A_encoder_intrinsic}, which shows the correlation between retrieved and ground truth patterns in CA3, when the EC patterns are encoded and the intrinsic transition is iterated (a) once and (b) five times.
The performance in CA3 for one transition decreases similar to that in EC (Figure~\ref{fig:model_A_uncorr_full_loop_1_step}), whereas after five transitions the performance is almost perfect for the 700 latest patterns.
When more transitions are performed the performance of most patterns gets perfect except for some `old' outliers that can be seen in Figure~\ref{fig:model_A_uncorr_intrinsic_all}.
For those exceptions the intrinsic dynamics either relaxes on a wrong position in the intrinsic sequence such that the output is just a shifted version of the stored sequence or they relax to a spurious sequence, where none of the retrieved patterns is similar to the corresponding ground truth pattern.
For the majority of the patterns, however, it is possible to recall the entire sequence in correct order within a short `sequence completion phase' (\emph{i.e,}\ phase until the CA3 dynamics has relaxed on the intrinsic sequence), which is longer for early pattern.
The three possibilities of sequence relaxation in CA3, which depend on the cue quality are illustrated in Figure~\ref{fig:error_illustration}.

An initial decrease of performance in CA3 within the first intrinsic transition followed by a steady increase as more intrinsic transitions are performed is consistent with the findings by~\citet{Bayati2018}, who have already observed this in the case of off\-line learning \citep[see][Fig. 7 top left]{Bayati2018}.
While in off\-line learning all patterns are stored at the same time, and therefore the denoising property of CA3 is equally important for all patterns, in online learning we have shown that it is less important for recently stored patterns and becomes more important for older patterns. 

\subsection{Storing Sequences through Plasticity in CA3}\label{sec:comparison_standard}

\begin{figure}[htbp!]
\begin{center}
\includegraphics[scale=0.23, trim=0 0 0 100, clip]{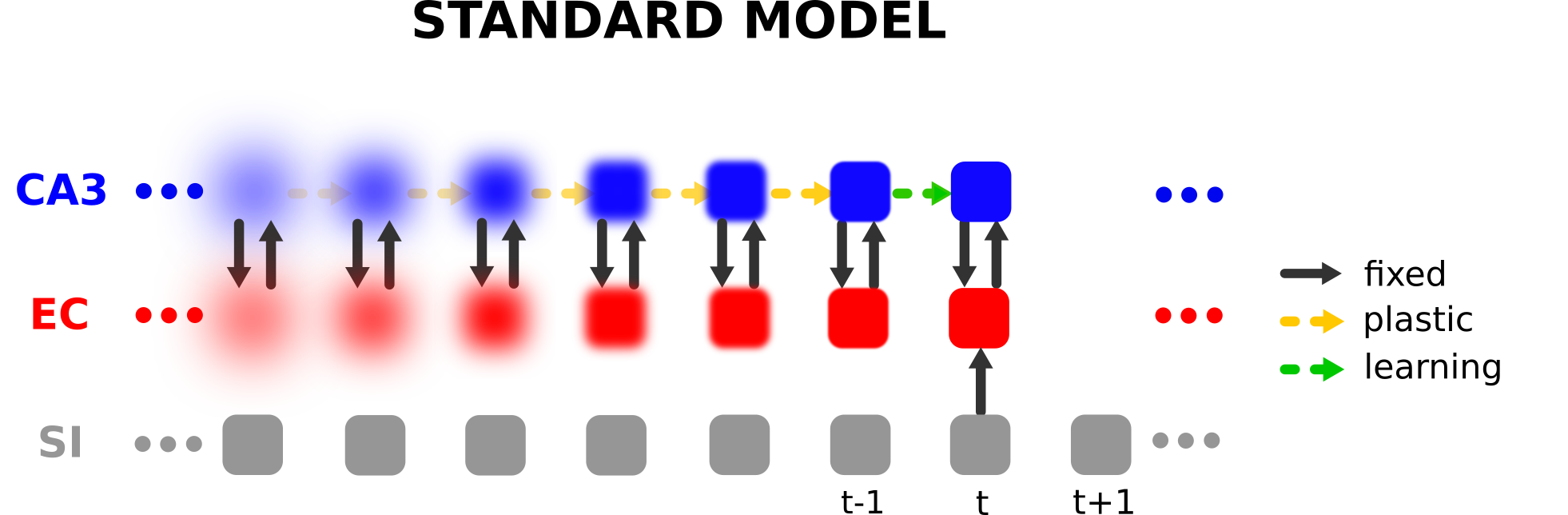}
\caption{Illustration of hetero-association and forgetting over time in the standard framework. At time step $t$ the current EC pattern (inferred from the corresponding SI pattern) is projected to CA3 via a fixed pathway EC$\rightarrow$CA3 (here we simply used the identity mapping) and hetero-associated (indicated by the green dashed arrows) with the previous CA3 pattern. The learned associations weaken over time indicated by the increasing transparency of the arrows, leading to a degraded reconstruction (forgetting) in CA3 and thus also in EC indicated by the increasing blur. Also compare with the illustration for \emph{Model-A} and \emph{Model-B} in Figure~\ref{fig:schema_ModelA} and Figure~\ref{fig:schema_ModelB}, respectively.
}
\label{fig:schema_standard}
\end{center}
\end{figure}

\begin{figure}[htbp!]
\begin{center}
\subfigure[Recall performance (learning rate 0.01)]{
\includegraphics[scale=0.4, trim=10 5 32 40, clip]{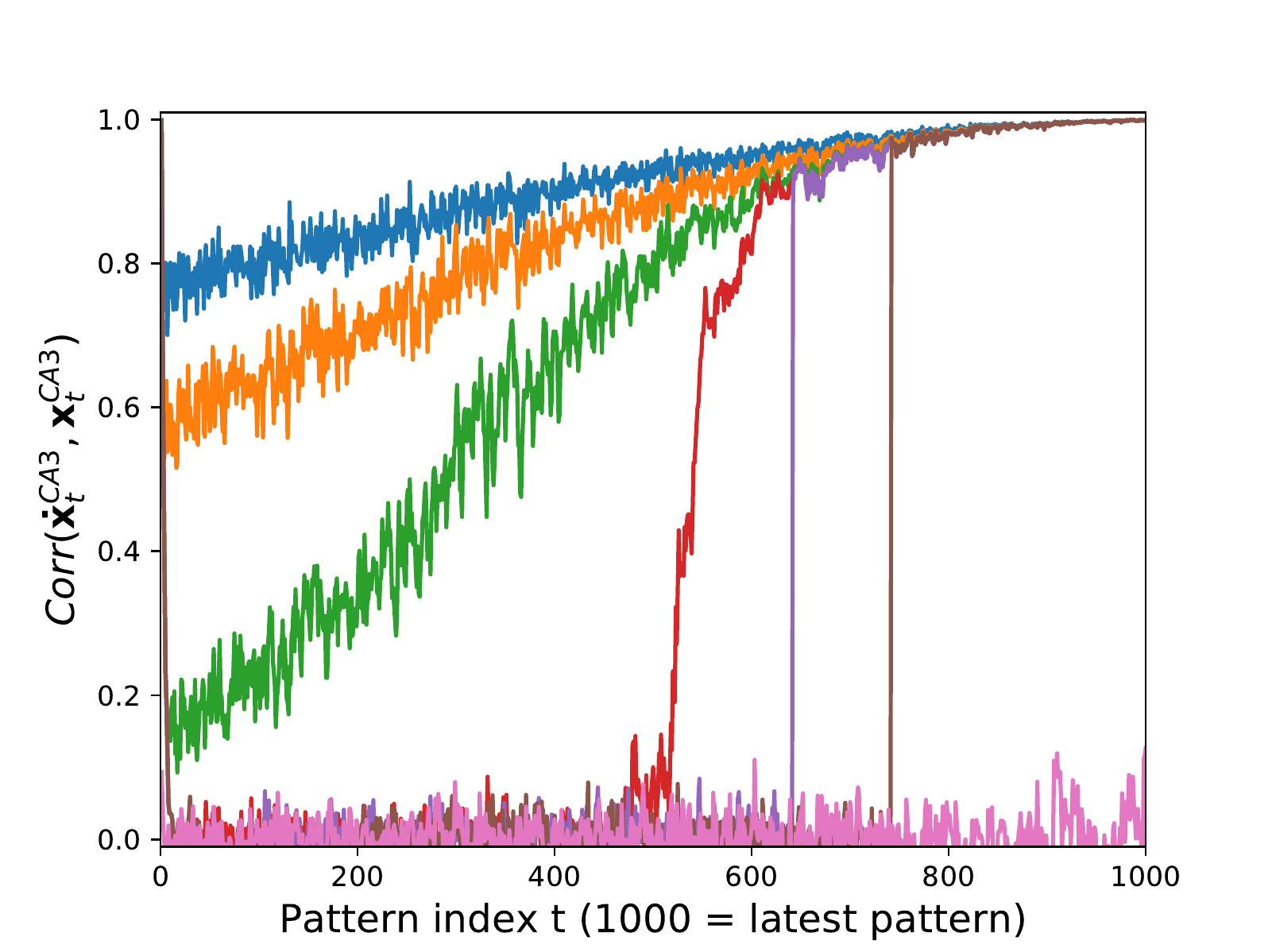}\label{fig:model_standard_performance_0025}}
\subfigure[Recall performance (learning rate 0.025)]{
\includegraphics[scale=0.4, trim=10 5 32 40, clip]{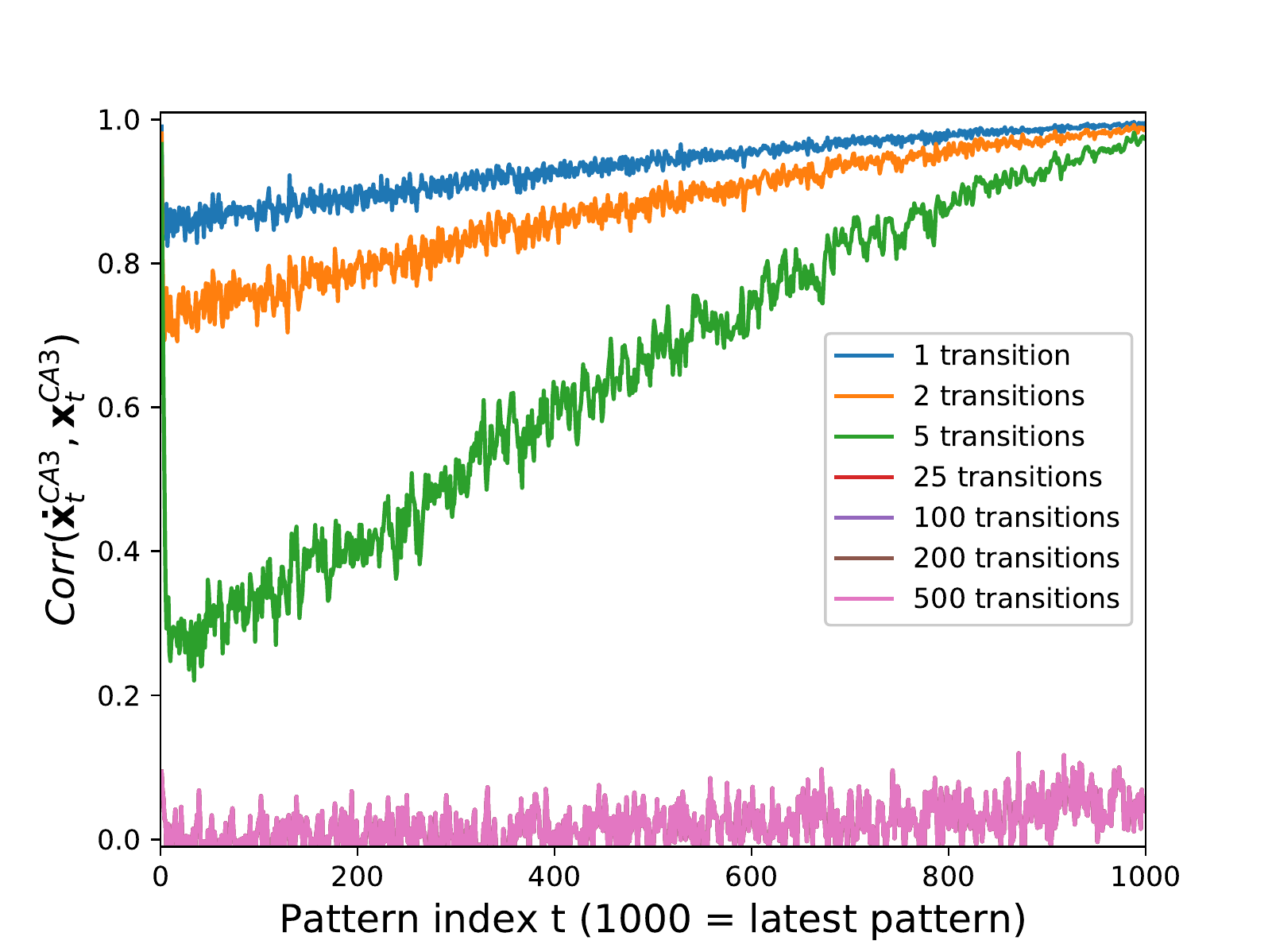}\label{fig:model_standard_performance_001}}
\caption{
Recall performance in the standard framework, where uncorrelated CA3 patterns have been hetero-associated online one pattern pair at a time
 with a learning rate of (a) 0.01 and (b) 0.025 . 
Shown is the network performance after a different number of iterations through the recurrent CA3 dynamics (e.g. for 5 transitions {\tiny $\vect{x}^{CA3}(t-5)\rightarrow \dot{\vect{x}}^{CA3}(t-4) \cdots \dot{\vect{x}}^{CA3}(t)$}).
Notice, that 
the curves for 25, 100, and 200 transitions in (b) are almost perfectly overlaid by the curve for 500 transitions and thus not visible.}
\label{fig:model_standard_performance}
\end{center}
\end{figure}

Plasticity in the standard framework~\citep{nadel1997memory} is assumed in the recurrent connections of CA3, so that online storage of sequences is performed by 
 hetero-associating the previous with the current pattern in CA3 one pattern pair at a time~\citep{Levy-1996}.
This process is illustrated in Figure~\ref{fig:schema_standard}.
We performed experiments on online hetero-association of uncorrelated patterns one pattern pair at a time, using the same CA3 sequence with the same dimensionality and activity as in the previous experiments. We also evaluated the recall performance for a different number of transitions, \emph{i.e.}\ asking the question how well a pattern following a starting pattern after 1, 2, 5, 25, ... time steps can be recalled.
The results are shown in Figure~\ref{fig:model_standard_performance_0025}, where predicting only the next pattern works reasonably well, and even better than in \emph{Model-A} (compare with Figure~\ref{fig:model_A_uncorr_full_loop_1_step}).
However, as the number of transitions increases the performance decreases rapidly so that recalling patterns after 25 transitions only works for the most recent third of patterns and recalling for 500 transitions and more does not work for any of the patterns.
While the performance for this model decreases rapidly with the number of transitions, the performance for \emph{Model-A} increases with the number of iterations as the model is able to recover through the sequence relaxation process in CA3 as illustrated in Figure~\ref{fig:model_A_encoder_intrinsic}).
Notice, that the process is also very sensitive to the learning rate as illustrated in Figure~\ref{fig:model_standard_performance_001}. Using a learning rate smaller than 0.025 or larger than 0.01 leads to a much worse performance (results not shown).
In terms of online sequence storage the standard framework is thus no reasonable alternative to the concept of intrinsic sequences as proposed by the CRISP framework.

\subsection{Storing a Sequence of Temporally Correlated Patterns in \\\emph{Model-A}}\label{sec:storing_temporally_correlated_patterns_in_model_A}

In the previous section we have shown that the model can recall sequences of uncorrelated patterns from a single input pattern. 
However, assuming uncorrelated input patterns is a rather unrealistic assumption, it is more realistic to assume correlated patterns such as grid cell activity~\citep{solstad2006grid, hafting2008hippocampus}.
We therefore trained the same model as in the previous section but using the \emph{RAND-CORR} dataset in which patterns are temporally correlated, \emph{i.e.} successive patterns have a correlation of 0.8.

\subsubsection{One-Shot Encoding and Decoding Performance}

\begin{figure}[htbp!]
\begin{center}
\subfigure[Encoder]{
\includegraphics[scale=0.4, trim=10 5 32 40, clip]{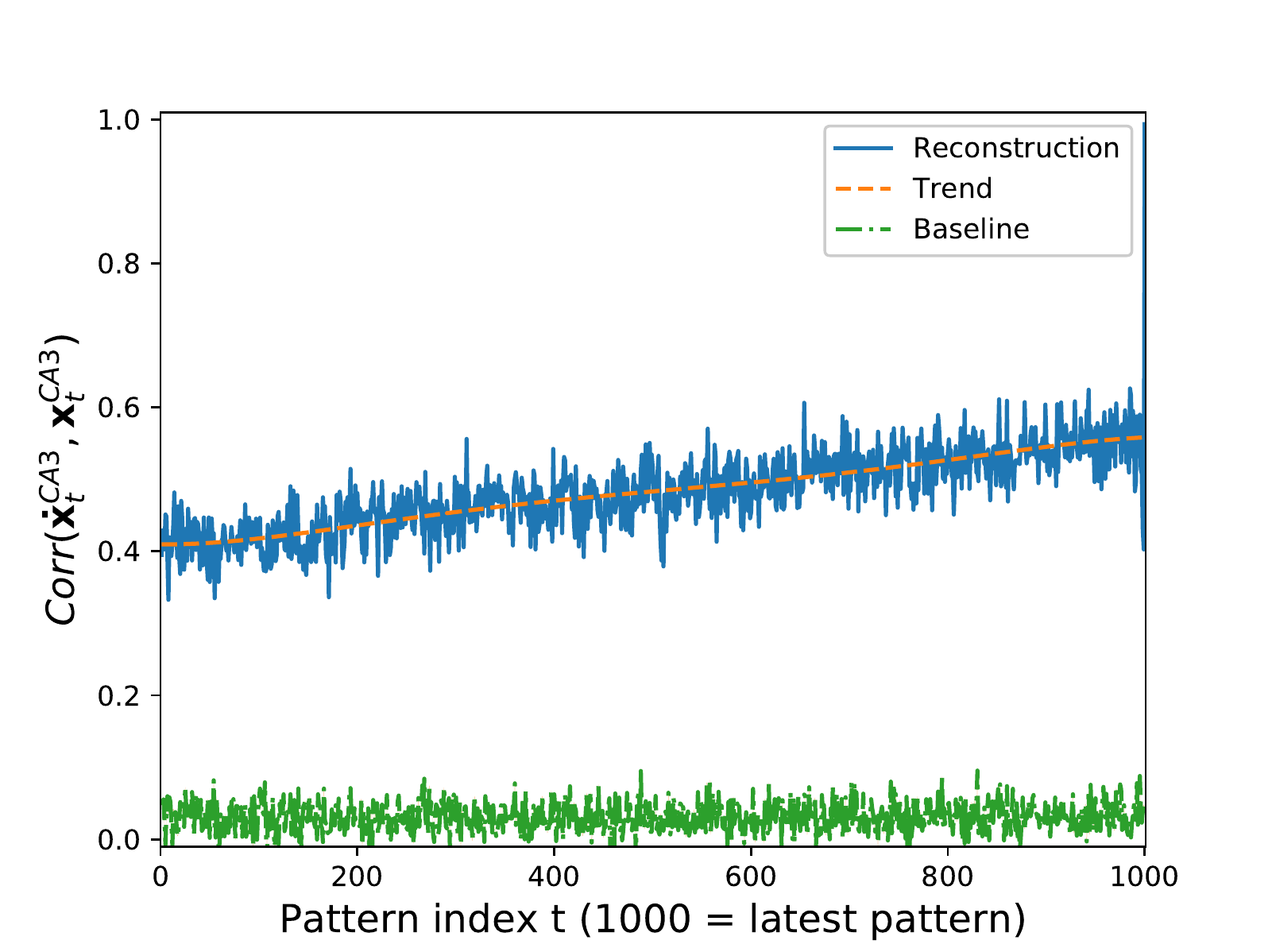}\label{fig:model_A_crosscorr_encoder}}
\subfigure[Decoder]{
\includegraphics[scale=0.4, trim=10 5 32 40, clip]{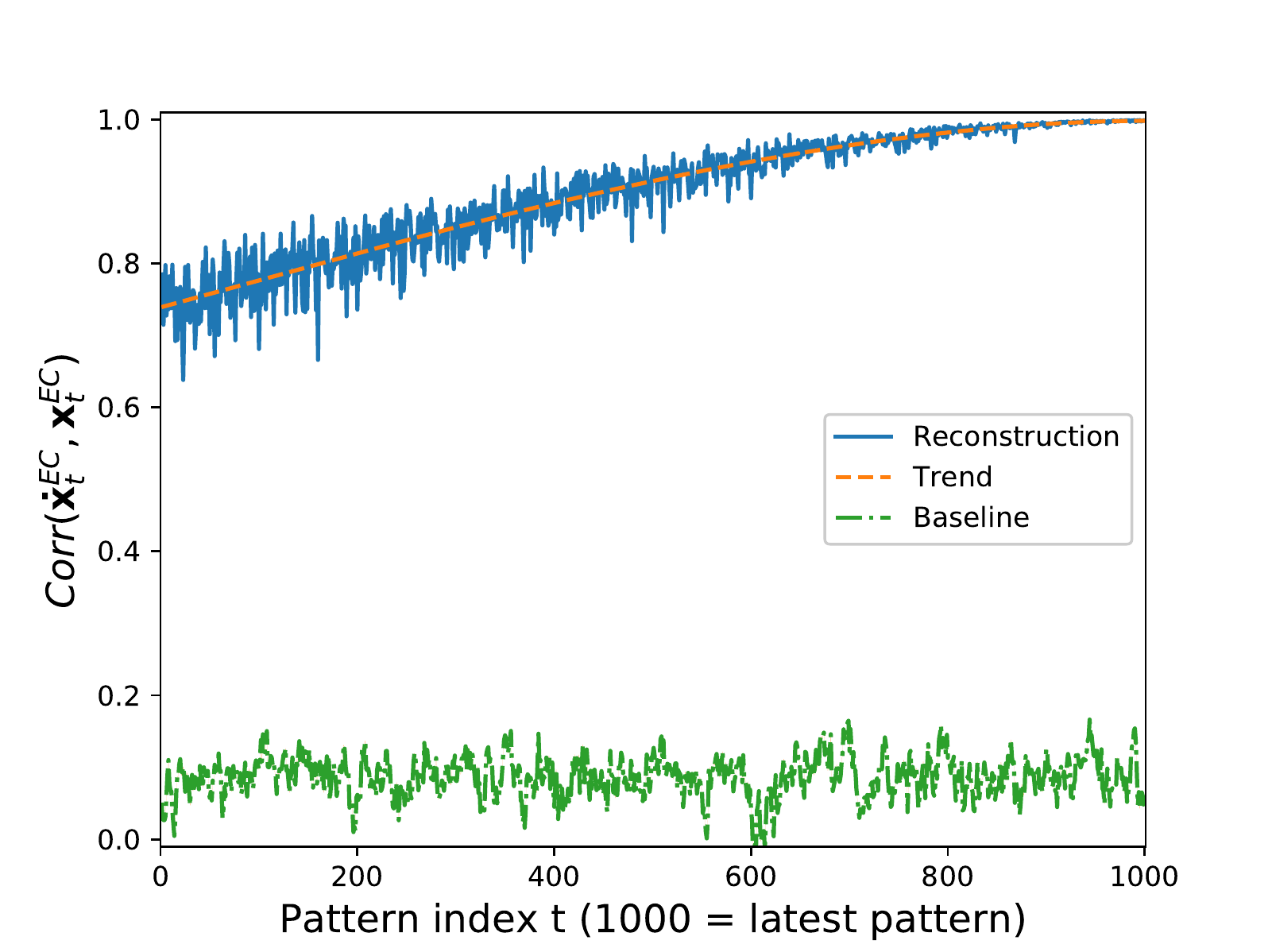}\label{fig:model_A_crosscorr_decoder}}
\caption{
Encoding and decoding performance of \emph{Model-A} on the \emph{RAND-CORR} dataset for (a) encoder ({\tiny $\vect{\vect{x}}^{EC}(t)\rightarrow \dot{\vect{x}}^{CA3}(t)$}) and (b) decoder ({\tiny $\vect{\vect{x}}^{CA3}(t)\rightarrow \dot{\vect{x}}^{EC}(t)$}). For clarity the intrinsic dynamics has not been used and see also Figure~\ref{fig:model_A_uncorr_encoder_decoder} for comparison with uncorrelated data.
}
\label{fig:model_A_crosscorr_encoder_decoder}
\end{center}
\end{figure}

Figure~\ref{fig:model_A_crosscorr_encoder} shows the performance of the encoder, which is much worse compared to the performance when uncorrelated patterns are stored as shown in Figure~\ref{fig:model_A_uncorr_encoder}. 
In contrast, the performance of the decoder as shown in Figure~\ref{fig:model_A_crosscorr_decoder} is very similar to the performance when using uncorrelated input patterns as shown in Figure~\ref{fig:model_A_uncorr_decoder}.
This illustrates that one can easily associate a set of very different input patterns (uncorrelated) with a set of very similar output patterns (correlated) even in online learning, but the reverse association is difficult.
Notice however that both association can be learned perfectly with this network when performing mini-batch learning for 100 epochs (results not shown), so that this is only problematic in online learning.
Although the encoder performs poorly, the performance in EC is surprisingly good when a pattern is encoded and directly decoded as shown in Figure~\ref{fig:model_A_crosscorr_encode_decode}. 

\subsubsection{Recall Performance}

\begin{figure}[t!]
\begin{center}
\subfigure[Encoder + decoder]{
\includegraphics[scale=0.4, trim=10 5 32 40, clip]{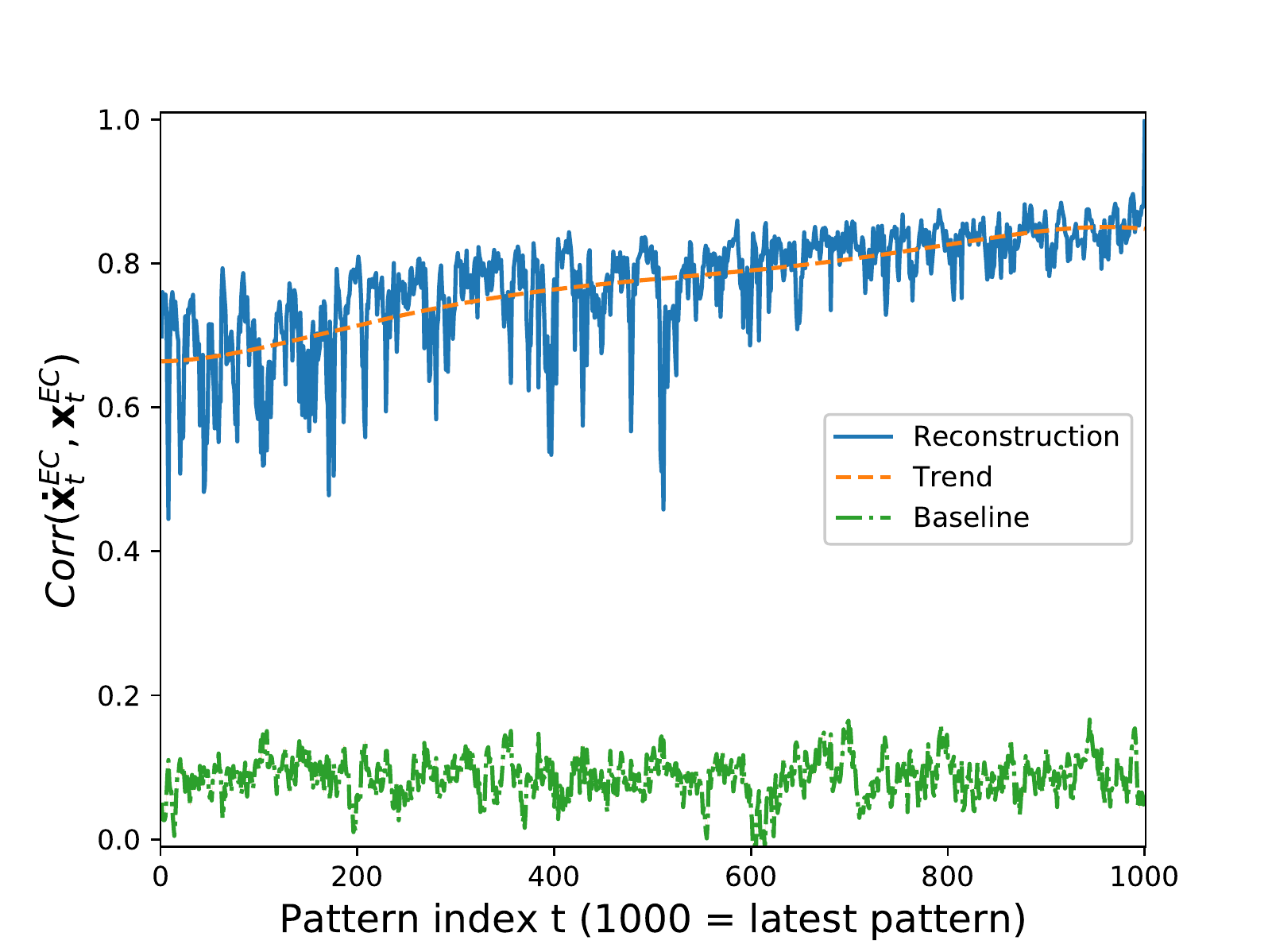}\label{fig:model_A_crosscorr_encode_decode}}
\subfigure[Intrinsic recall for 1 transition]{
\includegraphics[scale=0.4, trim=10 5 32 40, clip]{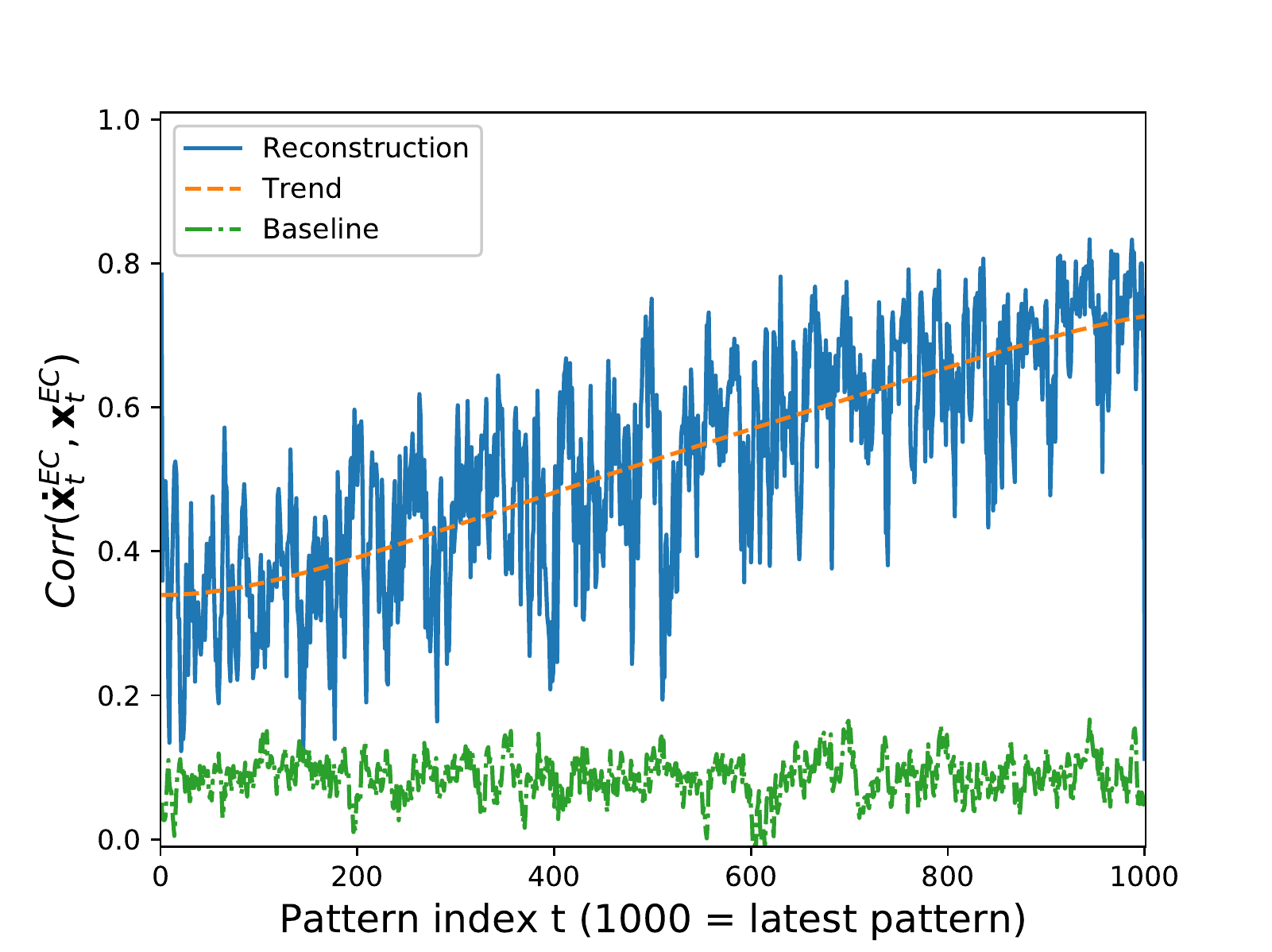}\label{fig:model_A_crosscorr_full_loop_1_step}}
\subfigure[Intrinsic recall for 5 transitions]{
\includegraphics[scale=0.4, trim=10 5 32 40, clip]{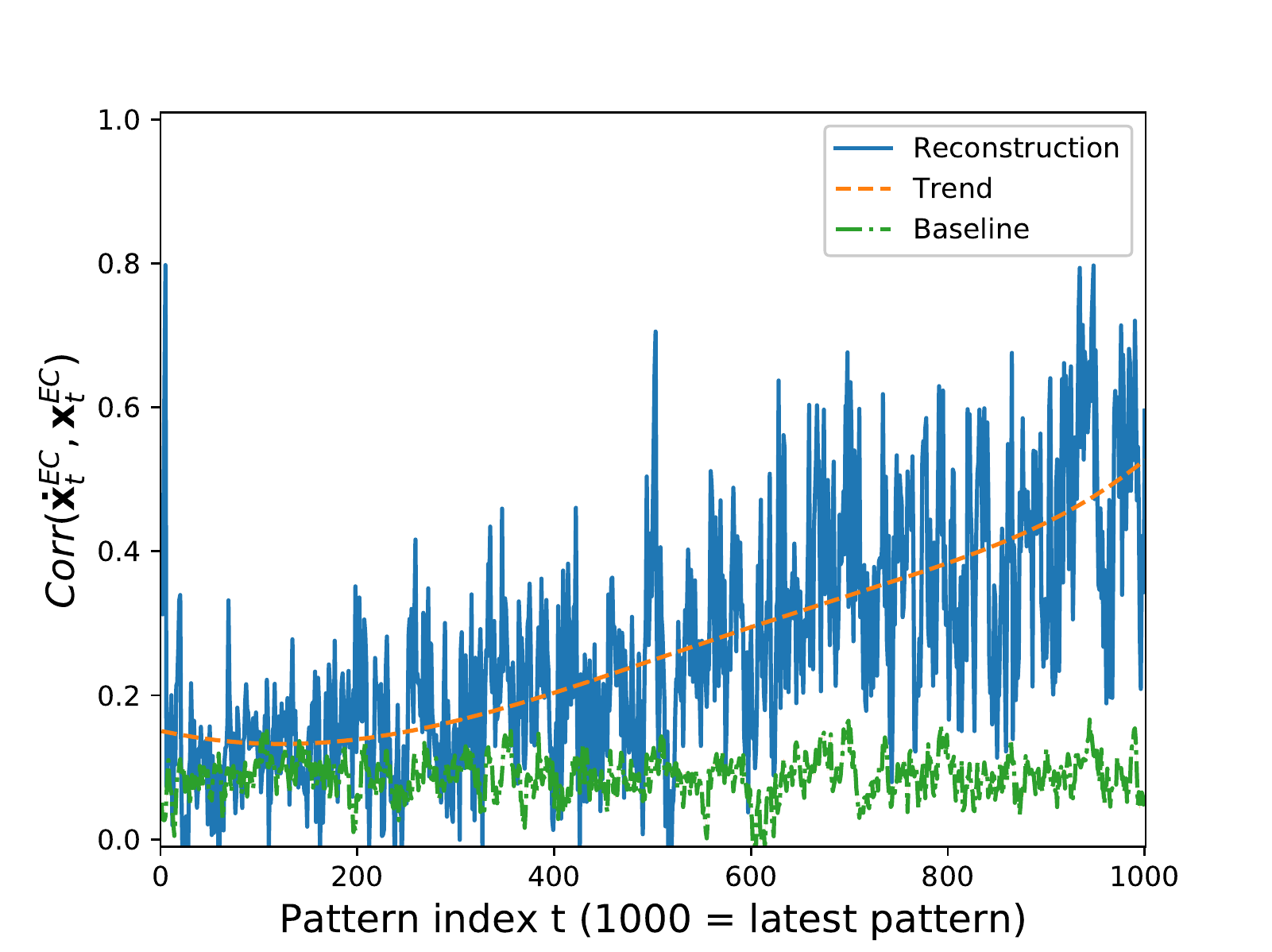}\label{fig:model_A_crosscorr_intrinsic_5_steps}}
\subfigure[Intrinsic recall for 1000 transitions]{
\includegraphics[scale=0.4, trim=10 5 32 40, clip]{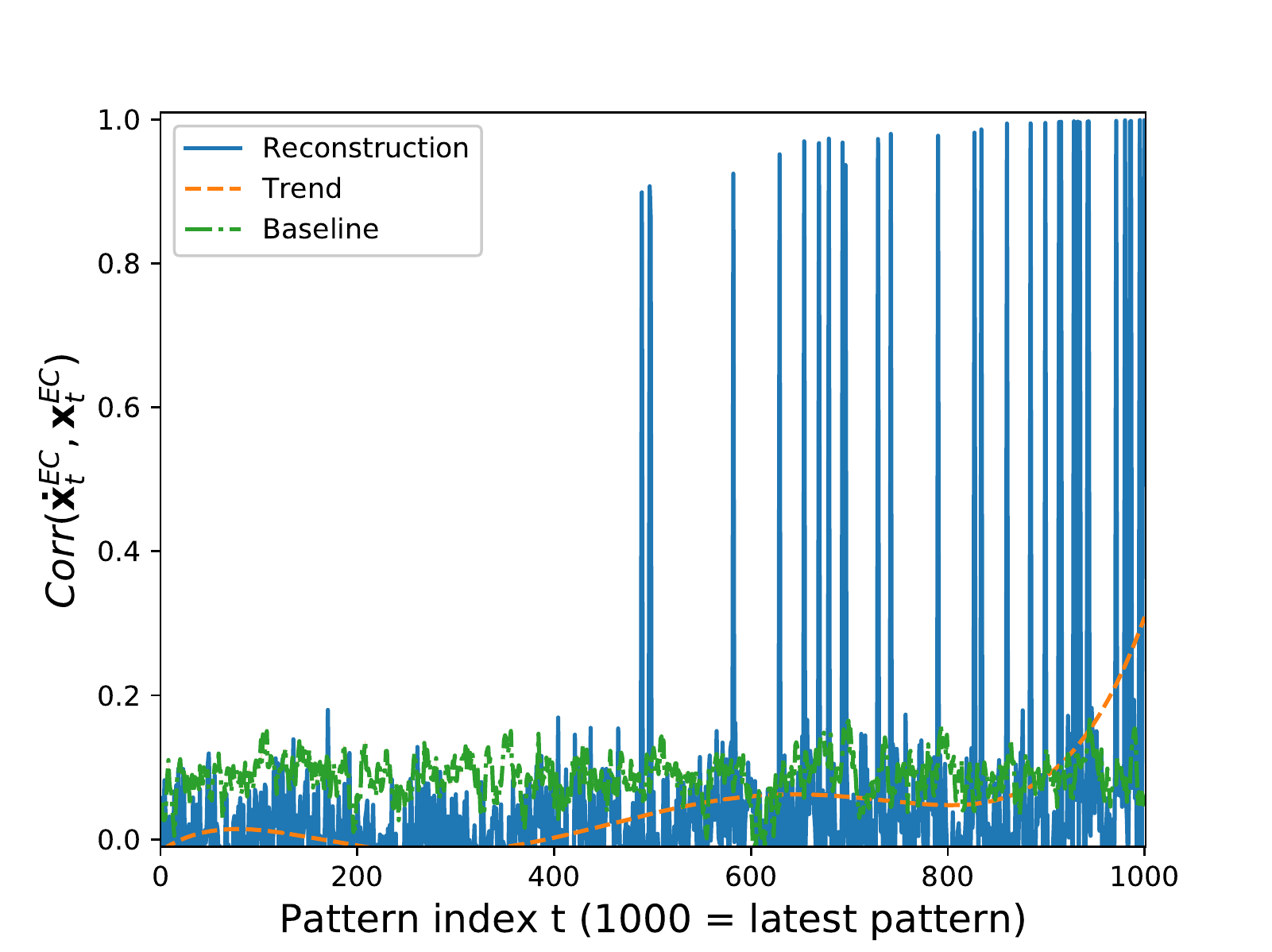}\label{fig:model_A_crosscorr_intrinsic_all}}
\caption{Recall performance of \emph{Model-A} on the \emph{RAND-CORR} dataset. For comparison and detailed description of the subplots see Figure~\ref{fig:model_A_uncorr_all}, which shows the recall performance of \emph{Model-A} on the \emph{RAND} dataset.
}
\label{fig:model_A_crosscorr_all}
\end{center}
\end{figure}

\begin{figure}[t!]
\begin{center}
\subfigure[Encoder + dynamics (1 transition)]{
\includegraphics[scale=0.4, trim=10 5 32 40, clip]{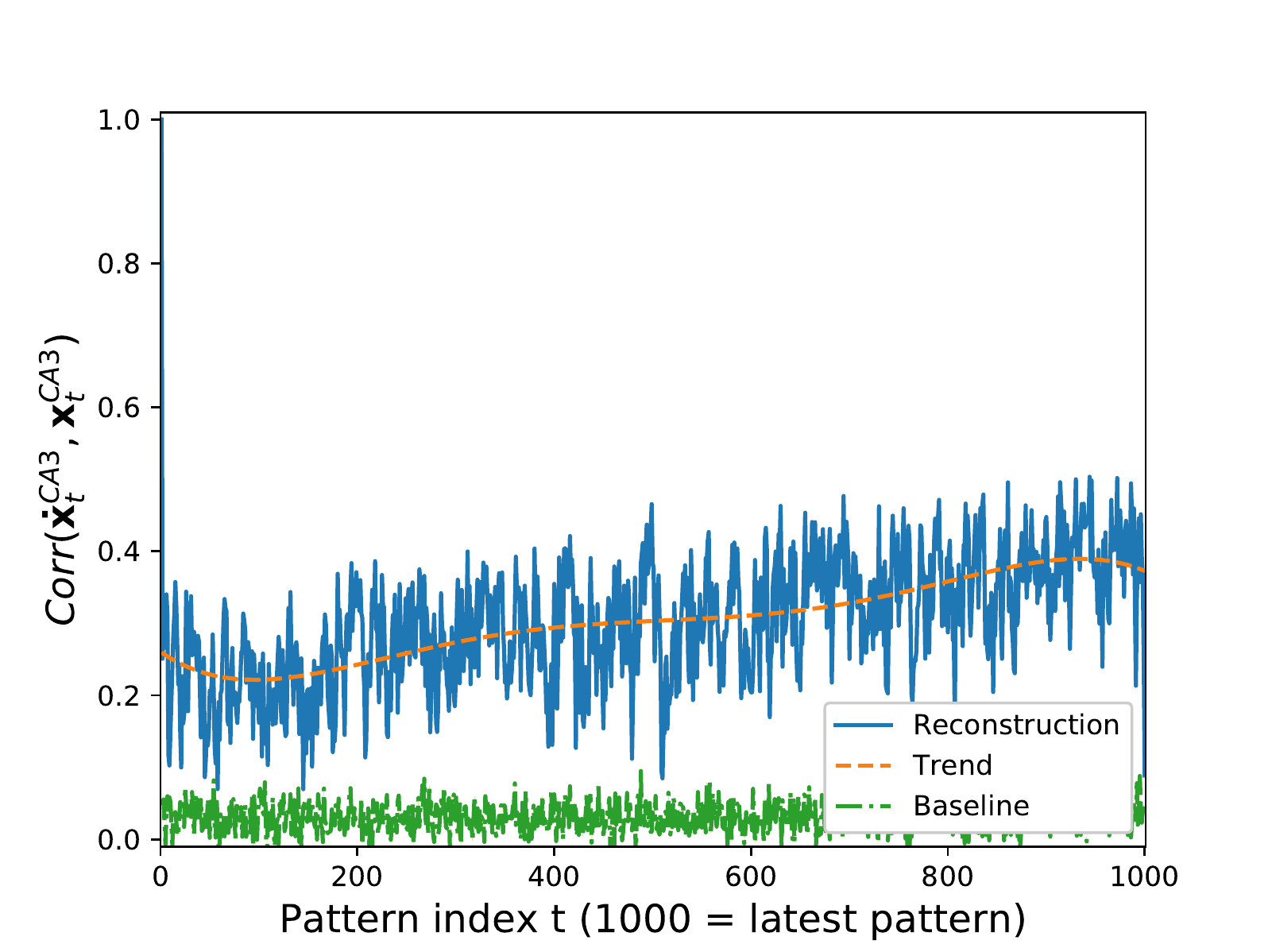}\label{fig:model_A_crosscorr_encoder_dynamic_one_step}}
\subfigure[Encoder + dynamics (5 transitions)]{
\includegraphics[scale=0.4, trim=10 5 32 40, clip]{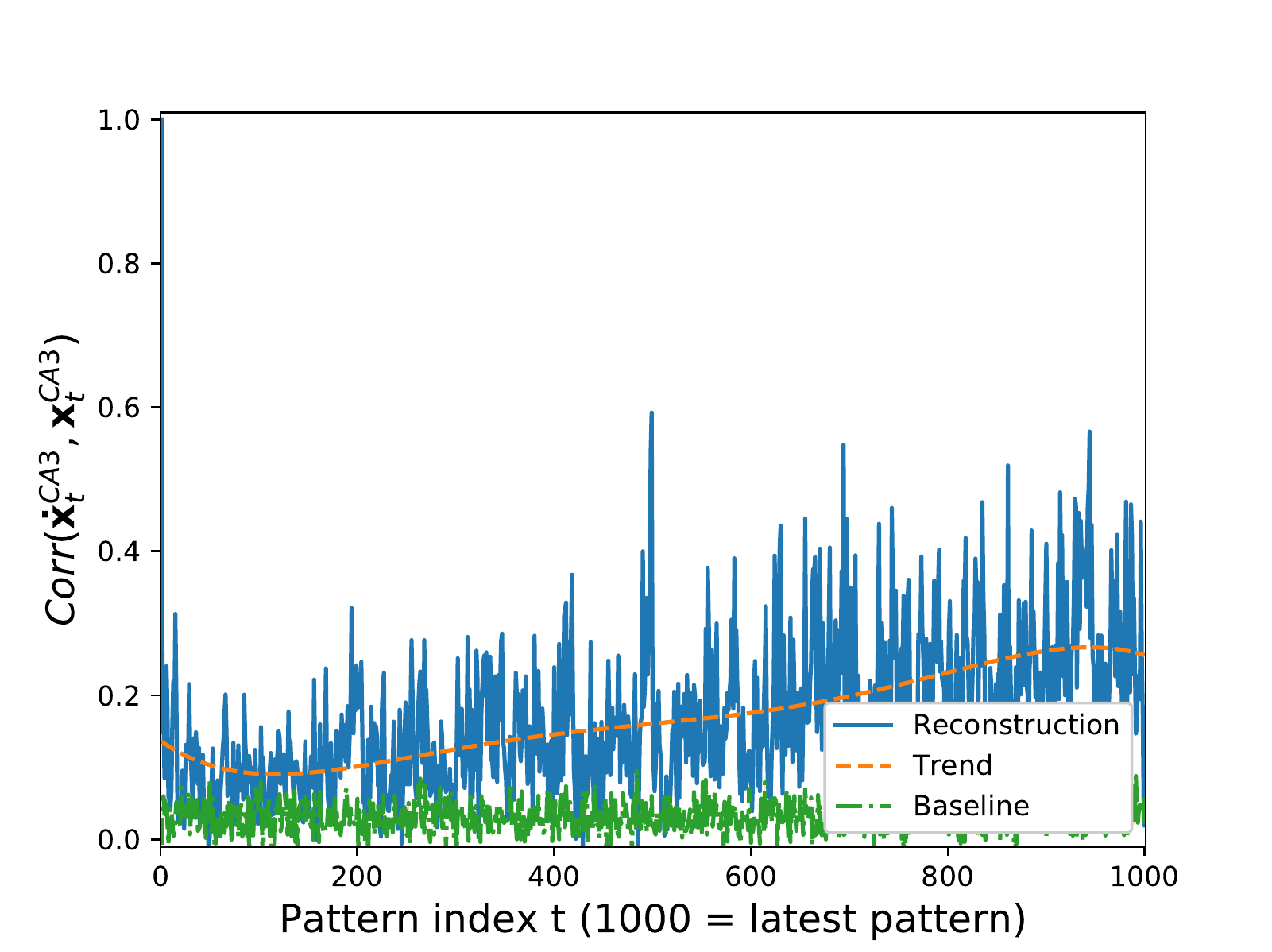}\label{fig:model_A_crosscorr_encoder_dynamic_five_step}}
\caption{Intrinsic performance in CA3 of \emph{Model-A} on the {\bf{\emph{CORR-RAND}}} dataset, where each pattern is encoded, the intrinsic dynamic is iterated (a) once
({\tiny $\vect{\vect{x}}^{EC}(t-1)\rightarrow \dot{\vect{x}}^{CA3}(t-1)\rightarrow \dot{\vect{x}}^{CA3}(t)$})
and (b) five times ({\tiny $\vect{\vect{x}}^{EC}(t-5)\rightarrow \dot{\vect{x}}^{CA3}(t-5)\rightarrow \dot{\vect{x}}^{CA3}(t-4) \cdots \dot{\vect{x}}^{CA3}(t)$}).
Also compare with the performance of \emph{Model-A} on the \emph{RAND} dataset shown in Figure~\ref{fig:model_A_encoder_intrinsic}.
}
\label{fig:model_A_crosscorr_encoder_intrinsic}
\end{center}
\end{figure}

Even though the combination of encoding and decoding still works reasonably well, even for a single intrinsic transition the performance in EC drops already significantly as shown in Figure~\ref{fig:model_A_crosscorr_full_loop_1_step}.
This gets worse as more transitions are performed, which is illustrated for five intrinsic transitions in Figure~\ref{fig:model_A_crosscorr_intrinsic_5_steps} and finally for 1000 intrinsic transitions shown in Figure~\ref{fig:model_A_crosscorr_intrinsic_all}. 
The reason is that the input pattern cannot trigger the correct sequence in CA3. 
This is illustrated in Figure~\ref{fig:model_A_crosscorr_encoder_intrinsic}, which shows the correlation between retrieved and ground truth patterns in CA3, when the EC patterns are encoded and the intrinsic transition is iterated (a) once and (b) five times.
In contrast to the performance in CA3 for uncorrelated data as shown in Figure~\ref{fig:model_A_encoder_intrinsic}, which improves as more intrinsic transitions are performed, the performance for correlated data decreases for almost all patterns as more intrinsic transitions are performed.
As for the outliers in Section~\ref{Intrinsic_Recall_Performance_MODEL_A}, but in this case for almost all patterns, the CA3 dynamics either converges to a shifted version of the intrinsic sequence or to a spurious sequence.
A good performance of the encoder is hence crucial for the performance of the entire system, which should also be able to cope with correlated data.

\subsection{Storing a Sequence of Temporally Correlated Patterns in \emph{Model-B}}\label{sec:storing_temporally_correlated_patterns_in_modelb}

\begin{figure}[htbp!]
\begin{center}
\includegraphics[scale=0.23, trim=0 0 0 50, clip]{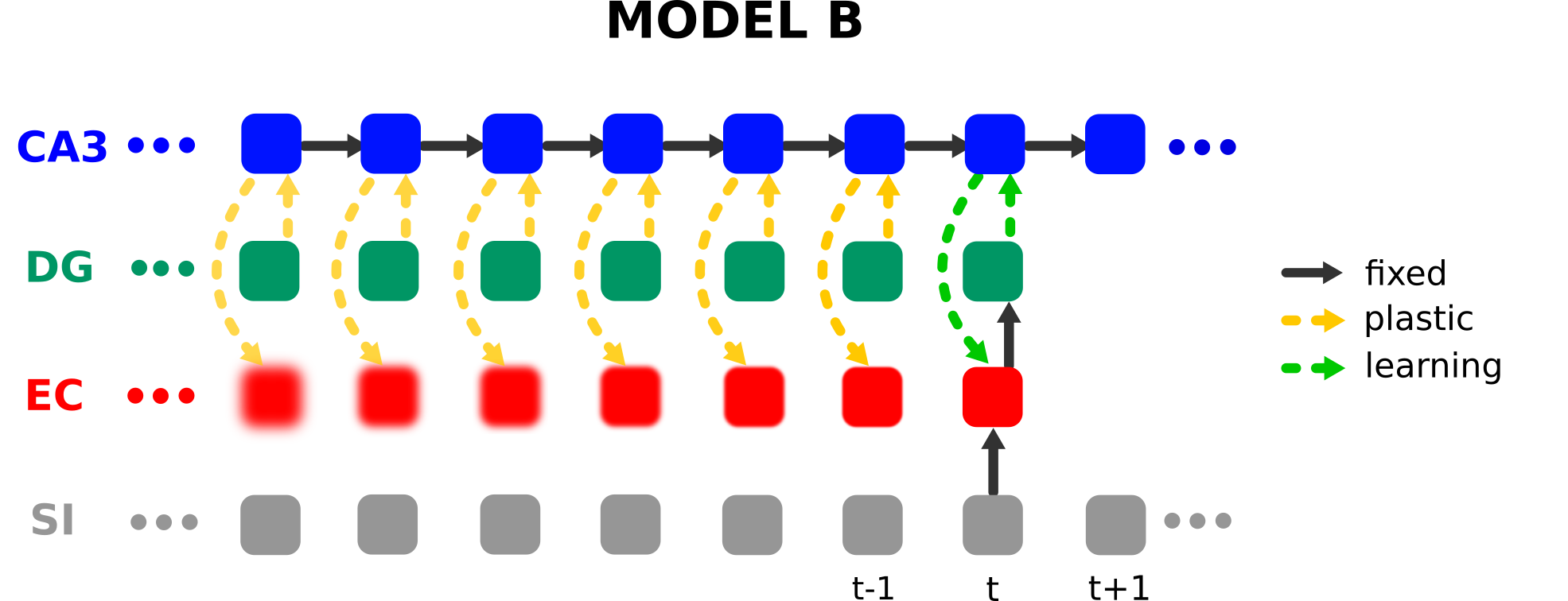}
\caption{
Illustration of hetero-association and forgetting over time in \emph{Model-B}. At time step $t$ the DG pattern (inferred via EC from the SI pattern) is hetero-associated with the CA3, which in turn is hetero-associated with the pattern in EC (indicated by the green dashed arrows). The learned associations weaken over time (indicated by the increasing transparency of the arrows) leading to a degraded reconstruction (forgetting) in EC (indicated by the increasing blurring). However, if you compare this illustration with the illustration for \emph{Model-A} (Figure~\ref{fig:schema_ModelA}) the forgetting indicated by the blurring of the EC patterns is reduced significantly, emphasizing the better performance of \emph{Model-B}. 
}
\label{fig:schema_ModelB}
\vspace{-0.5cm}
\end{center}
\end{figure}

As shown in the previous section, correlations between input patterns are problematic for the encoder, but a good performance of the encoder is crucial for the overall performance of the system as all successive parts of the system depend on the encoder.
To overcome this problem we have added subregion DG to \emph{Model-B}, which serves as a generic way of decorrelating input patterns.
We trained \emph{Model-B} with size $N=1000$ on \emph{RAND-CORR} dataset using the same setup as before.
For clarity Figure~\ref{fig:schema_ModelB} illustrates how the input patterns are hetero-associated with the intrinsic patterns in \emph{Model-B} and how these associations weaken over time. 
This is similar to the illustration for \emph{Model-A} (Figure~\ref{fig:schema_ModelA}) except that here the forward hetero-association is between the pattern in DG and CA3. 

\subsubsection{One-Shot Encoding and Decoding Performance}

\begin{figure}[tbp!]
\begin{center}
\subfigure[Encoder]{
\includegraphics[scale=0.4, trim=10 5 32 40, clip]{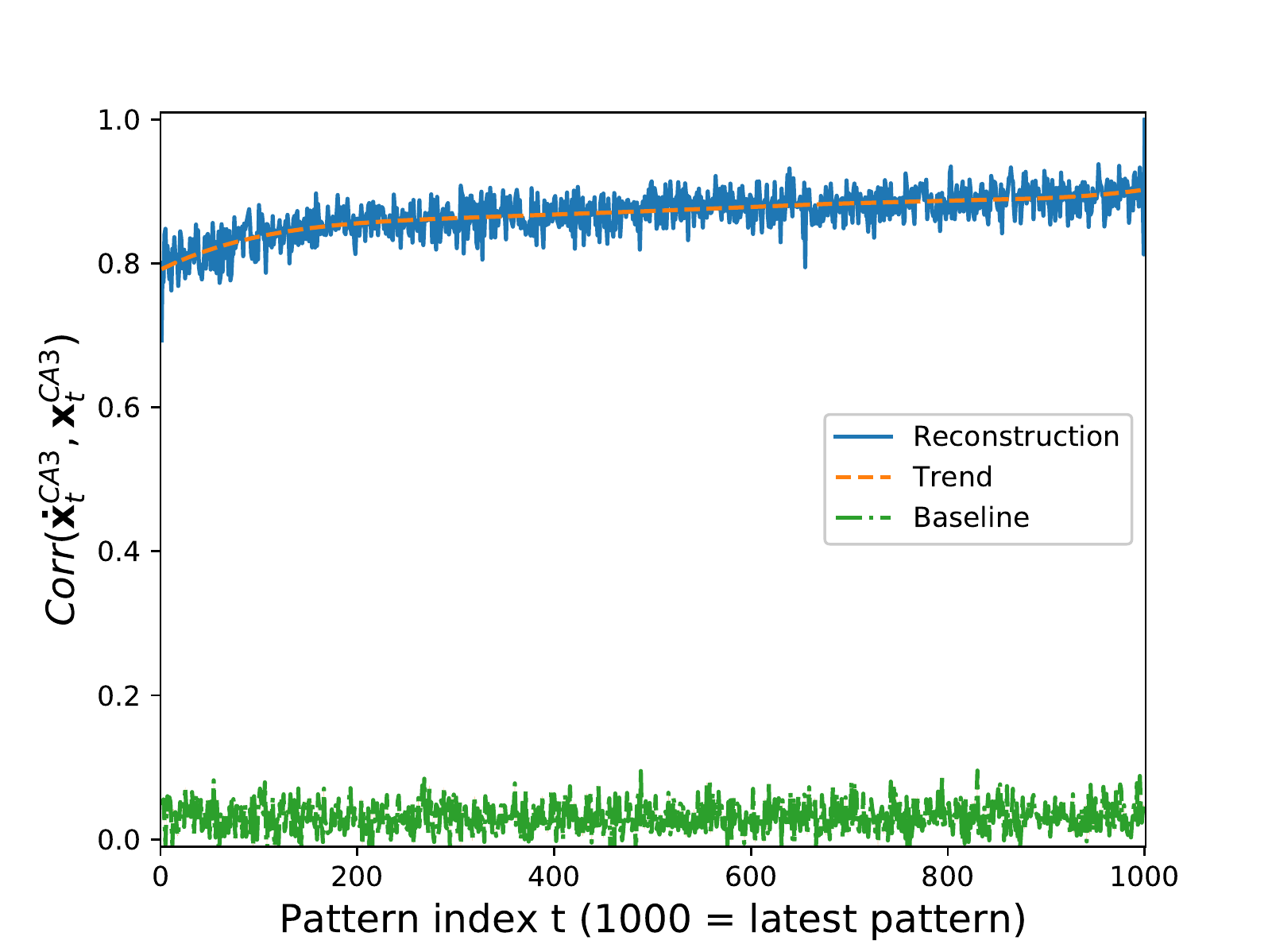}\label{fig:model_B_crosscorr_encoder}}
\subfigure[Encoder + dynamics (1 transition)]{
\includegraphics[scale=0.4, trim=10 5 32 40, clip]{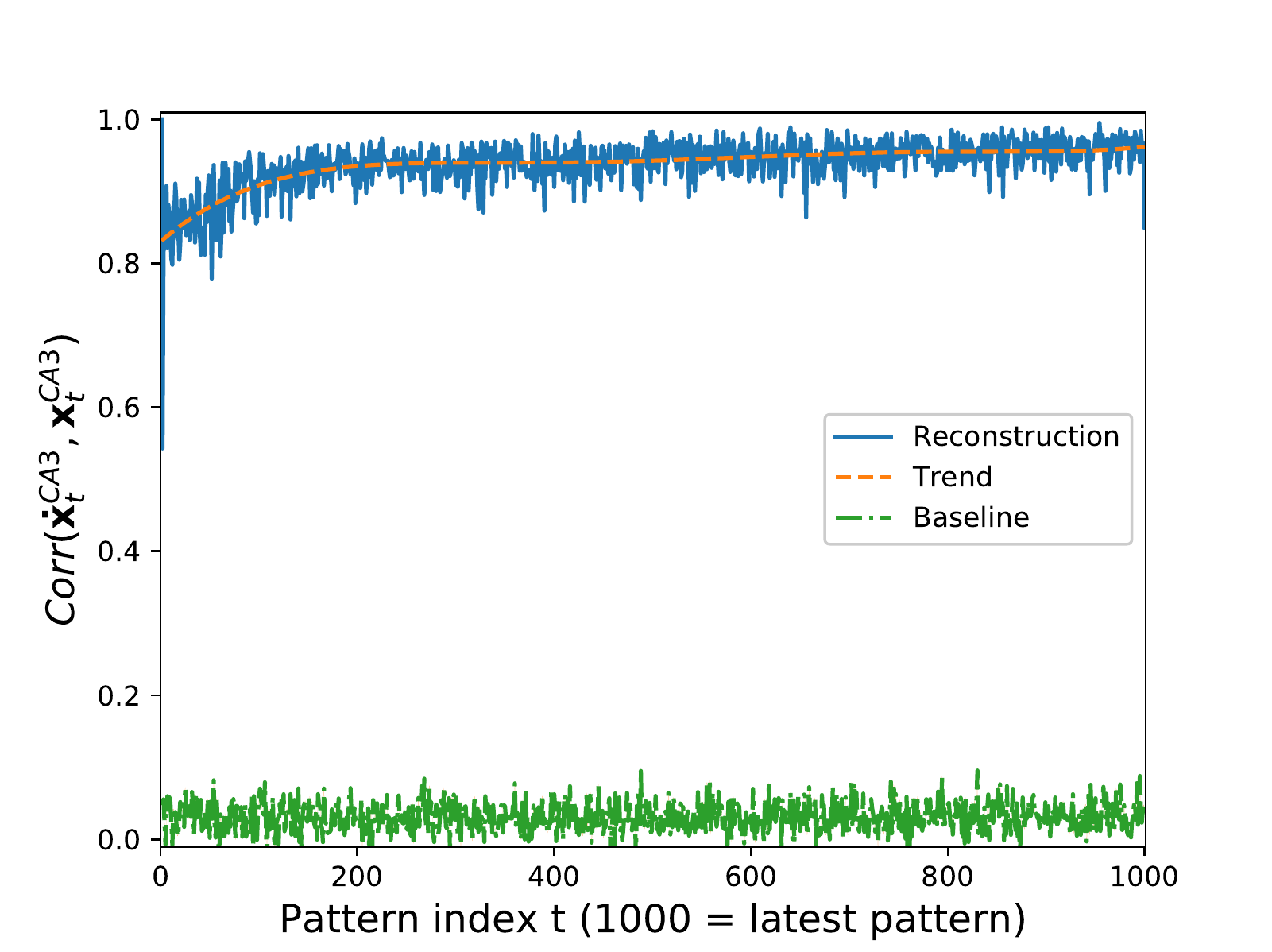}\label{fig:model_B_crosscorr_encoder_dynamic_one_step}}
\caption{Encoding performance of \emph{Model-B} on the \emph{RAND-CORR} dataset for (a) encoder ({\tiny $\dot{\vect{x}}^{EC}(t)\rightarrow \dot{\vect{x}}^{CA3}(t)$}) and (b) encoder and one intrinsic transition ({\tiny $\dot{\vect{x}}^{EC}(t-1)\rightarrow \dot{\vect{x}}^{CA3}(t-1)\rightarrow \dot{\vect{x}}^{CA3}(t)$}). See also Figure~\ref{fig:model_A_crosscorr_encoder_decoder} and Figure~\ref{fig:model_A_uncorr_encoder_decoder} for the encoding performance of \emph{Model-A} on uncorrelated and correlated data.
}
\label{fig:model_B_encoder_decoder_crosscorr}
\end{center}
\end{figure}

\begin{figure}[t!]
\begin{center}
\subfigure[Encoder + decoder]{
\includegraphics[scale=0.4, trim=10 5 32 40, clip]{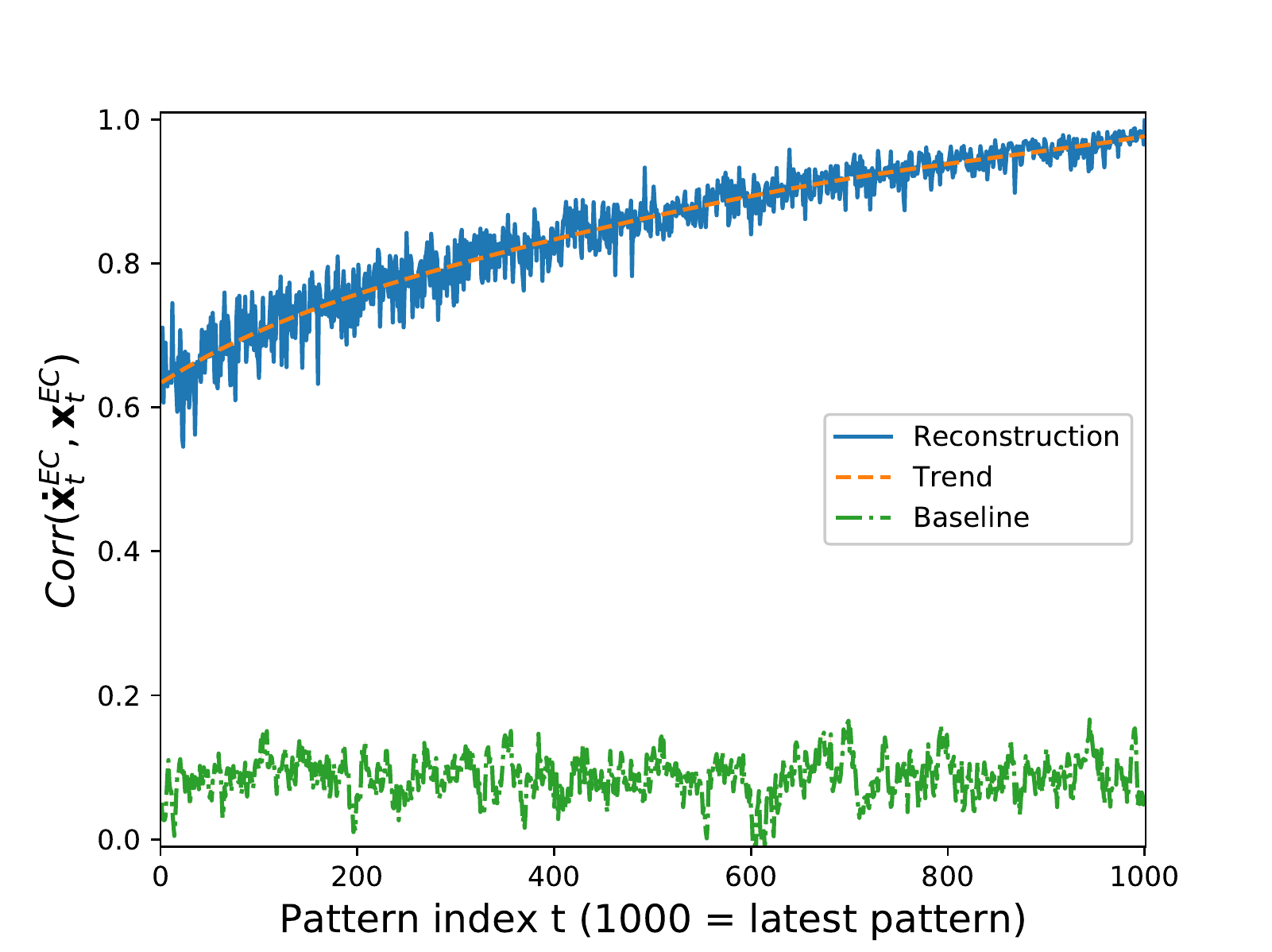}\label{fig:model_B_crosscorr_encode_decode}}
\subfigure[Intrinsic recall for 1 transition]{
\includegraphics[scale=0.4, trim=10 5 32 40, clip]{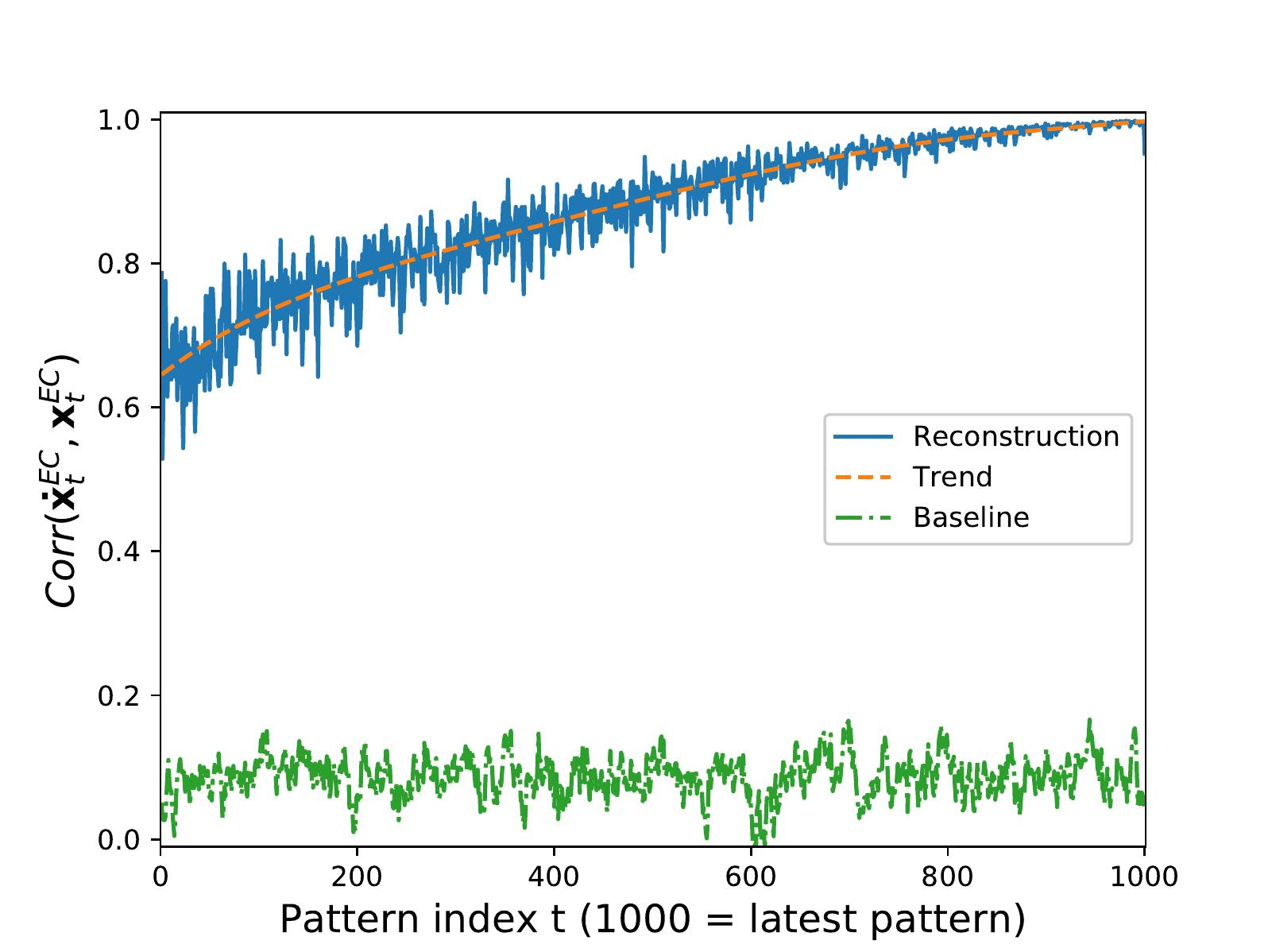}\label{fig:model_B_crosscorr_full_loop_1_step}}
\subfigure[Intrinsic recall for 5 transitions]{
\includegraphics[scale=0.4, trim=10 5 32 40, clip]{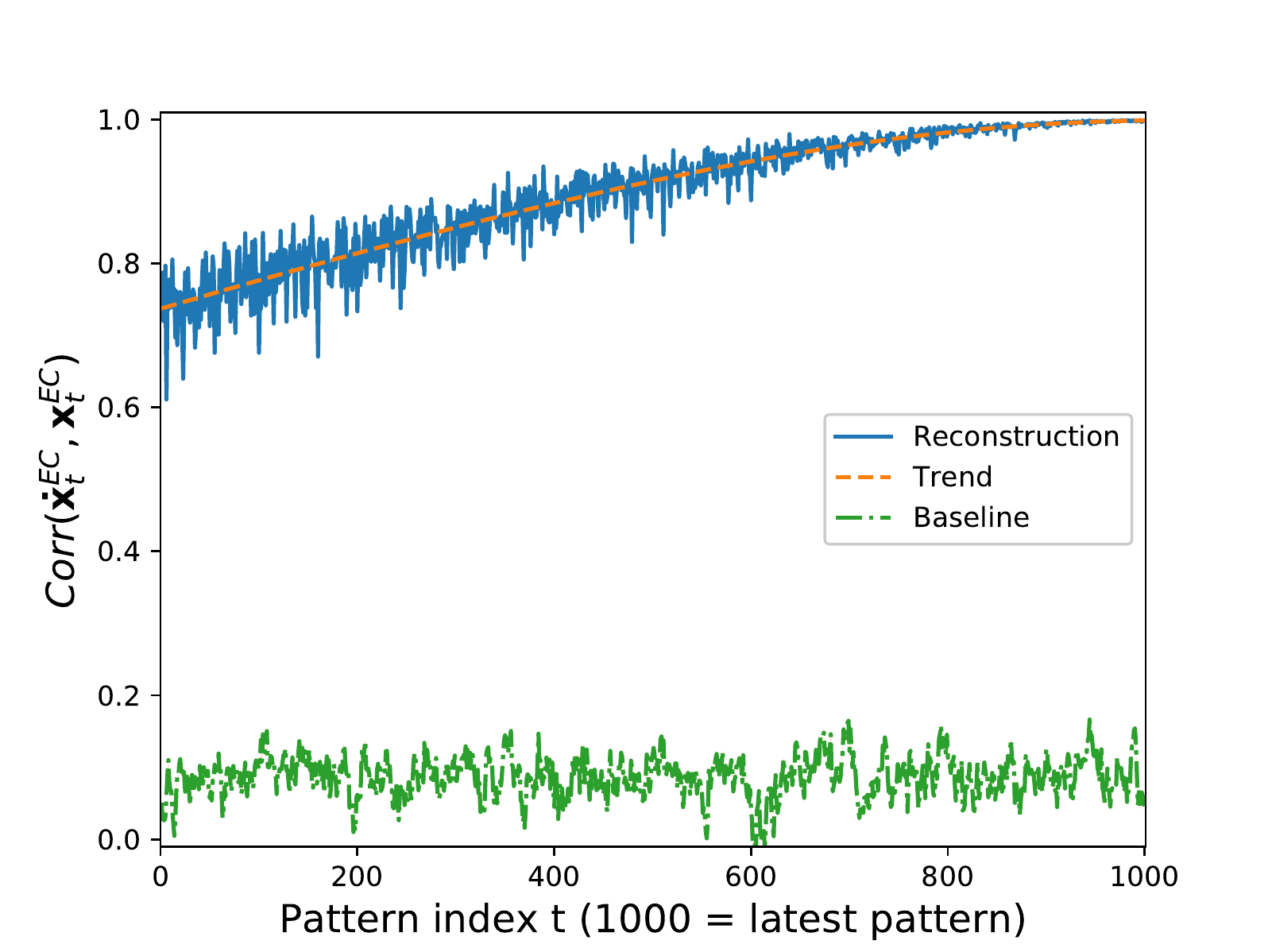}\label{fig:model_B_crosscorr_intrinsic_5_steps}}
\subfigure[Intrinsic recall for 1000 transitions]{
\includegraphics[scale=0.4, trim=10 5 32 40, clip]{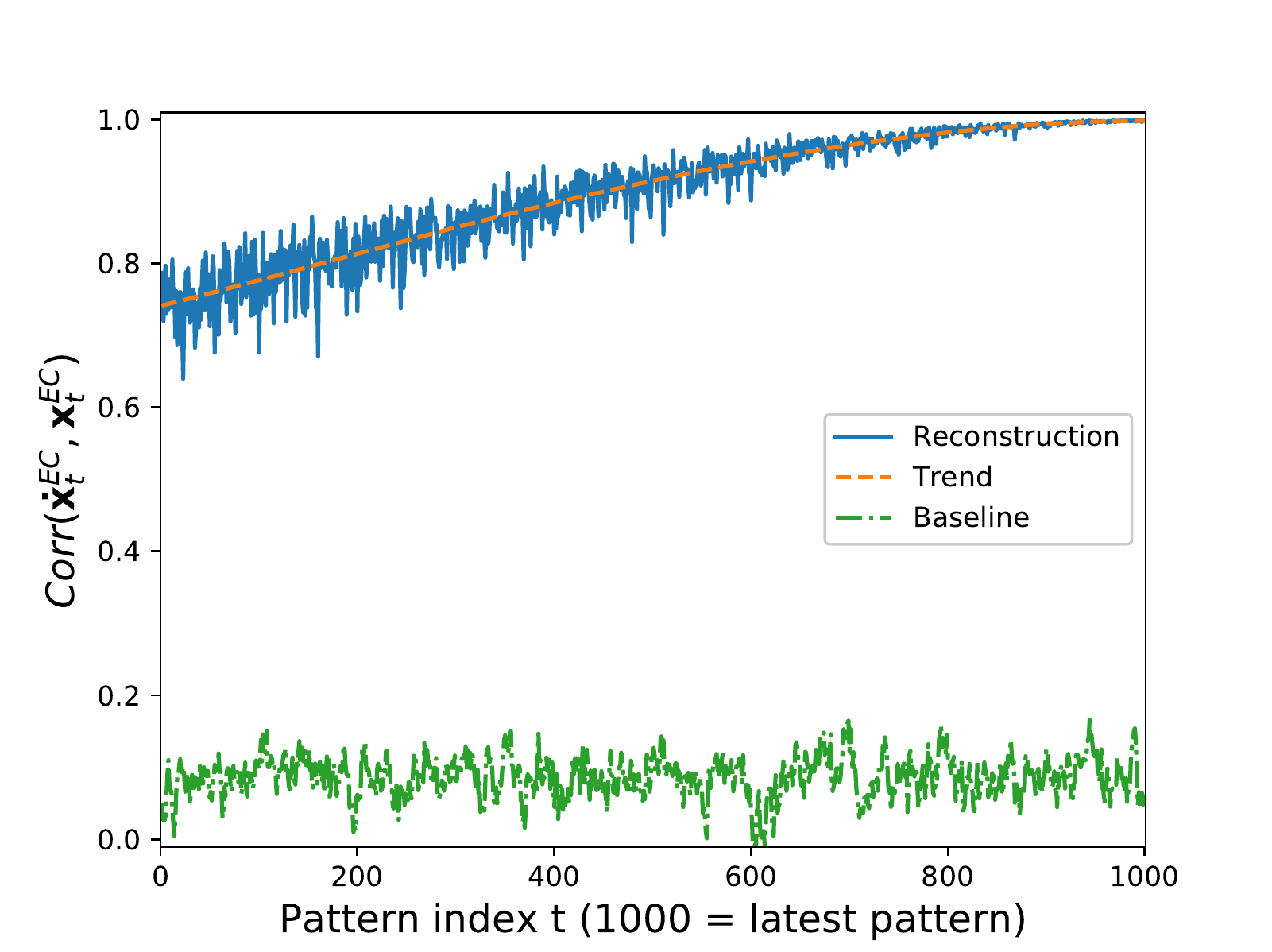}\label{fig:model_B_crosscorr_intrinsic_all}}
\caption{
Recall performance of \emph{Model-B} on the \emph{RAND-CORR} dataset. For comparison and detailed description of the subplots see Figure~\ref{fig:model_A_uncorr_all}, which shows the performance of \emph{Model-A} on the \emph{RAND} dataset. Also compare with the performance of \emph{Model-A} on the \emph{RAND-CORR} dataset shown in Figure~\ref{fig:model_A_crosscorr_all}.
}
\label{fig:model_B_all_crosscorr}
\end{center}
\vspace{-0.4cm}
\end{figure}

By introducing DG the maximum correlation of $0.8$ between two patterns in the dataset is reduced from $0.8$ in EC to $0.45$ in DG. 
Thus it does not fully decorrelate the patterns, but it is sufficient to achieve a good performance of the encoder (EC~$\rightarrow$~DG~$\rightarrow$~CA3), which according to Figure~\ref{fig:model_B_crosscorr_encoder} is $0.87$ on average and more or less equal for all patterns.
The performance of the decoder is not shown, as it is equivalent to the one for \emph{Model-A} shown in Figure~\ref{fig:model_A_crosscorr_decoder}.
Figure~\ref{fig:model_B_crosscorr_encode_decode} shows the performance in EC when patterns are encoded and directly decoded without intrinsic dynamics, which is very close to the performance of the decoder and thus indicates that the errors made by the encoder are compensated by the decoder.

\subsubsection{Recall Performance}

As shown in Figure~\ref{fig:model_B_crosscorr_encoder_dynamic_one_step} the correlation between retrieved and true patterns in CA3 increases to $0.94$ on average after a single intrinsic transition and further increases with the number of intrinsic transitions (data not shown). 
In contrast to \emph{Model-A} the performance in EC does not initially decrease for earlier patterns after one intrinsic transition, which indicates that the encoder provides a much better encoded EC pattern to CA3.
We identified a correlation of approximately 0.8 and above to lead to an increase in performance in EC through the intrinsic dynamics, while a value below leads to an decrease as shown for the uncorrelated data and \emph{Model-A} shown in Figure~\ref{fig:model_A_uncorr_all}.
As already seen for the intrinsic patterns in CA3, the performance in EC improves further with the number of intrinsic transitions when recalling patterns in EC as shown in Figure~\ref{fig:model_B_all_crosscorr} (a\,--\,d).
After five transitions the performance in EC is already very close to the decoder performance shown in Figure~\ref{fig:model_A_crosscorr_decoder}.
We also performed the same experiments with uncorrelated patterns, which as expected performs even better (results not shown).


\subsection{Improving the Performance of \emph{Model-A} on Temporally Correlated Data through Dreaming}\label{sec:experiments_dreaming}

\begin{figure}[htbp!]
\begin{center}
\includegraphics[scale=0.23, trim=0 0 0 50, clip]{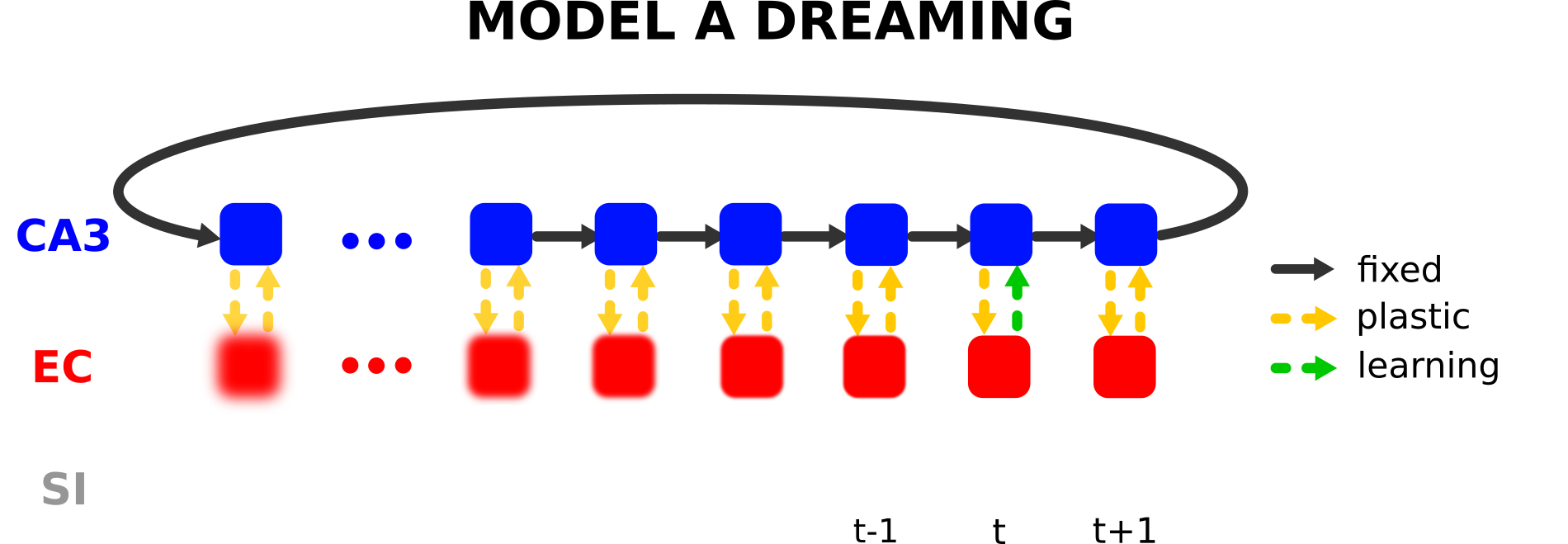}
\caption{
Illustration of the dreaming process in \emph{Model-A}. The intrinsic sequences is looped through several times where at each time step $t$ the EC pattern is reconstructed from the intrinsic pattern via pathway CA3 $\rightarrow$ EC and the forward pathway EC $\rightarrow$ CA3 is updated (indicated by the green dashed arrow) by hetero-associating the reconstructed EC pattern with the corresponding CA3 pattern. Through this process the quality of the EC $\rightarrow$ CA3 pathway increases leading to a better recall performance without any external input from SI. Also compare with the illustration for \emph{Model-A} without dreaming in Figure~\ref{fig:schema_dreaming}.}
\label{fig:schema_dreaming}
\end{center}
\end{figure}

As discussed in the previous sections the encoder in \emph{Model-A} cannot deal with correlated EC patterns, whereas the decoder performs equally well on uncorrelated and correlated EC patterns.
Introducing DG helps to overcome this problem as shown in the previous section, but we can also use the notably property of our model that it can `bootstrap', \emph{i.e.}\ improve the encoder without any external input.
We therefore took the trained model of Section~\ref{sec:storing_temporally_correlated_patterns_in_model_A} and retrained it as described in Section~\ref{sec:retraining_of_plastic_pathways_dreaming}. For clarity the process of dreaming is illustrated in Figure~\ref{fig:schema_dreaming}.

\subsubsection{Performance after Re-training/Dreaming}

\begin{figure}[htbp!]
\begin{center}
\subfigure[Encoder]{
\includegraphics[scale=0.4, trim=10 5 32 40, clip]{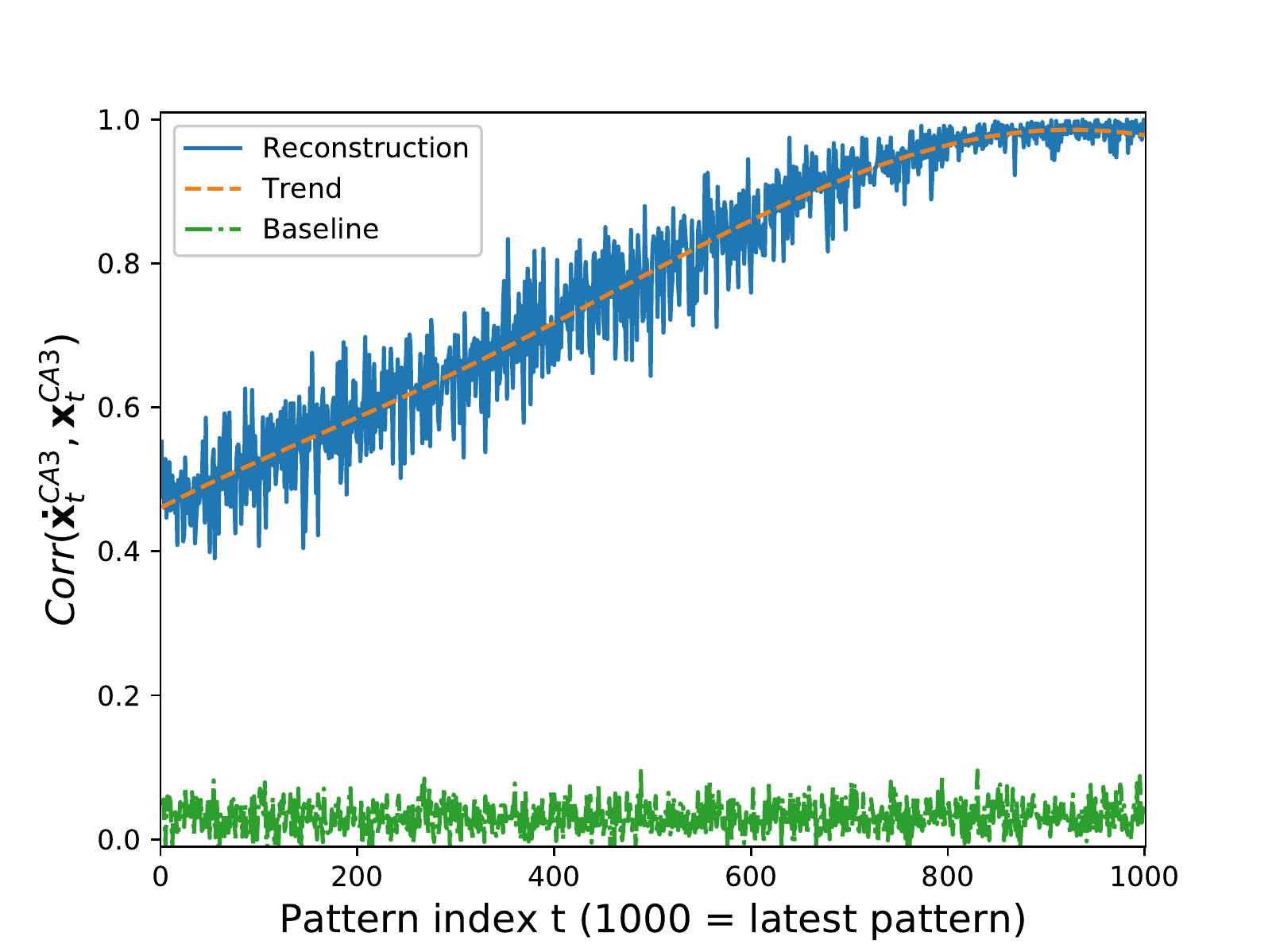}\label{fig:model_A_dreaming_crosscorr_encoder}}
\subfigure[Encoder + dynamics (1 transition)]{
\includegraphics[scale=0.4, trim=10 5 32 40, clip]{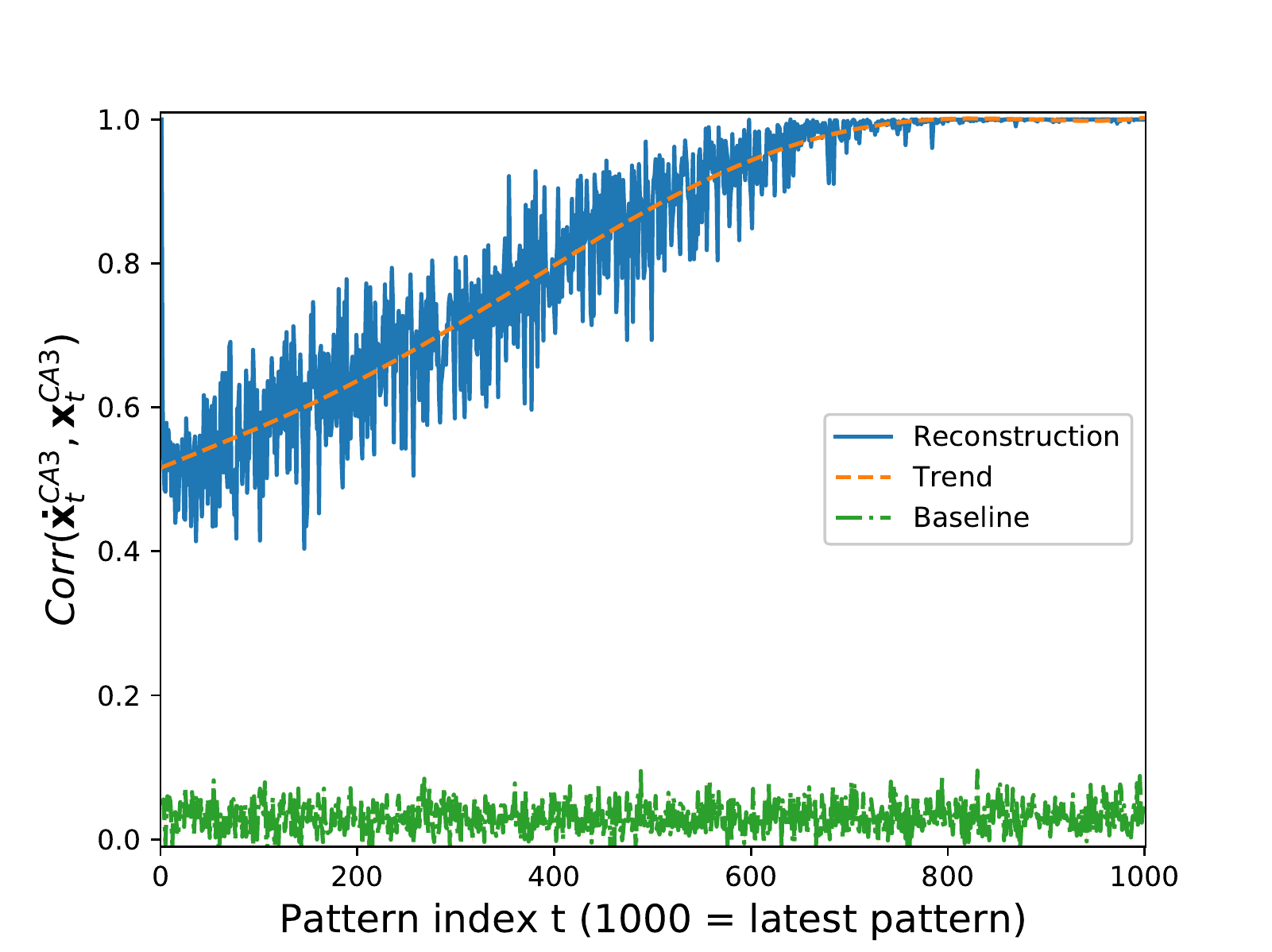}\label{fig:model_A_dreaming_crosscorr_encoder_dynamic_one_step}}
\caption{Encoding performance of \emph{Model-A} on the \emph{RAND-CORR} dataset after retraining/dreaming for (a) encoder ({\tiny $\dot{\vect{x}}^{EC}(t)\rightarrow \dot{\vect{x}}^{CA3}(t)$}) and (b) encoder and one intrinsic transition ({\tiny $\dot{\vect{x}}^{EC}(t-1)\rightarrow \dot{\vect{x}}^{CA3}(t-1)\rightarrow \dot{\vect{x}}^{CA3}(t)$}). See also Figure~\ref{fig:model_B_crosscorr_encode_decode} for the encoding performance of \emph{Model-B} on correlated data.
}
\label{fig:model_A_dreaming_encoder_decoder_crosscorr}
\end{center}
\end{figure}

Figure~\ref{fig:model_A_dreaming_crosscorr_encoder} shows the performance of the encoder after retraining/dreaming, which improves significantly compared to the performance before as shown in Figure~\ref{fig:model_A_crosscorr_encode_decode}.
However, the encoder of \emph{Model-B} has still a better performance as shown in Figure~\ref{fig:model_B_crosscorr_encoder}.  
Figure~\ref{fig:model_A_dreaming_crosscorr_encoder_dynamic_one_step} shows that the correlation between the recalled and ground truth intrinsic patterns improves through the dynamics although this is still much worse than for \emph{Model-B} as shown in Figure~\ref{fig:model_B_crosscorr_encoder_dynamic_one_step}.
As the decoder does not change through retraining/dreaming we refer to Figure~\ref{fig:model_A_crosscorr_decoder} for its performance.

\begin{figure}[t!]
\begin{center}
\subfigure[Encoder + decoder]{
\includegraphics[scale=0.4, trim=10 5 32 40, clip]{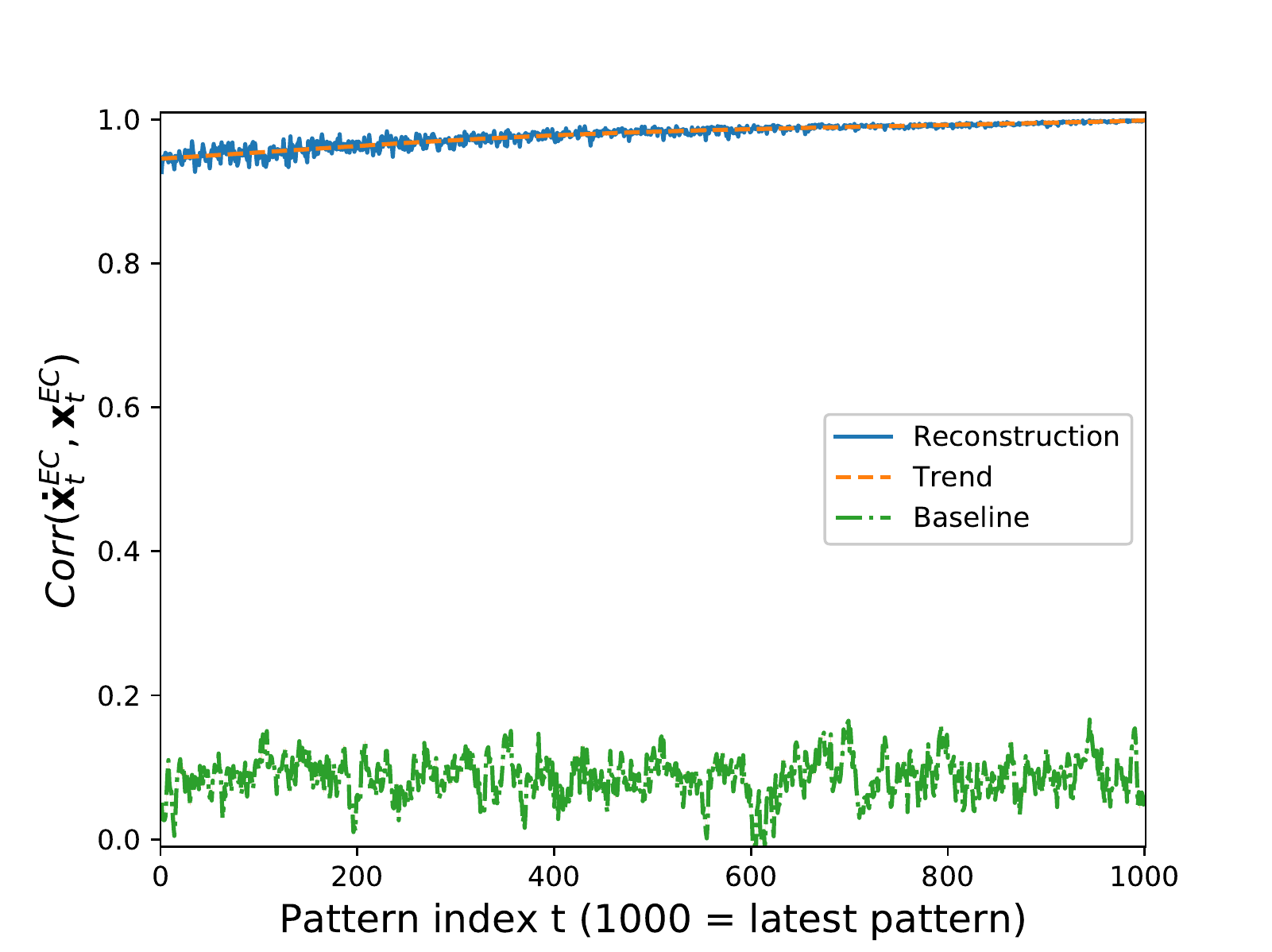}\label{fig:model_A_dreaming_crosscorr_encode_decode}}
\subfigure[Intrinsic recall for 1 transition]{
\includegraphics[scale=0.4, trim=10 5 32 40, clip]{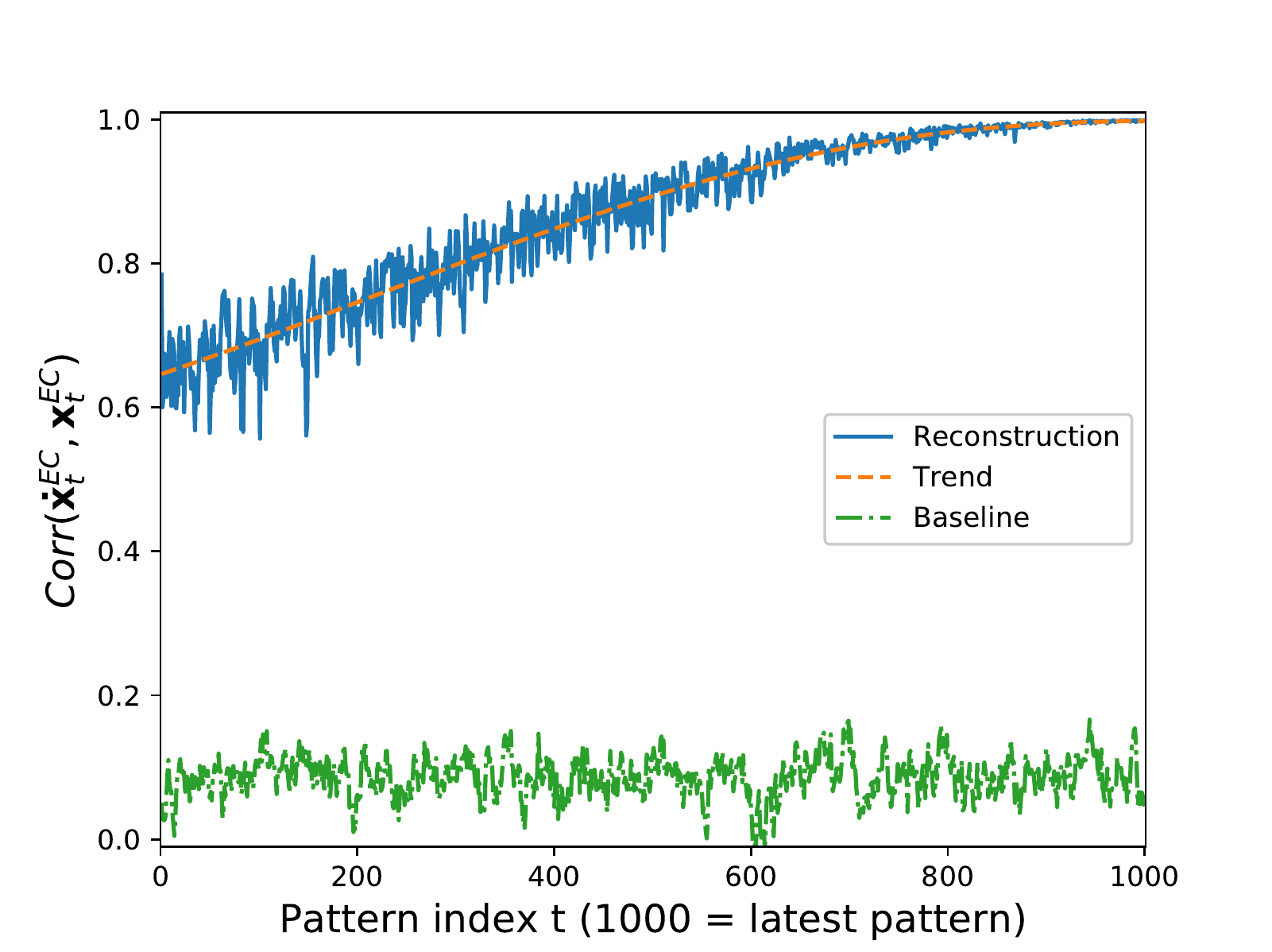}\label{fig:model_A_dreaming_crosscorr_full_loop_1_step}}
\subfigure[Intrinsic recall for 5 transitions]{
\includegraphics[scale=0.4, trim=10 5 32 40, clip]{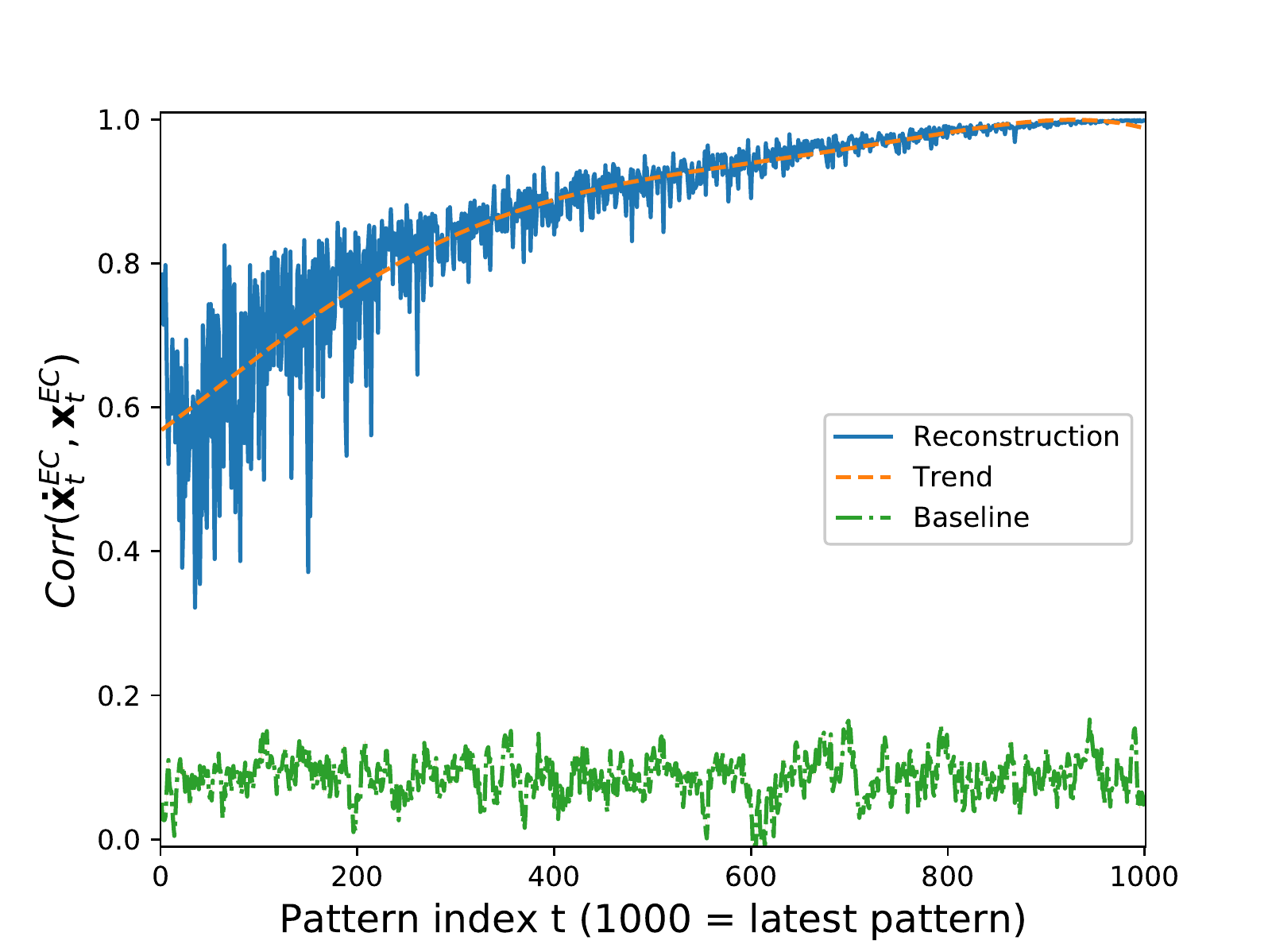}\label{fig:model_A_dreaming_crosscorr_intrinsic_5_steps}}
\subfigure[Intrinsic recall for 1000 transitions]{
\includegraphics[scale=0.4, trim=10 5 32 40, clip]{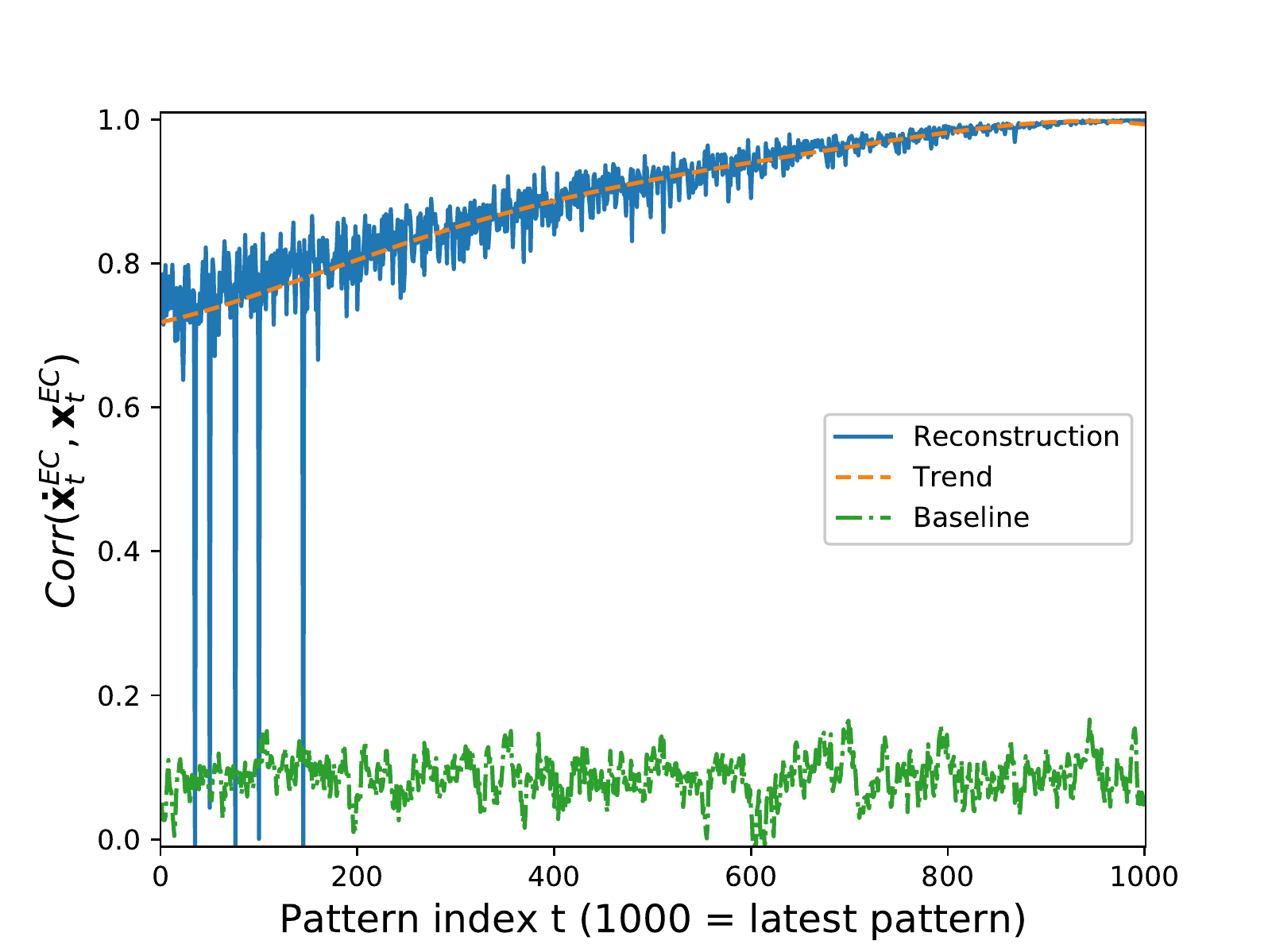}\label{fig:model_A_dreaming_crosscorr_intrinsic_all}}
\caption{
Recall performance of \emph{Model-A} on the \emph{RAND-CORR} dataset after retraining/dreaming. For comparison and detailed description of the subplots see Figure~\ref{fig:model_A_uncorr_all}, which shows the performance of \emph{Model-A} on the \emph{RAND} dataset. Also compare with the performance of \emph{Model-B} on the \emph{RAND-CORR} dataset shown in Figure~\ref{fig:model_A_crosscorr_all}.
}
\label{fig:model_A_dreaming_all_crosscorr}
\end{center}
\end{figure}

Figure~\ref{fig:model_A_dreaming_crosscorr_encode_decode} shows the performance in EC after dreaming when the patterns are encoded and directly decoded, which is remarkably good. 
On first sight this seems to be contradictory as the individual performance of encoder and decoder is much worse. 
It becomes plausible, however, when realizing that we retrain the encoder to become the inverse transformation of the decoder. 
Let us recall that the decoder should learn the transformation $\vect{x}^{CA3}\rightarrow \vect{x}^{EC}$, but as online learning is imperfect the decoder learned the transformation $\vect{x}^{CA3}\rightarrow \dot{\vect{x}}^{EC}$ instead. 
Similarly, the encoder, although trained on the exact patterns, has only learned the imperfect transformation $\vect{x}^{EC} \rightarrow \dot{\vect{x}}^{CA3}$. 
By retraining the encoder on pattern pairs $\dot{\vect{x}}^{EC} $ and $ \vect{x}^{CA3}$, this transformation becomes $\dot{\vect{x}}^{EC} \rightarrow \vect{x}^{CA3}$, which is the inverse transformation of the decoder, so that it is not surprising that reconstruction without intrinsic dynamics is almost perfect after retraining/dreaming.
 
Even though the encoder is retrained on imperfect inputs $\dot{\vect{x}}^{EC}$, they seem to be good enough to improve the encoder, and the advantage of multiple training epochs seems to outweigh the disadvantage of using imperfect input patterns. This helps to improve the intrinsic recall as shown in Figure~\ref{fig:model_A_dreaming_all_crosscorr} for one, five, and 1000 transitions.

\subsection{Storing a Sequence of Handwritten Digits with \emph{Model-B}}\label{sec:exp_hippo_mnist}

Artificial data is helpful when analyzing a model as we have full control of the input statistics.
However, what we are really interested in is storing a sequence of meaningful real world data.
We therefore trained \emph{Model-B} with $N=200$ and $N=1000$ on the \emph{MNIST} dataset using the same setup as used in the previous experiments. 

\subsubsection{Comparison of Small and Large Models}

\begin{figure}[t!]
\begin{center}
\subfigure[Encoder]{
\includegraphics[scale=0.4, trim=10 5 32 40, clip]{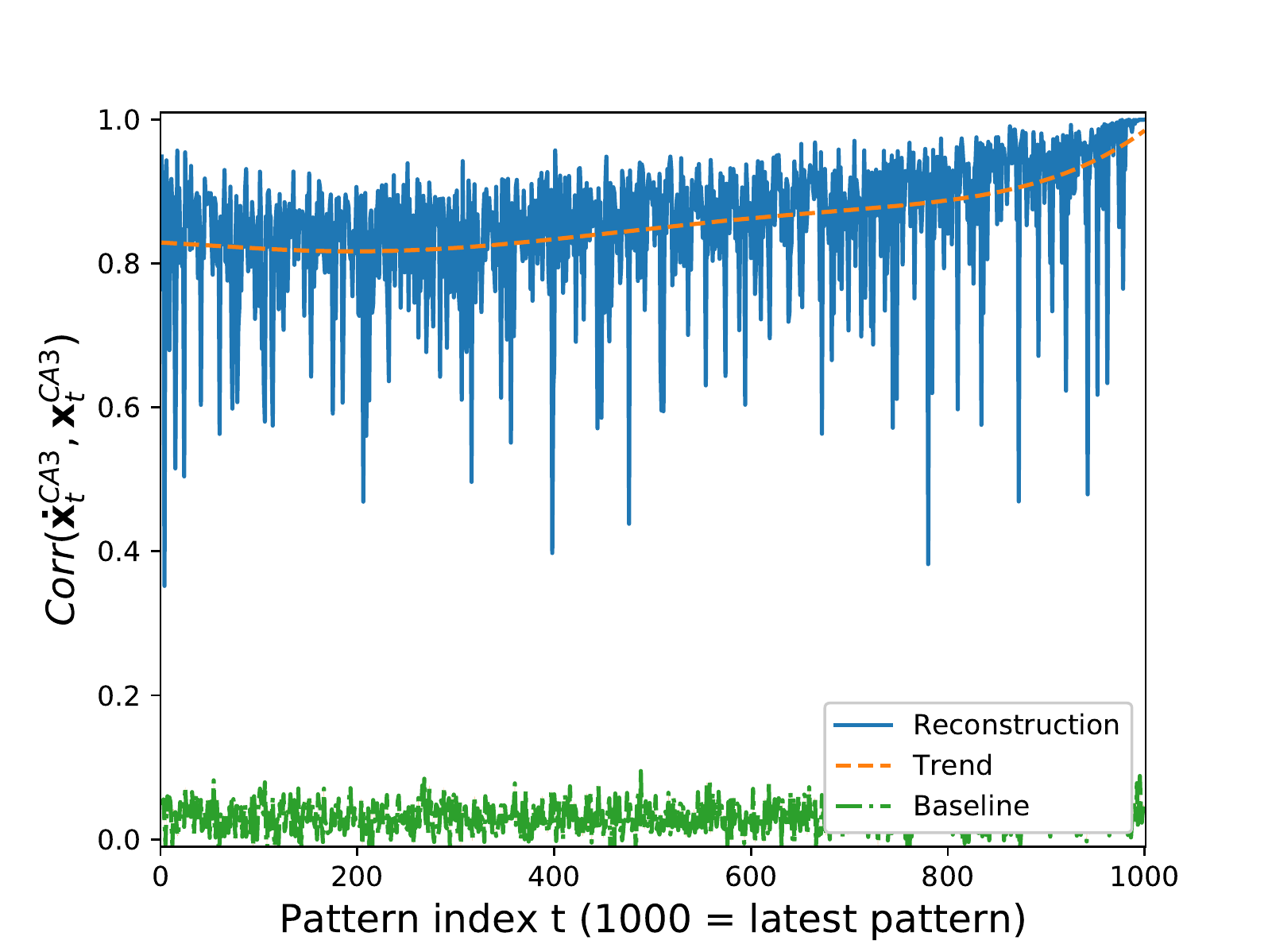}\label{fig:model_B_1000_mnist_encoder}}
\subfigure[Intrinsic recall for 1000 transitions]{
\includegraphics[scale=0.4, trim=10 5 32 40, clip]{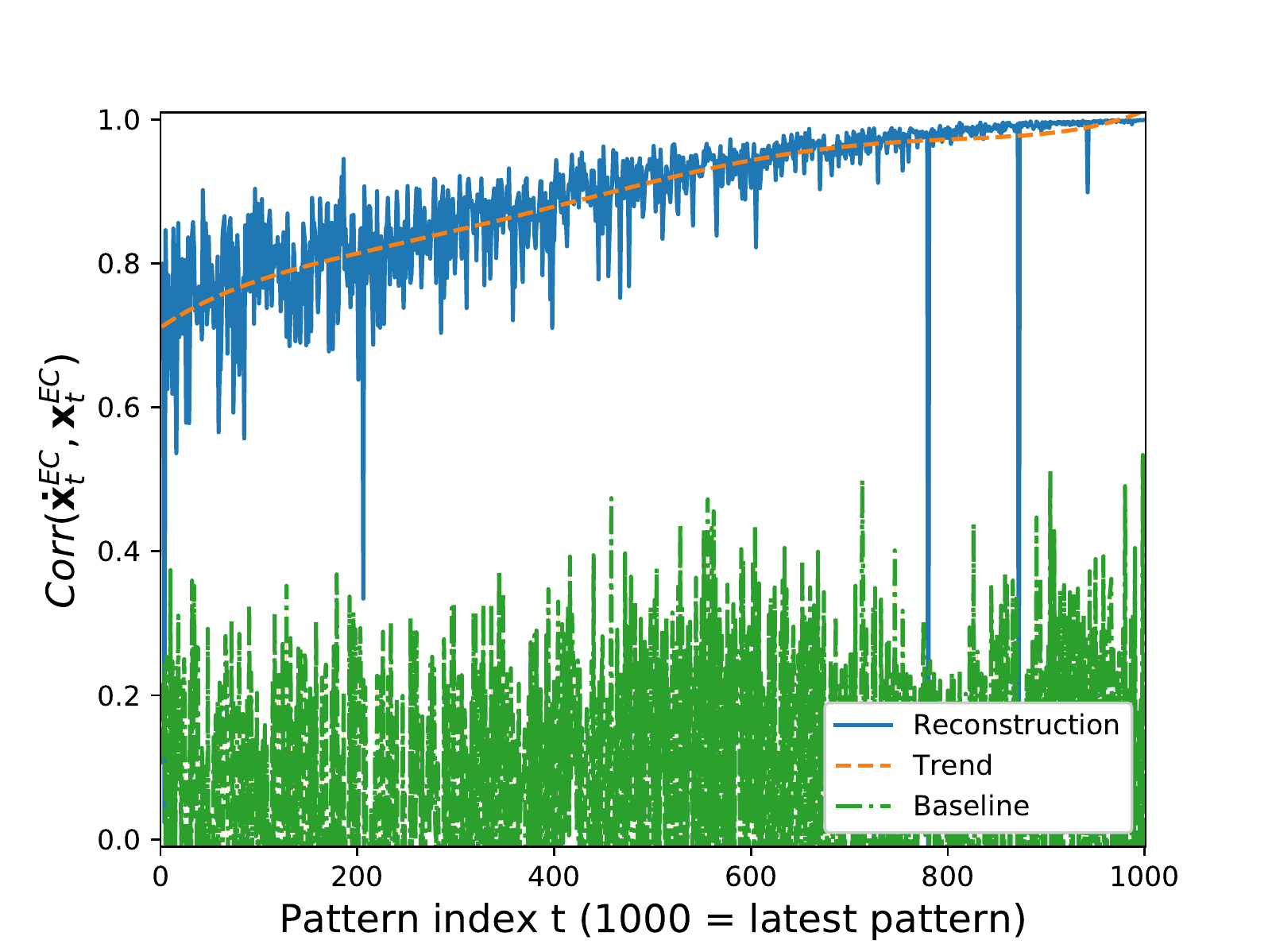}\label{fig:model_B_1000_mnist_intrinsic_all}}
\caption{Performance of \emph{Model-B} on the \emph{MNIST} dataset for $N=1000$ for (a) the encoder, and (b) the full intrinsic recall. 
Also compare with the performance of \emph{Model-B} on the \emph{RAND-CORR} dataset shown in Figure~\ref{fig:model_B_crosscorr_encoder} and Figure~\ref{fig:model_B_crosscorr_intrinsic_all}.
}
\label{fig:model_B_1000_MNIST_encoder_intrinsic_all}
\end{center}
\end{figure}

\begin{figure}[t!]
\begin{center}
\subfigure[Encoder]{
\includegraphics[scale=0.4, trim=10 5 32 40, clip]{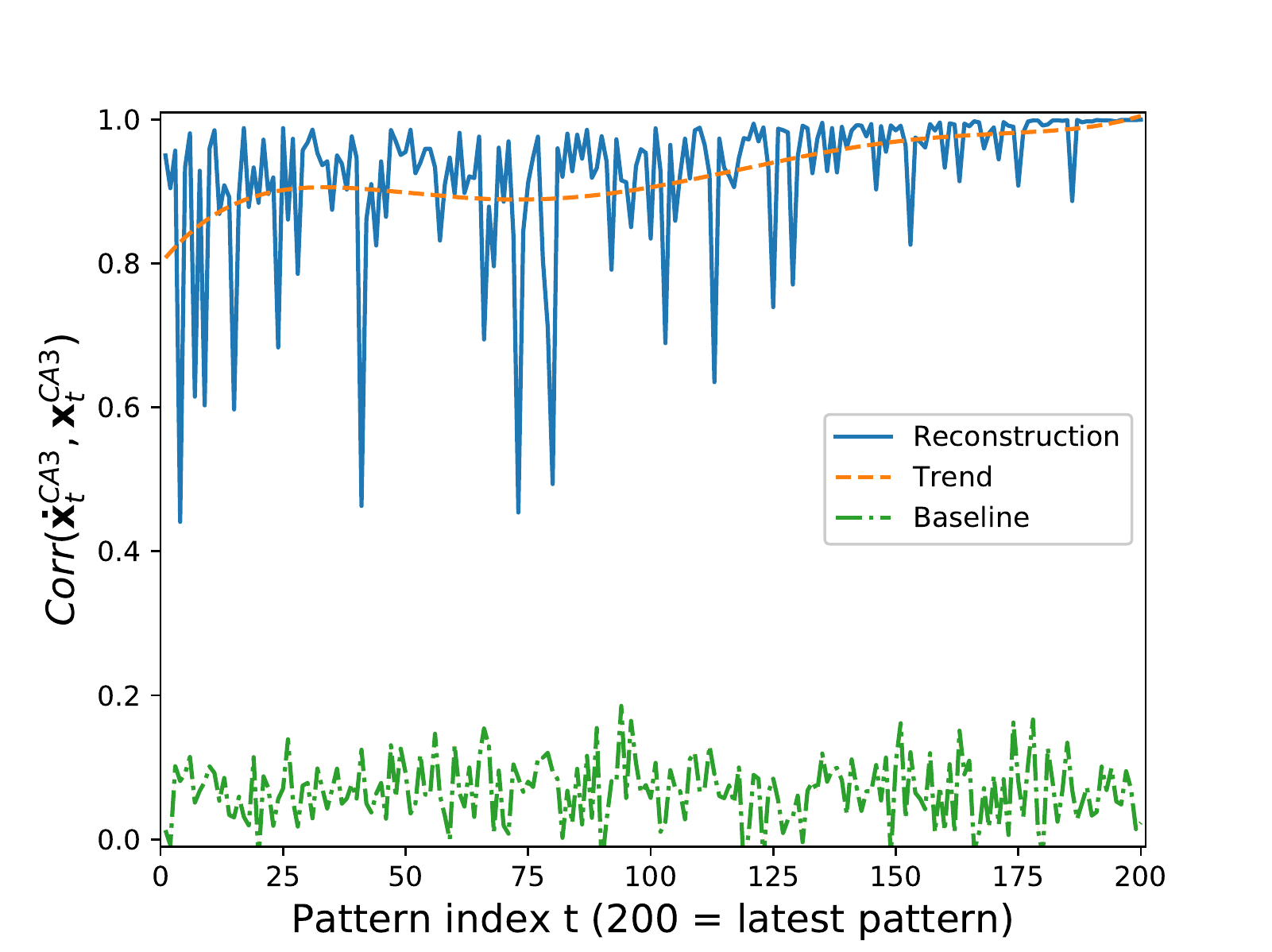}\label{fig:model_B_200_mnist_encoder}}
\subfigure[Intrinsic recall for 200 transitions]{
\includegraphics[scale=0.4, trim=10 5 32 40, clip]{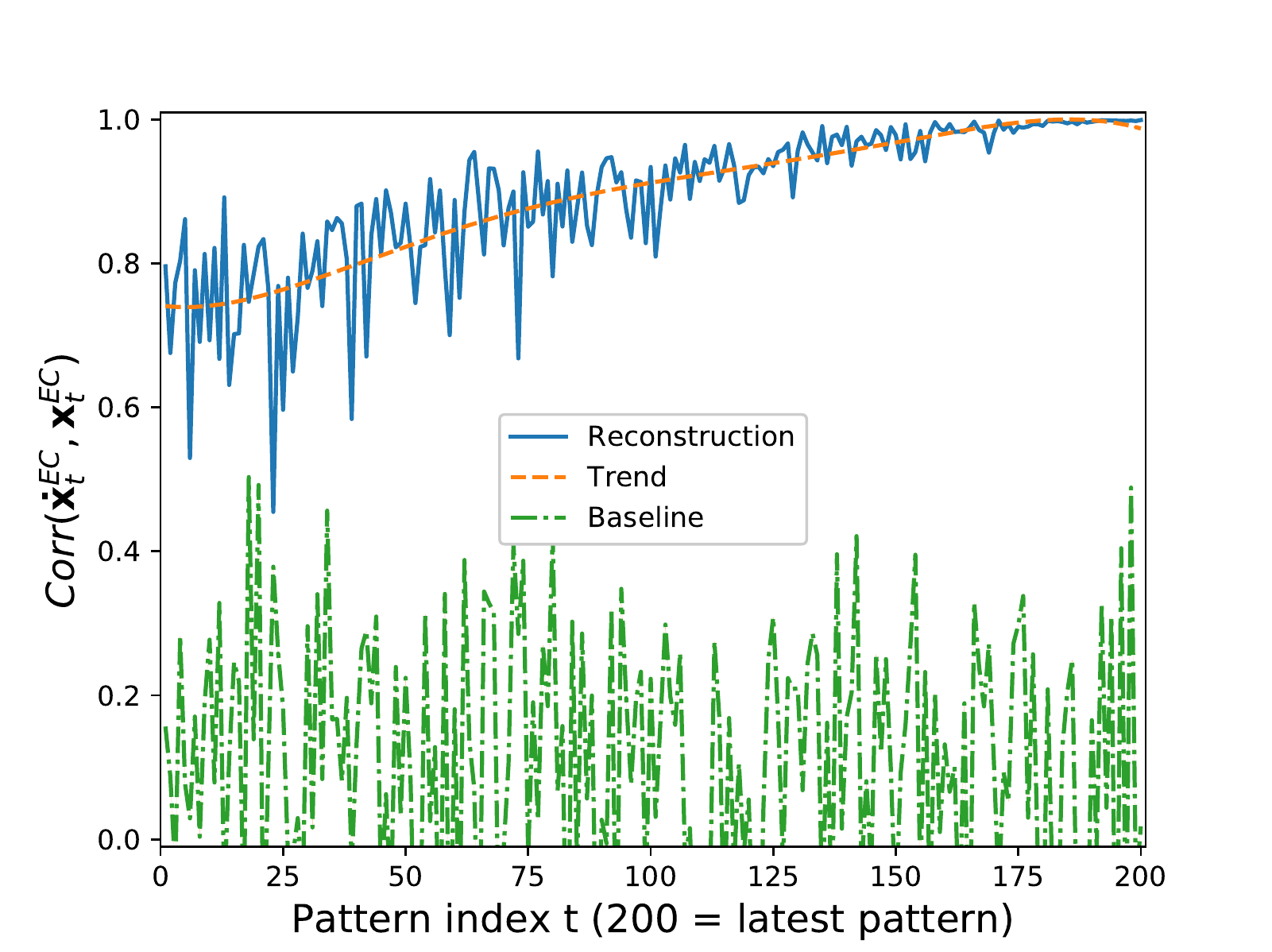}\label{fig:model_B_200_mnist_intrinsic_all}}
\caption{Performance of \emph{Model-B} on the \emph{MNIST} for $N=200$ for (a) the encoder, and (b) the full intrinsic recall. 
Also compare with the performance of \emph{Model-B} on the \emph{RAND-CORR} dataset shown in Figure~\ref{fig:model_B_crosscorr_encoder} and Figure~\ref{fig:model_B_crosscorr_intrinsic_all}.
}
\label{fig:model_B_200_MNIST_encoder_intrinsic_all}
\end{center}
\end{figure}

Figures~\ref{fig:model_B_1000_mnist_encoder} and (b) show the performance of the encoder and full intrinsic recall on \emph{MNIST}, respectively,
which apart from a higher variance are comparable to the performance on the \emph{RAND-CORR} dataset shown in Figure~\ref{fig:model_B_crosscorr_encoder}~and Figure~\ref{fig:model_B_crosscorr_intrinsic_all}.
We also trained a model that is scaled down by a factor of five leading to $N=200$ and thus to a sequence length of 200. 
The model is trained with the exact same setup as before and the performance of the encoder as well as that of the full intrinsic recall is shown in Figure~\ref{fig:model_B_200_MNIST_encoder_intrinsic_all}.
As a sequence length of 1000 is impractical for visualization purposes we focus in the following on the smaller model with $N=200$.
We have verified that all results of the smaller and larger model are qualitatively the same (results not shown), illustrating how nicely the model scales with $N$ without changing the training setup at all.
Furthermore, the decoder performance is not shown, since it is similar to that of the full intrinsic recall.

\subsubsection{Visualization of Input, Reconstructed, and Recalled Patterns}

\begin{figure}[htbp!]
\begin{center}
\subfigure[Input data]{\hspace{-2pt}
\includegraphics[scale=0.471, trim=0 0 0 0, clip]{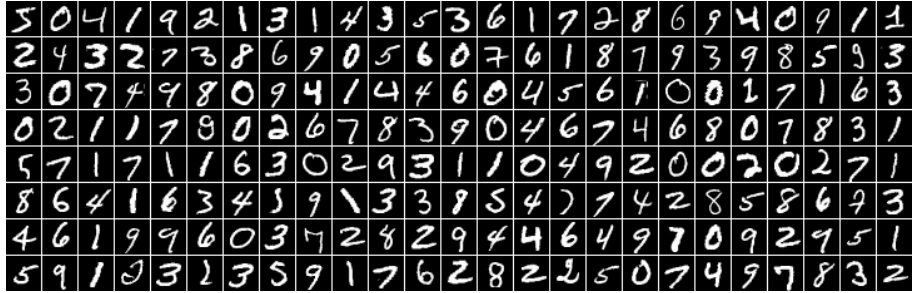}\label{fig:mnist_200_data}
}
\subfigure[Visualized ground truth (via SI~$\rightarrow$~EC~$\rightarrow$~SI)]{\hspace{-2pt}
\includegraphics[scale=0.471, trim=0 0 0 0, clip]{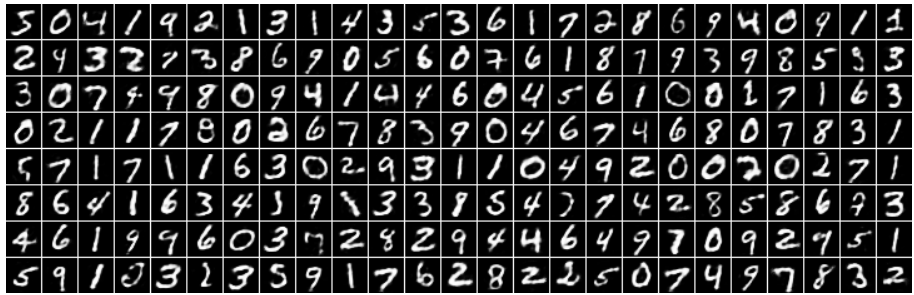}\label{fig:mnist_200_ae_rec}
}
\subfigure[Full intrinsic recall (200 transitions) ]{\hspace{-2pt}
\includegraphics[scale=0.471, trim=0 0 0 0, clip]{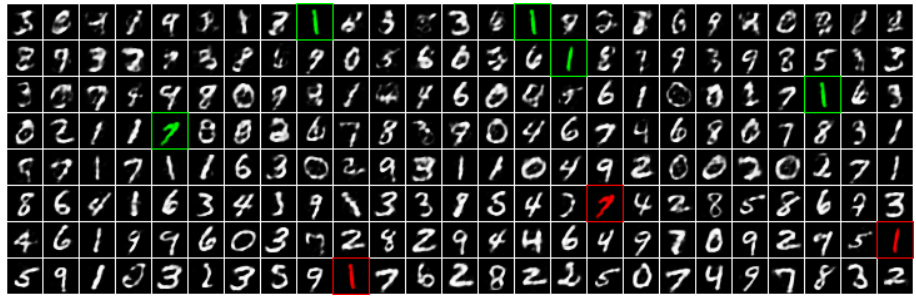}\label{fig:mnist_200_intrinsic_rec}
}
\caption{Visualization of \emph{MNIST} sequences, which read left to right, top to bottom. Thus the oldest pattern (index = 0) is located top left and the latest pattern (index 199) bottom right. (a) Sequence of input images as provided to SI. (b) Reconstruction of the same sequence encoded and directly decoded through SI~$\rightarrow$~EC~$\rightarrow$~SI (visualized EC ground truth). (c) Full intrinsic recall with 200 transitions for each pattern, which is visually indistinguishable from the reconstruction from the CA3 ground truth (CA3~$\rightarrow$~EC~$\rightarrow$~SI, not shown!) except for the patterns highlighted in green. Those patterns have a doppelganger in the training sequence, which is highlighted in red and whose subsequence is recalled instead.
}
\label{fig:mnist_200_data_all}
\end{center}
\end{figure}

When using real world data the performance of the network is confounded by information loss due to the compression of the input images in SI into the EC representation. 
Therefore we consider the EC representation as input ground truth, as we are not interested in the performance of the pathway SI~$\rightarrow$~EC~$\rightarrow$~SI, and visualize it by reconstructing the images from the EC representations.
Figure~\ref{fig:mnist_200_data} shows the entire input sequence of 200 \emph{MNIST} patterns stored in \emph{Model-B} with $N=200$. 
Figure~\ref{fig:mnist_200_ae_rec} shows the sequence when it has been encoded into a 200 dimensional hidden representation (EC) and decoded back to input space (SI). 
As the 784 dimensional real valued \emph{MNIST} patterns are compressed to 200 dimensional binary patterns there is some information loss leading to a visualized EC ground truth that is a smoothed version of the input data.
Notice, that this information loss depends on the size of the model relative to the input dimensionality, so that in case of $N=1000$ the 1000 dimensional hidden representation leads to an almost perfect reconstruction (data not shown).
Figure~\ref{fig:mnist_200_intrinsic_rec} shows the sequence of reconstructed EC patterns after full intrinsic recall, which are equivalent or at least similar to the visualized EC ground truth patterns shown in Figure~\ref{fig:mnist_200_ae_rec}.
The results are also visually indistinguishable from the reconstruction from the CA3 ground truth via CA3~$\rightarrow$~EC~$\rightarrow$~SI (results not shown) except for the patterns we highlighted in green. 
Those patterns have a doppelganger in the training sequence the system cannot distinguish from one another so that the sequence is recalled from the pattern that has been stored later highlighted in red.
Figures~\ref{fig:model_B_200_MNIST_encoder_intrinsic_all} and~\ref{fig:mnist_200_data_all} are from the same simulation.

\subsubsection{Visualization of Individual Recall Subsequences}

\begin{figure}[htbp!]
\begin{center}
\includegraphics[scale=0.95, trim=60 125 45 120, clip]{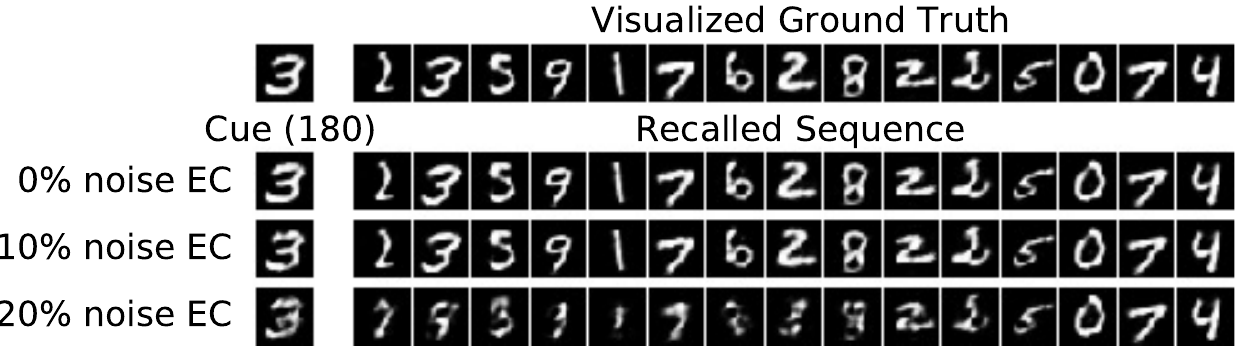}
\includegraphics[scale=0.95, trim=55 123 45 108, clip]{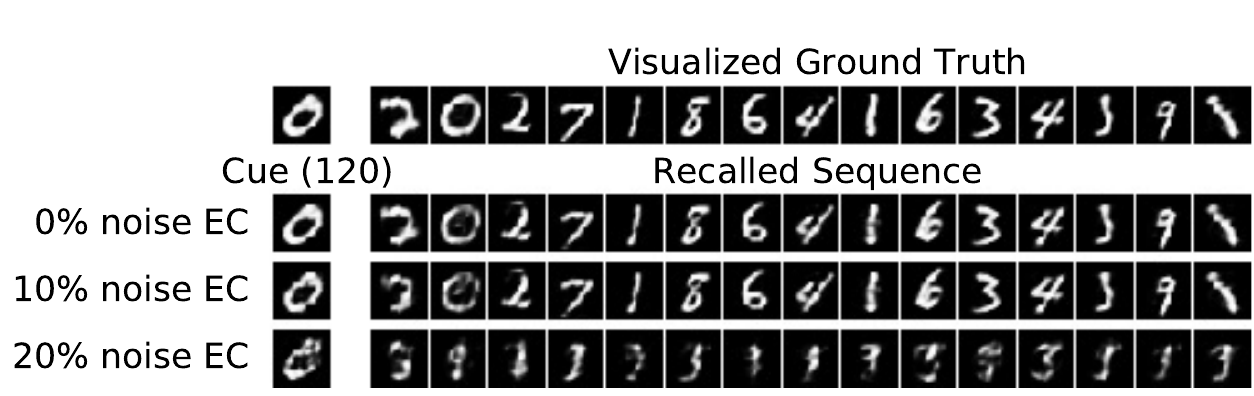}
\includegraphics[scale=0.95, trim=55 123 45 108, clip]{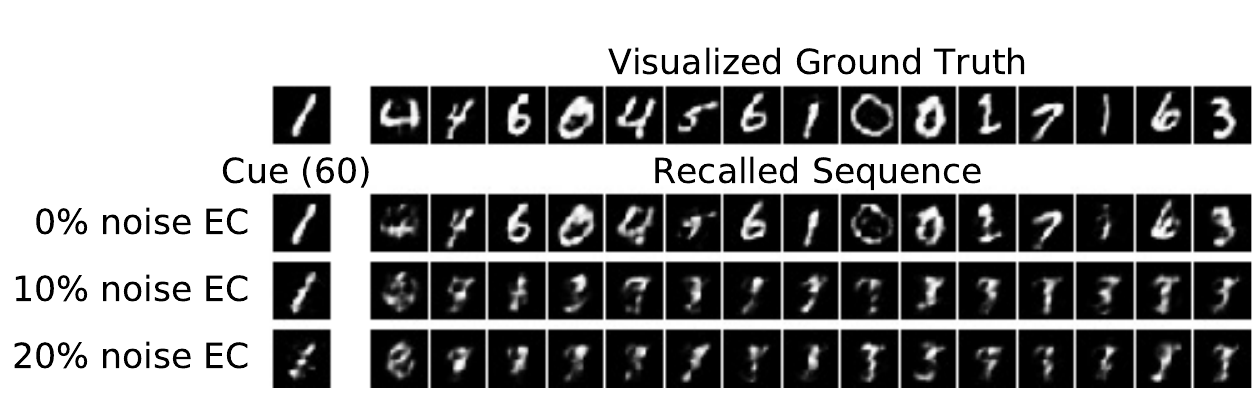}
\includegraphics[scale=0.95, trim=55 123 45 108, clip]{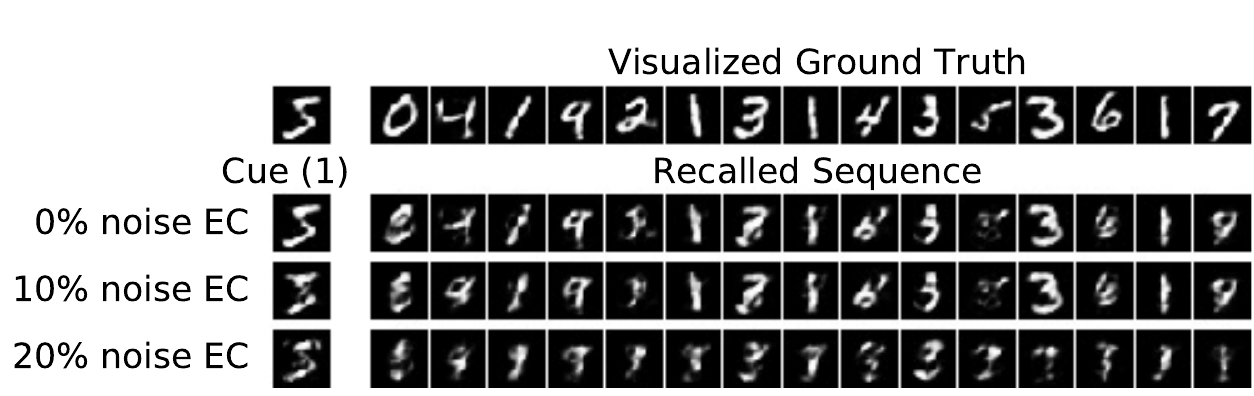}
\caption{Illustration of the recalled subsequences for different cues and noise levels for \emph{Model-B} on the \emph{MNIST} for $N=200$. 
Each cue is encoded to a CA3 pattern via pathway SI~$\rightarrow$~EC~$\rightarrow$~DG~$\rightarrow$~CA3. The intrinsic transition is iterated for 15 transitions where in each transition the current pattern in CA3 is decoded via pathway CA3~$\rightarrow$~EC~$\rightarrow$~SI to visualize the reconstructed digit pattern in SI.
The first row shows the visualized EC ground truth cue and subsequence that have been decoded via pathway EC~$\rightarrow$~SI to SI. The second row shows the reconstructions in SI when the exact pattern is presented in EC, and the remaining two rows show the results when either 10\% or 20\% binary noise is added in to the corresponding EC pattern. 
}
\label{fig:mnist_reconstruction_examples1}
\end{center}
\end{figure}

To further investigate the recall process we show the first 15 patterns of the recalled subsequence for different cue patterns in Figure~\ref{fig:mnist_reconstruction_examples1}. 
The cues are taken from position 1 (early), 60, 120, and 180 (late).
The first row in each sub-figure shows the visualized EC ground truth subsequence starting with the cue pattern (\emph{i.e.}\ the SI reconstruction from the ground truth EC patterns), which is thus a subsequence of the full sequence shown in Figure~\ref{fig:mnist_200_ae_rec}.
The second row shows the cue and recalled subsequence when the exact cue has been presented and the remaining two rows show corrupted cues and recalled subsequences, where either 10\% or 20\% binary noise has been added to the corresponding EC pattern.
All visualized patterns have been reconstructed via pathway EC~$\rightarrow$~SI.
For all cues the system recalls the correct subsequence after a short `sequence completion phase' when no noise is added. Except for cue 60 the system also succeeds when 10\% noise is added. For 20\% noise only cue 180 recalls the correct sequence, suggesting  that the recall process is more noise sensitive for earlier patterns.
Furthermore, the recall quality decreases with decreasing pattern index as expected from the degraded CA3~$\rightarrow$~EC~$\rightarrow$~SI performance (see Figure~\ref{fig:model_A_crosscorr_decoder} and Figure~\ref{fig:model_B_crosscorr_intrinsic_all} for \emph{RAND-CORR} dataset for example).

\subsubsection{On the Difficulty of the \emph{MNIST} Dataset}

\begin{figure}[htbp!]
\begin{center}
\includegraphics[scale=0.41, trim=85 0 90 0, clip]{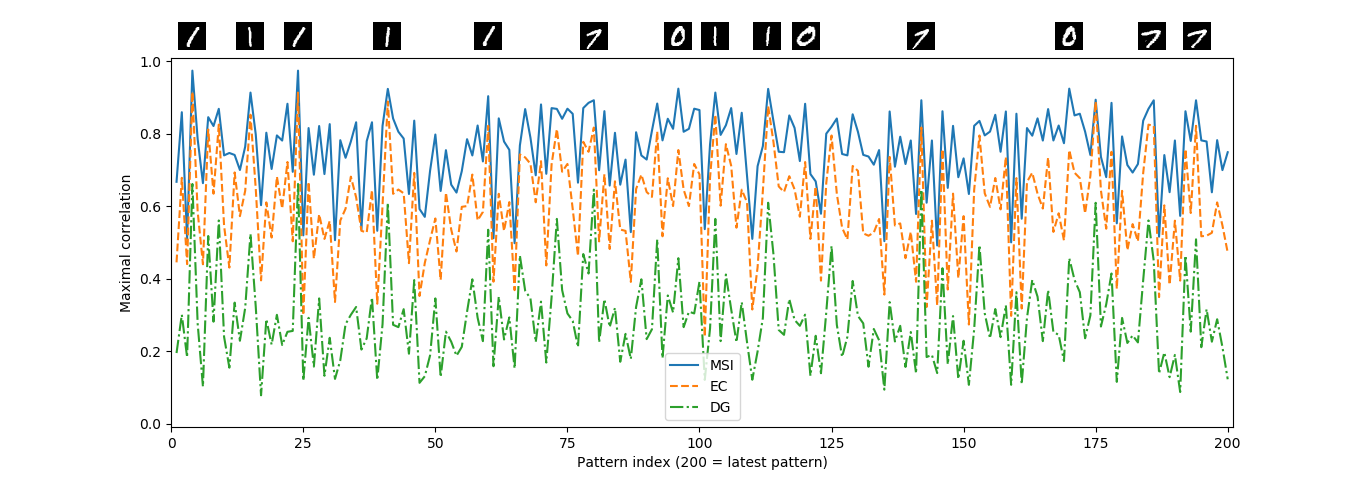}
\caption{Maximum of the correlation between each pattern and all the other patterns in subregions SI, EC, and DG for \emph{Model-B} on the \emph{MNIST} for $N=200$. For some of the peak values the corresponding patterns are shown above the corresponding position (pattern index). 
}
\label{fig:distribution_of_maximum_correlation}
\end{center}
\end{figure}

We have seen that a good recall, even in the presence of noise, works for late patterns, whereas the robustness to noise reduces with decreasing pattern index.
But there is a second property of the \emph{MNIST} dataset, which makes it sometimes hard for the model to relax to the correct position in the correct sequence in CA3.
Compared to the \emph{RAND-CORR} dataset, where we control the maximum correlation between two patterns in the input sequence to be approximately $0.8$, the \emph{MNIST} dataset has a much higher correlation for some of the patterns in the input sequence.
This is illustrated in Figure~\ref{fig:distribution_of_maximum_correlation}, which shows the maximal correlation between each pattern and all the other patterns within the sequence for subregions SI, EC, and DG, respectively.
While the maximal correlation reduces only slightly when the patterns are propagated from SI to EC, it reduces significantly when propagated further to DG. 
However, the relations within the distributions stay relatively the same, meaning that large correlation for a pattern in EC will still have a large correlation relative to the other patterns in DG.
We additionally show the patterns corresponding to the larges peaks, illustrating which patterns are most likely confused with another pattern in the sequence.
It is worth mentioning that due to the correlation the performance of \emph{Model-A} on the \emph{MNIST} dataset is much worse in comparison to the \emph{RAND-CORR} dataset (data not shown).

\begin{figure}[htbp!]
\vspace{-0.5cm}
\begin{center}
\includegraphics[scale=0.95, trim=55 125 45 115, clip]{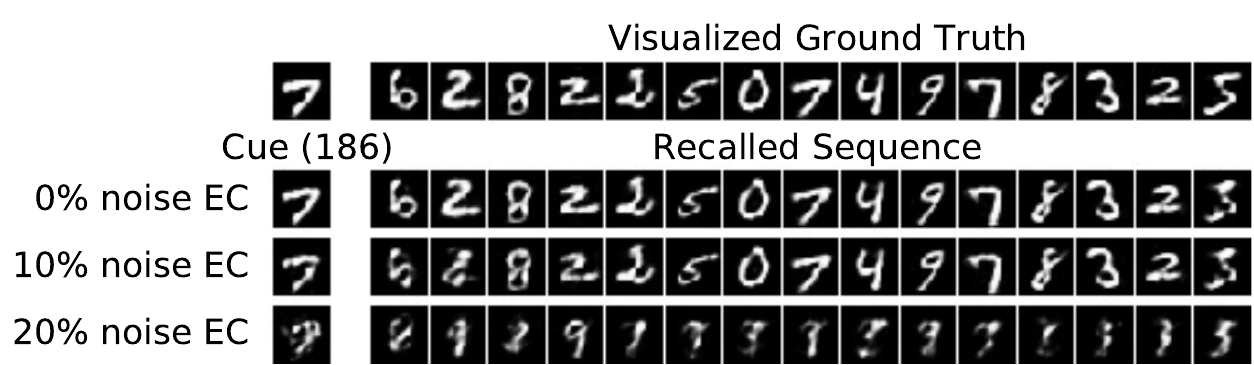}
\includegraphics[scale=0.95, trim=55 123 45 115, clip]{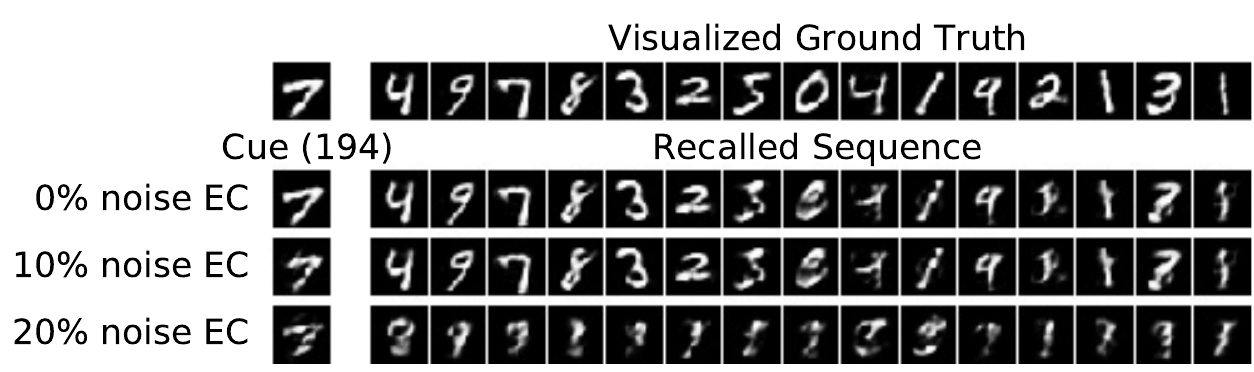}
\includegraphics[scale=0.95, trim=55 123 45 120, clip]{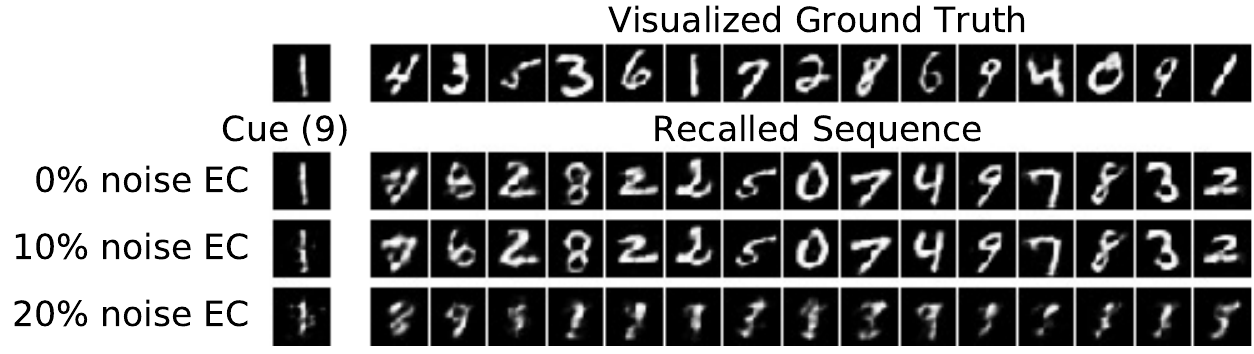}
\includegraphics[scale=0.95, trim=55 123 45 115, clip]{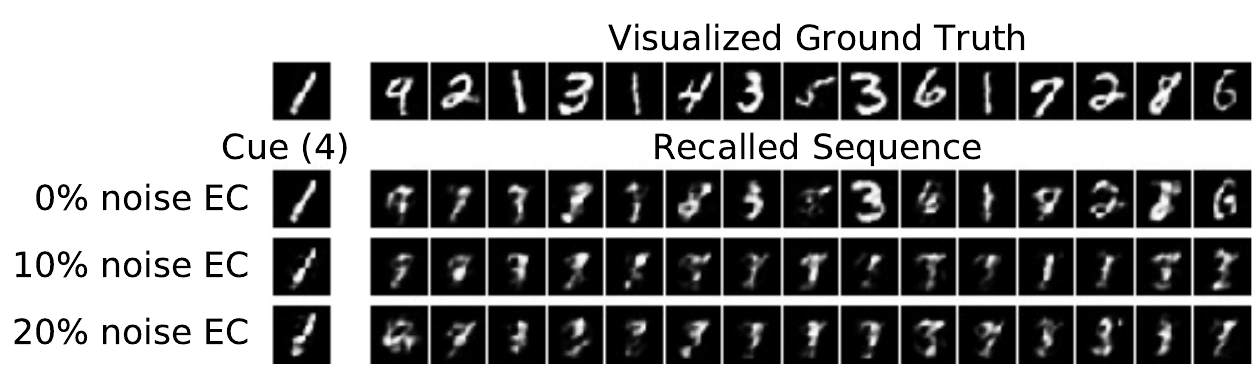}
\includegraphics[scale=0.95, trim=55 123 45 115, clip]{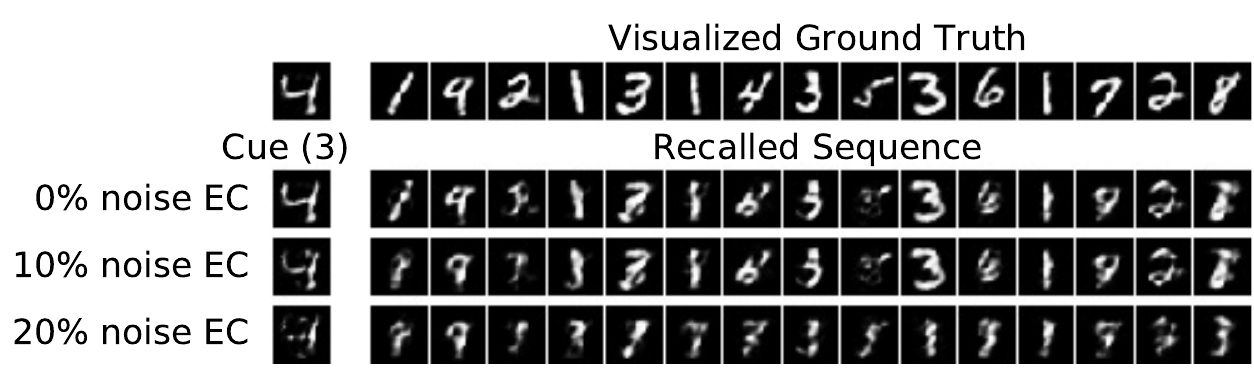}
\caption{Illustration of recalled subsequences for different pairs of cues that are highly correlated for \emph{Model-B} on the \emph{MNIST} for $N=200$. 
Cue 186 with 0\% noise is an example for correct relaxation and for relaxation to s spurious sequence with 20\% noise. Cue 9 is an example for a relaxation to a wrong position (See Figure~\ref{fig:error_illustration}). 
}
\label{fig:mnist_reconstruction_examples2}
\end{center}
\end{figure}

Figure~\ref{fig:mnist_reconstruction_examples2} illustrates the recall performance of two similar late patterns with index 186 and 194. These patterns are also shown on the right hand side of Figure~\ref{fig:distribution_of_maximum_correlation}.
Without noise the correct subsequence can be recalled with rather high quality, whereas in the presence of 10\% noise both cues recall a subsequence that is close to the visualized EC ground truth subsequence of the more recent pattern with index 194.
This implies that if a pattern has a doppelganger in the dataset it is more sensitive to noise because it can easily relax to the doppelganger's subsequence.
For 20\% noise the dynamics does not relax to either of the two subsequences within 15 intrinsic transitions.
The sensitivity of doppelganger patterns to noise gets even worse for smaller pattern indices, as can be seen by comparing the very similar cues with index 60 and four shown in Figure~\ref{fig:mnist_reconstruction_examples1} and Figure~\ref{fig:mnist_reconstruction_examples2}, respectively.
The cue with index nine is one of the five patterns highlighted in Figure~\ref{fig:mnist_200_data_all}, where even without the presents of noise the system gets confused by a doppelganger. In this case the doppelganger whose subsequence is recalled is the pattern with index 185 highlighted in red in Figure~\ref{fig:mnist_200_data_all}.
This can also be seen from Figure~\ref{fig:mnist_reconstruction_examples2} where the recalled sequences of the pattern with index 9 is same as the one for the pattern with index 186 shifted by one pattern.
In contrast, the cue with index three shown in Figure~\ref{fig:mnist_reconstruction_examples2}, which does not have a doppelganger, recalls the correct subsequence with rather good quality even for 20\% noise although it needs a longer `sequence completion phase' in this case.
For 50\% noise the system generates a spurious sequence in almost all cases and thus does not recall any meaningful pattern in SI.

\subsubsection{Visualization of Recalled Subsequences for Novel Input Patterns}

\begin{figure}[htbp!]
\begin{center}
\includegraphics[scale=0.95, trim=55 145 45 140, clip]{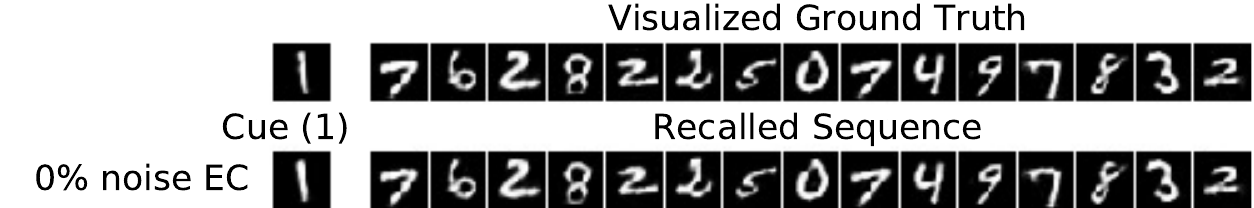}
\includegraphics[scale=0.95, trim=55 145 45 120, clip]{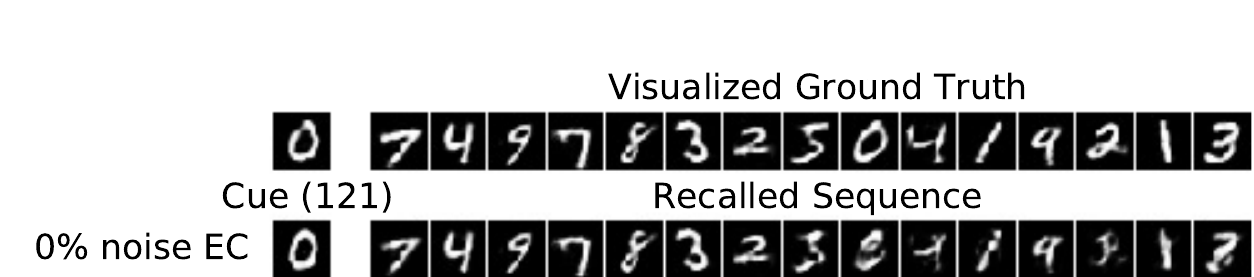}
\includegraphics[scale=0.95, trim=55 145 45 120, clip]{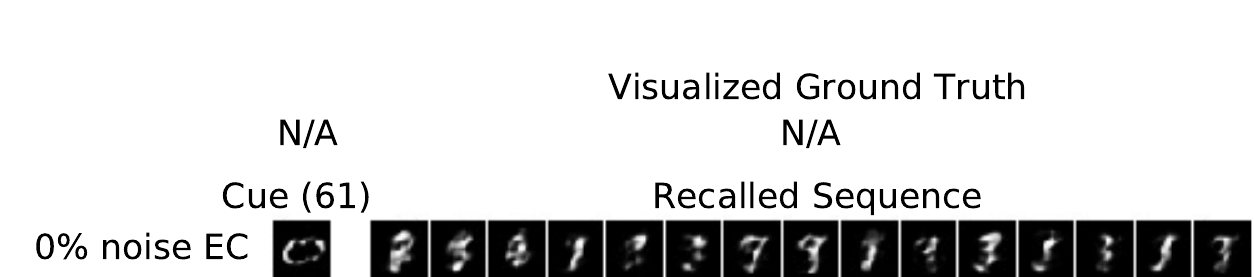}
\includegraphics[scale=0.95, trim=55 145 45 120, clip]{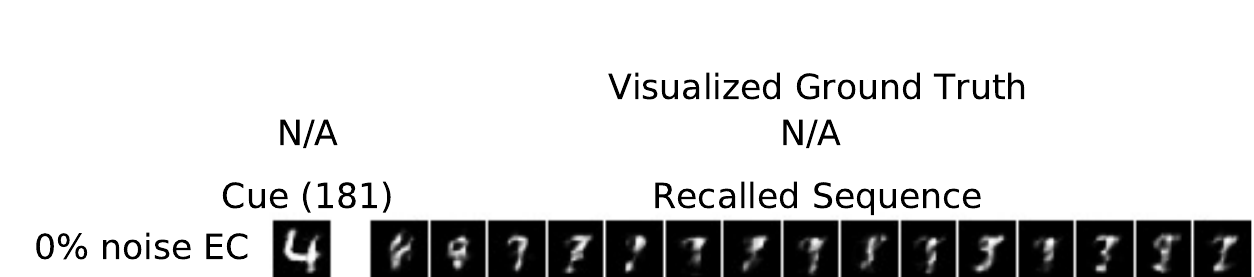}
\caption{Illustration of the recalled subsequences for different cues of the test data (novel patterns) for \emph{Model-B} on the \emph{MNIST} for $N=200$.
For the first two patterns the stored sequence is recalled starting from a similar stored pattern, while for the last two pattern spurious sequences are recalled. 
\vspace{-0.2cm}
}
\label{fig:mnist_reconstruction_test_examples}
\end{center}
\end{figure}

We have seen that the model is robust to noise to some extend, but when the cue presented differs too much from the correct pattern the model does not recall any pattern of the visualized EC ground truth  sequence.
This is a desired property of the model as it should only recall the corresponding sequence when the presented cue pattern has a high similarity to the visualized EC ground truth cue pattern.  
To verify this we selected four patterns of the test dataset that the model has not seen before and presented them as cues to the network. 
Two of those patterns are rather similar to at least one pattern in the training sequence while the other two test patterns differ sufficiently from all patterns in the training sequence.
The results are shown in Figure~\ref{fig:mnist_reconstruction_test_examples}. 
The top row in the first two subfigures shows the manually determined ground truth sequence to which the recalled sequence, shown in the second row, is most similar. 
For the other two cues there exists no ground truth as the recalled spurious sequence is significantly different from all shifted version of the ground truth sequence.

\subsection{Storing a Sequence of Natural Images with \emph{Model-B}}\label{sec:experiments_hippo_natural_images}

\begin{figure}[t!]
\begin{center}
\subfigure[Input data]{\hspace{-2pt}
\includegraphics[scale=0.41, trim=0 0 0 0, clip]{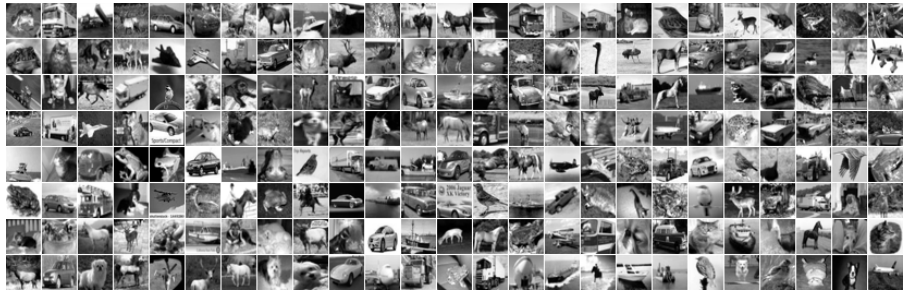}\label{fig:cifar_200_data}
}
\subfigure[Reconstruction via SI~$\rightarrow$~EC~$\rightarrow$~SI]{\hspace{-2pt}
\includegraphics[scale=0.41, trim=0 0 0 0, clip]{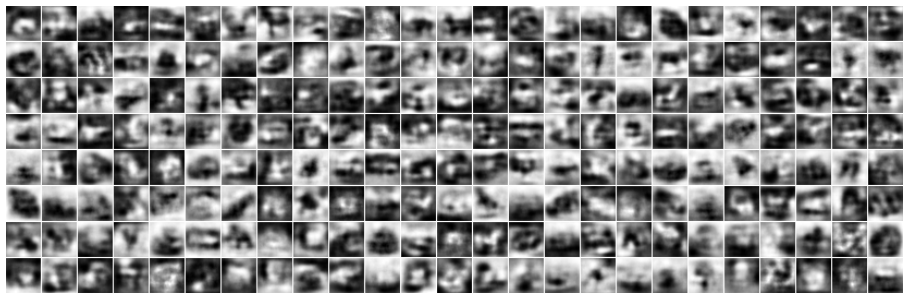}\label{fig:cifar_200_ae_rec}
}
\subfigure[Reconstruction via CA3~$\rightarrow$~EC~$\rightarrow$~SI / full intrinsic recall (200 transitions)]{\hspace{-2pt}
\includegraphics[scale=0.41, trim=0 0 0 0, clip]{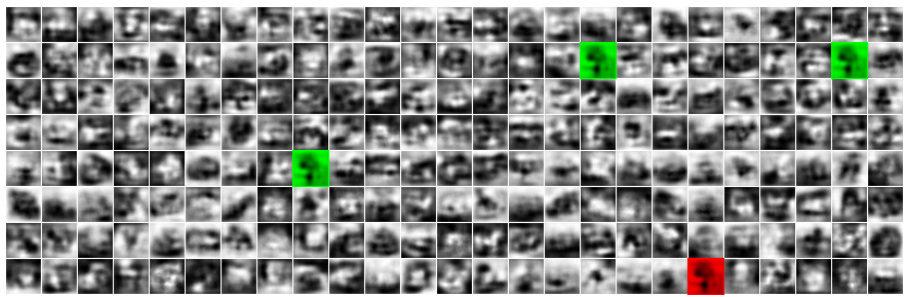}\label{fig:cifar_200_intrinsic_rec}
}
\caption{Visualization of \emph{CIFAR} sequences, which read left to right, top to bottom. Thus the oldest pattern (index = 0) is located top left and the latest pattern (index 199) bottom right. (a) Sequence of input images as provided to SI. (b) Reconstruction of the same sequence encoded and directly decoded through SI~$\rightarrow$~EC~$\rightarrow$~SI (visualized EC ground truth). (c) Full intrinsic recall with 200 transitions for each pattern, which is is visually indistinguishable from the reconstruction from the ground truth CA3 patterns (CA3~$\rightarrow$~EC~$\rightarrow$~SI) except for the patterns highlighted in green. For those patterns the system gets confused by the pattern in the training sequence highlighted in red.
}
\label{fig:cifar_200_data_all}
\end{center}
\end{figure}

We performed the same experiments as in the previous section for the \emph{CIFAR} dataset, which contains small natural images.
We only show selected results that demonstrate that the model has a similar performance as for the other datasets.
Figure~\ref{fig:cifar_200_data_all} 
shows the input sequence in SI, the sequence when reconstructed from the compressed 200 dimensional EC representation, and the patterns finally retrieved after a full intrinsic recall.

The compression of natural images to a 200 dimensional binary representation using a single layer network results in a large reconstruction error as can be seen by comparing Figure~\ref{fig:cifar_200_data} with Figure~\ref{fig:cifar_200_ae_rec}.
Even for the model with size $N=1000$ the reconstruction is far form perfect (data not shown).

\begin{figure}[htbp!]
\begin{center}
\subfigure[$N=200$]{
\includegraphics[scale=0.4, trim=10 5 32 40, clip]{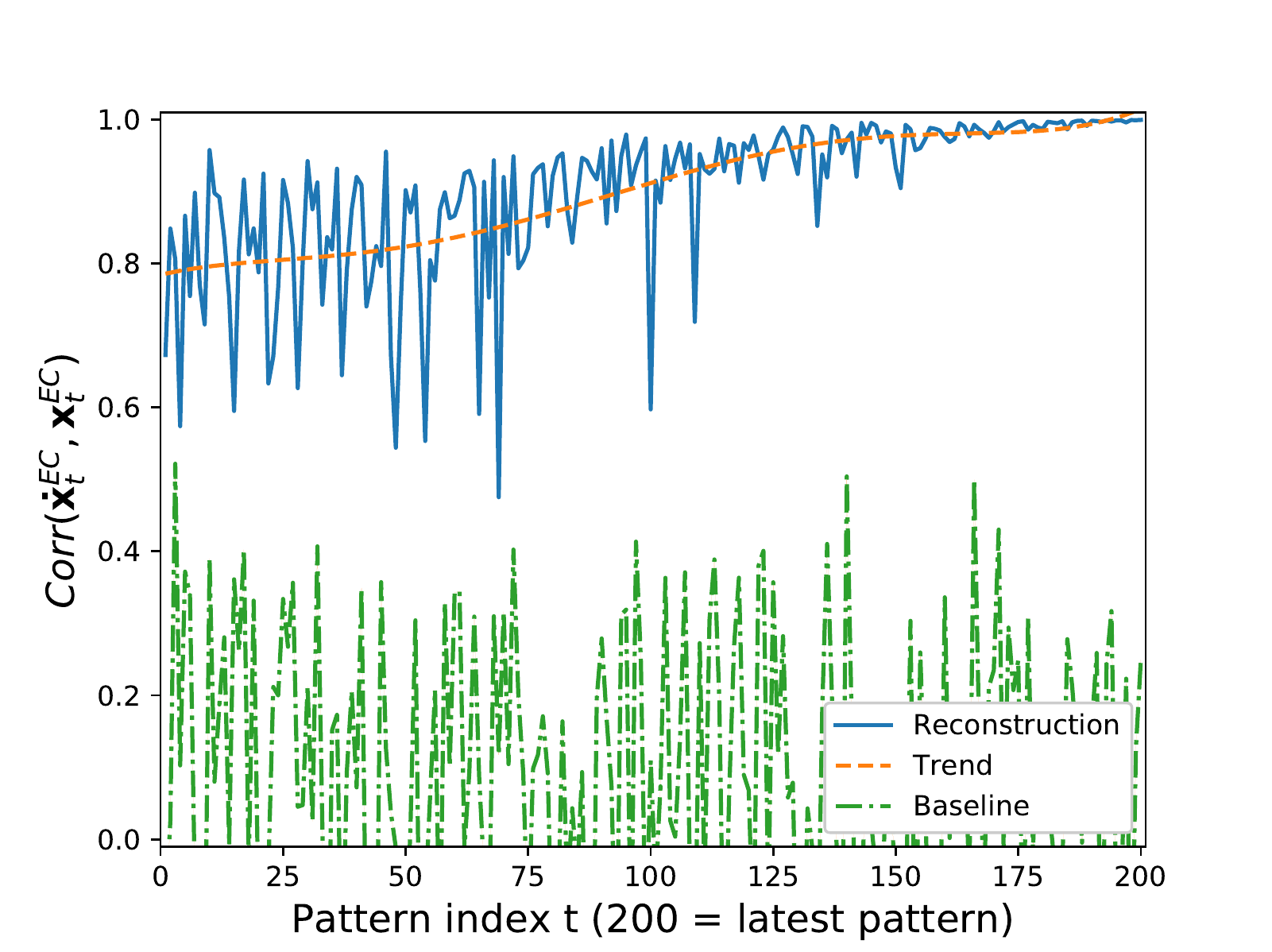}\label{fig:model_B_200_cifar_intrinsic_all}}
\subfigure[$N=1000$]{
\includegraphics[scale=0.4, trim=10 5 32 40, clip]{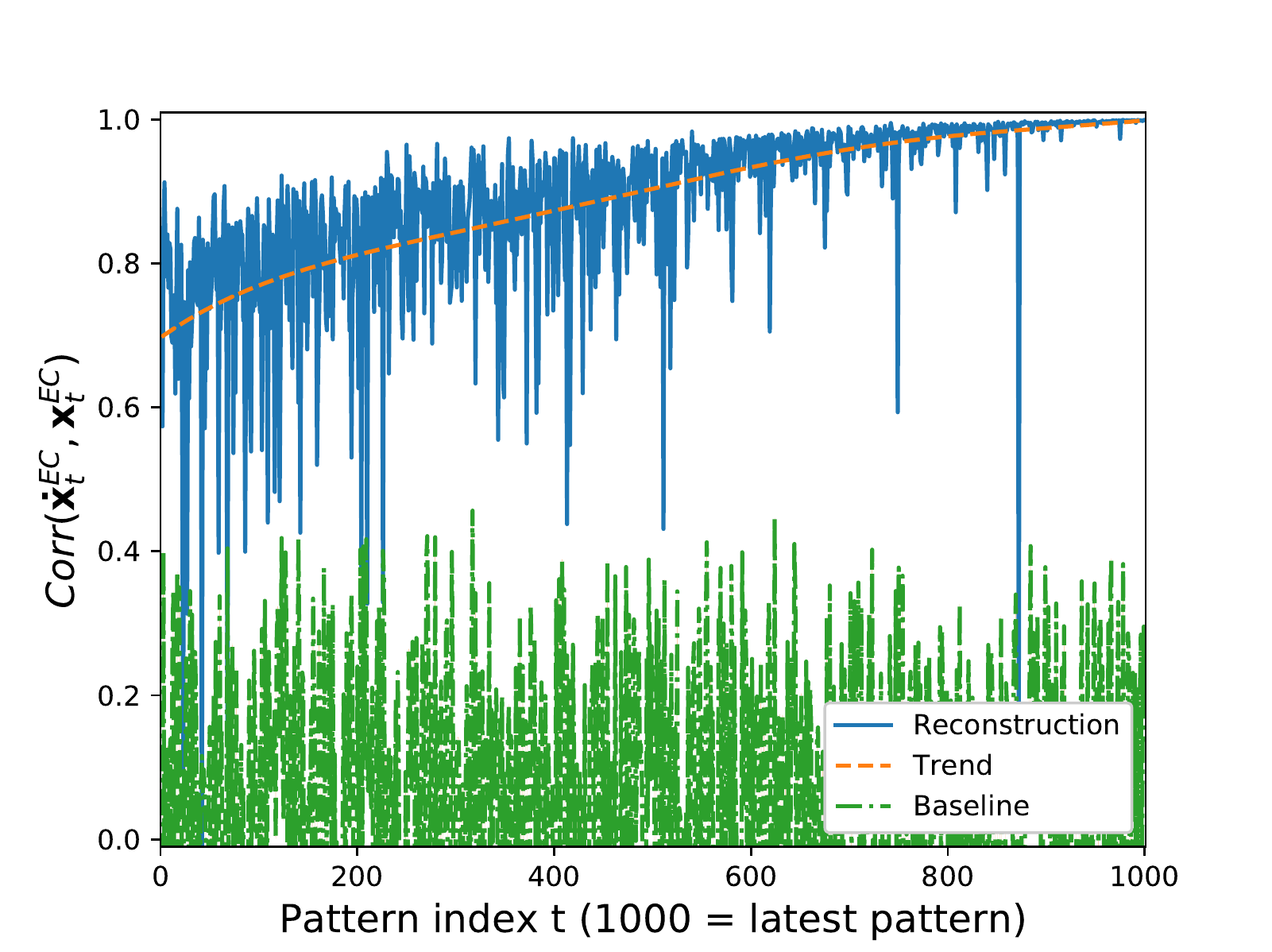}\label{fig:model_B_1000_cifar_intrinsic_all}}
\caption{Recall performance of \emph{Model-B} on the \emph{CIFAR} for (a) $N=200$ and (b) $N=1000$. 
Also compare with the performance of \emph{Model-B} on the \emph{MNIST} dataset shown in Figure~\ref{fig:model_B_200_MNIST_encoder_intrinsic_all}.
}
\label{fig:model_B_cifar_intrinsic_all}
\end{center}
\end{figure}

The full intrinsic reconstruction recall performance apart from the reconstruction error, is good for most of the patterns shown in Figure~\ref{fig:cifar_200_intrinsic_rec}.
This can also be seen from Figure~\ref{fig:model_B_200_cifar_intrinsic_all}, which shows the full recall performance for \emph{Model-B} with size $N=200$.
Figure~\ref{fig:model_B_1000_cifar_intrinsic_all} shows the results for $N=1000$. The results of both network sizes and both datasets, \emph{CIFAR} and \emph{MNIST} (Figures~\ref{fig:model_B_1000_MNIST_encoder_intrinsic_all} and~\ref{fig:model_B_200_MNIST_encoder_intrinsic_all}), are comparable.

\subsection{Average Activity in CA3}\label{sec:decreasing_ca3_activity}

\begin{figure}[htbp!]
\begin{center}
\subfigure[\emph{RAND}, $N=200$, CA3 activity 10 \%]{
\includegraphics[scale=0.4, trim=10 5 32 40, clip]{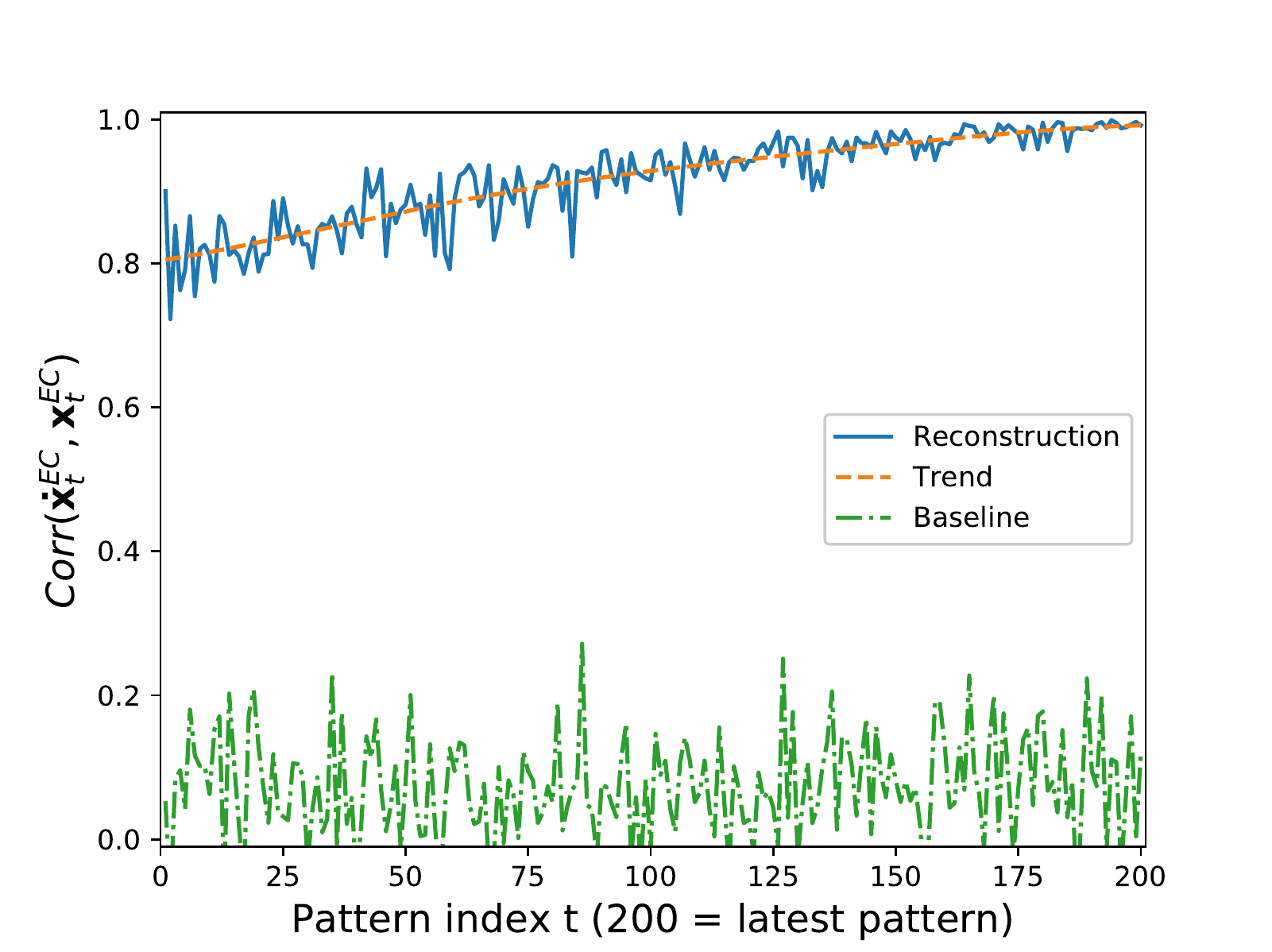}\label{fig:performance_intrinsic_steps_200_uncorr_activity_01}}
\subfigure[\emph{RAND}, $N=1000$, CA3 activity 3.2 \%]{
\includegraphics[scale=0.4, trim=10 5 32 40, clip]{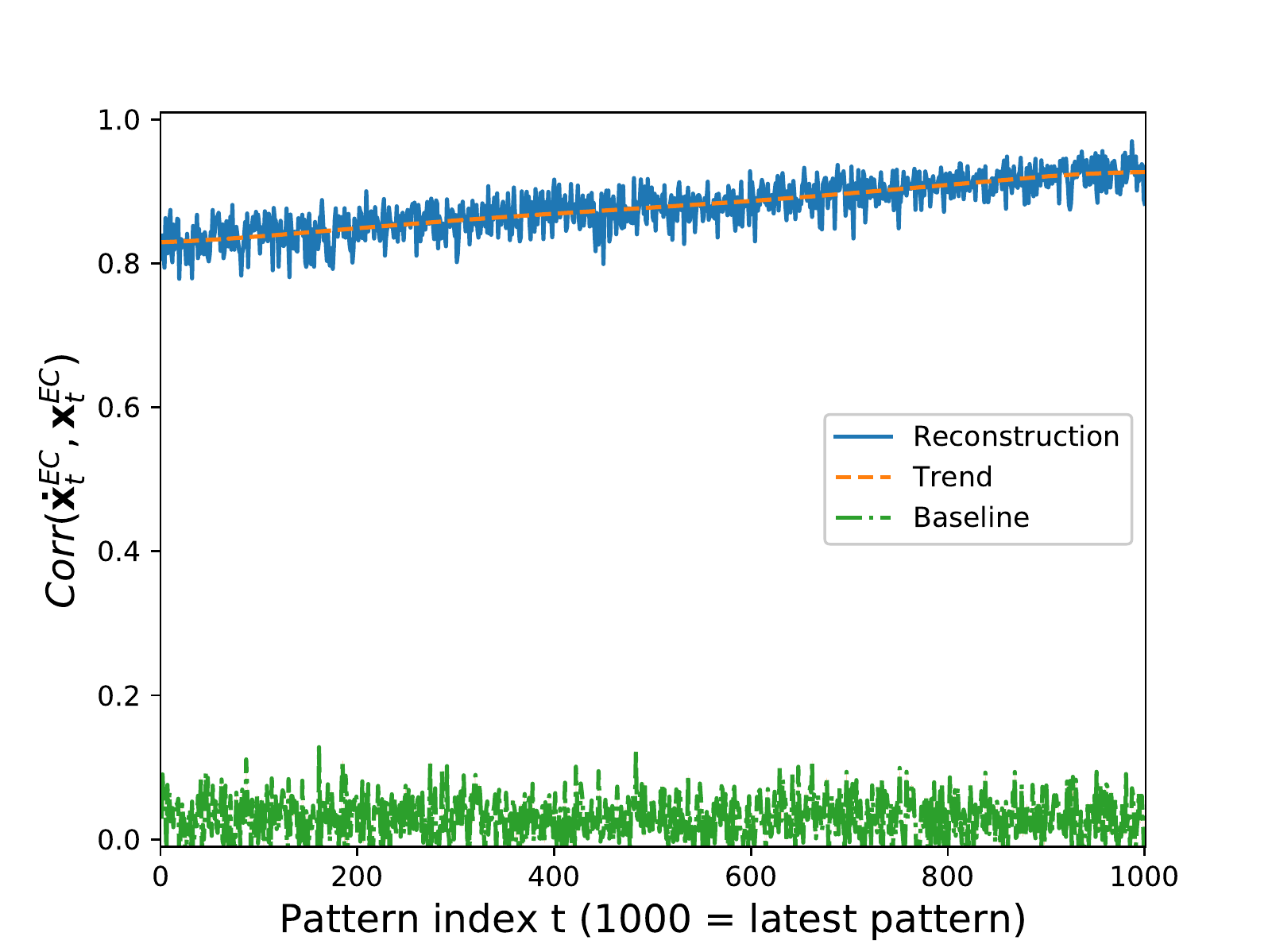}\label{fig:performance_intrinsic_steps_1000_uncorr_activity_0032}}
\subfigure[\emph{MNIST}, $N=200$, CA3 activity 10 \%]{
\includegraphics[scale=0.4, trim=10 5 32 40, clip]{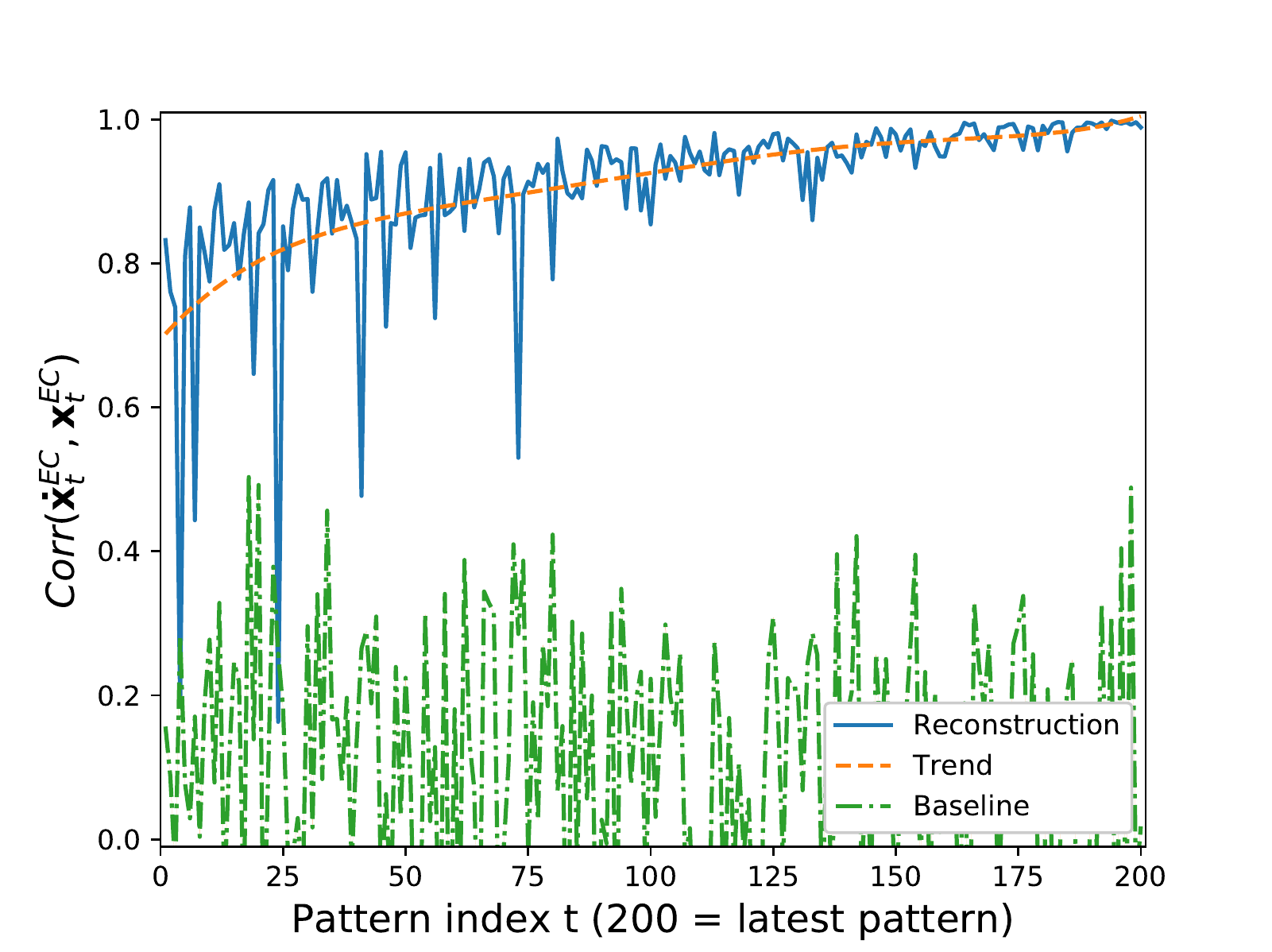}\label{fig:performance_intrinsic_steps_200_mnist_activity_01}}
\subfigure[\emph{MNIST}, $N=1000$, CA3 activity 3.2 \%]{
\includegraphics[scale=0.4, trim=10 5 32 40, clip]{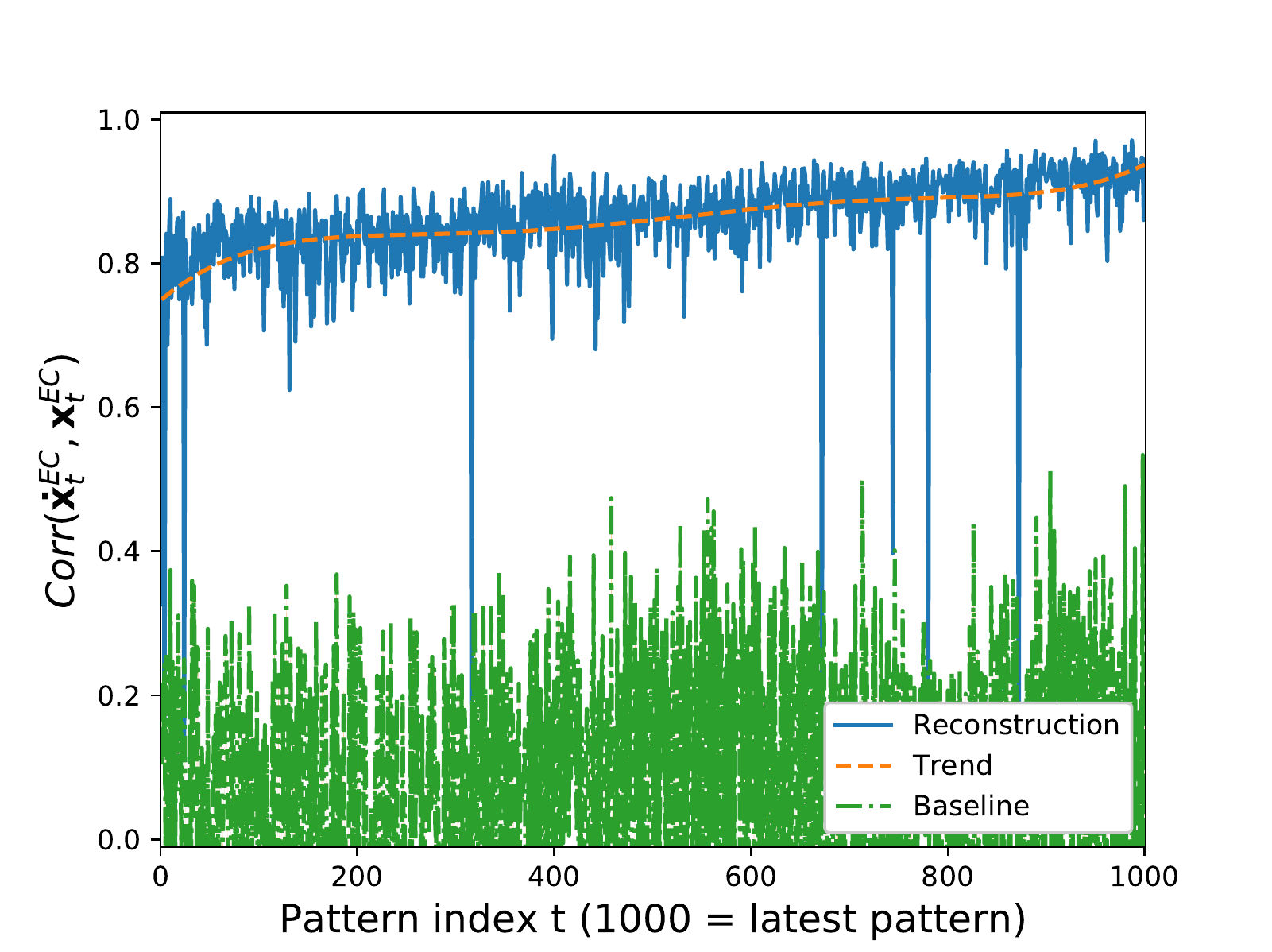}\label{fig:performance_intrinsic_steps_1000_mnist_activity_0032}}
\subfigure[\emph{RAND-CORR}, $N=200$, CA3 activity 10 \%]{
\includegraphics[scale=0.4, trim=10 5 32 40, clip]{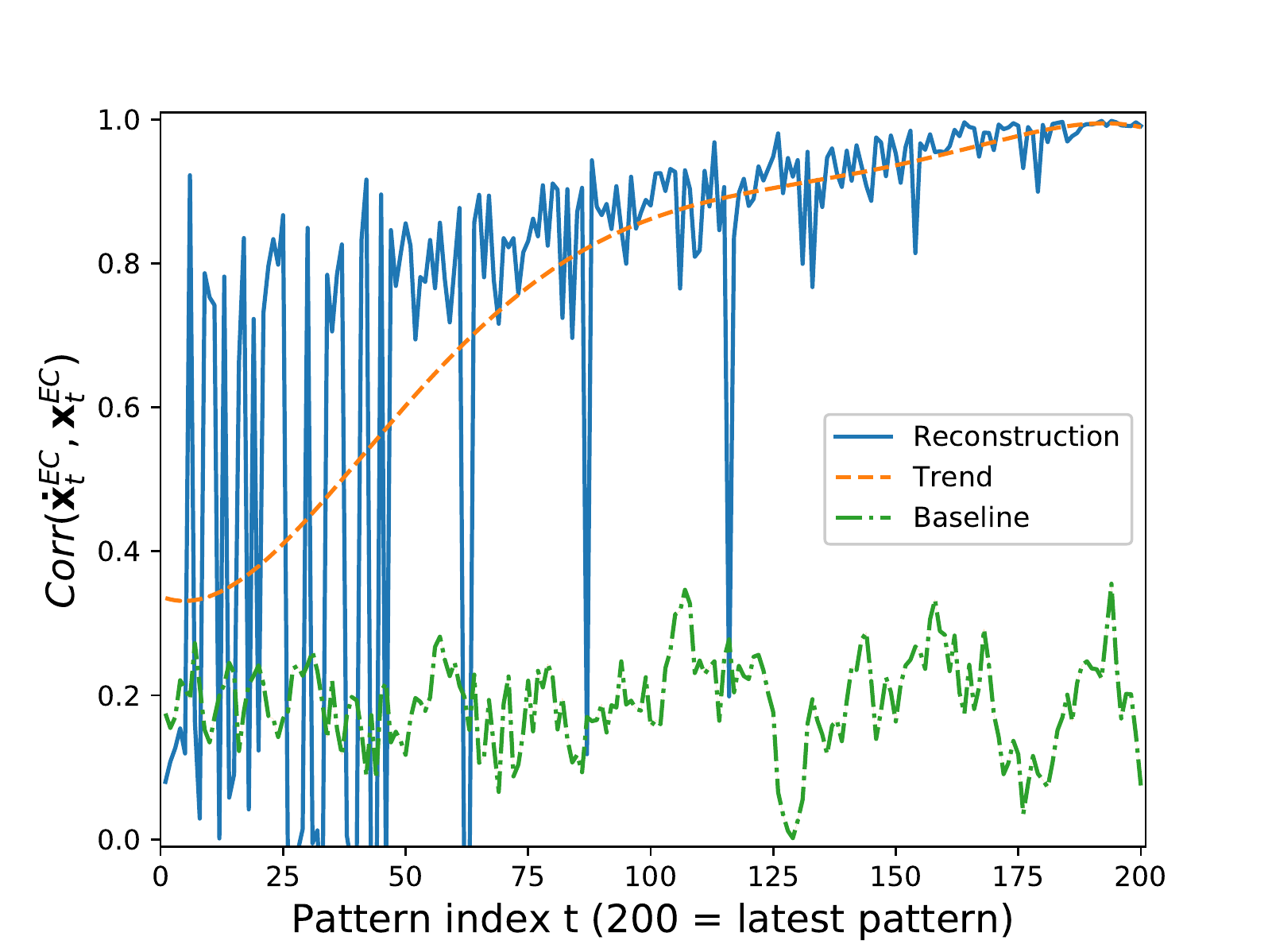}\label{fig:performance_intrinsic_steps_200_crosscorr_activity_01}}
\subfigure[\emph{RAND-CORR}, $N=1000$, CA3 activity 3.2 \%]{
\includegraphics[scale=0.4, trim=10 5 32 40, clip]{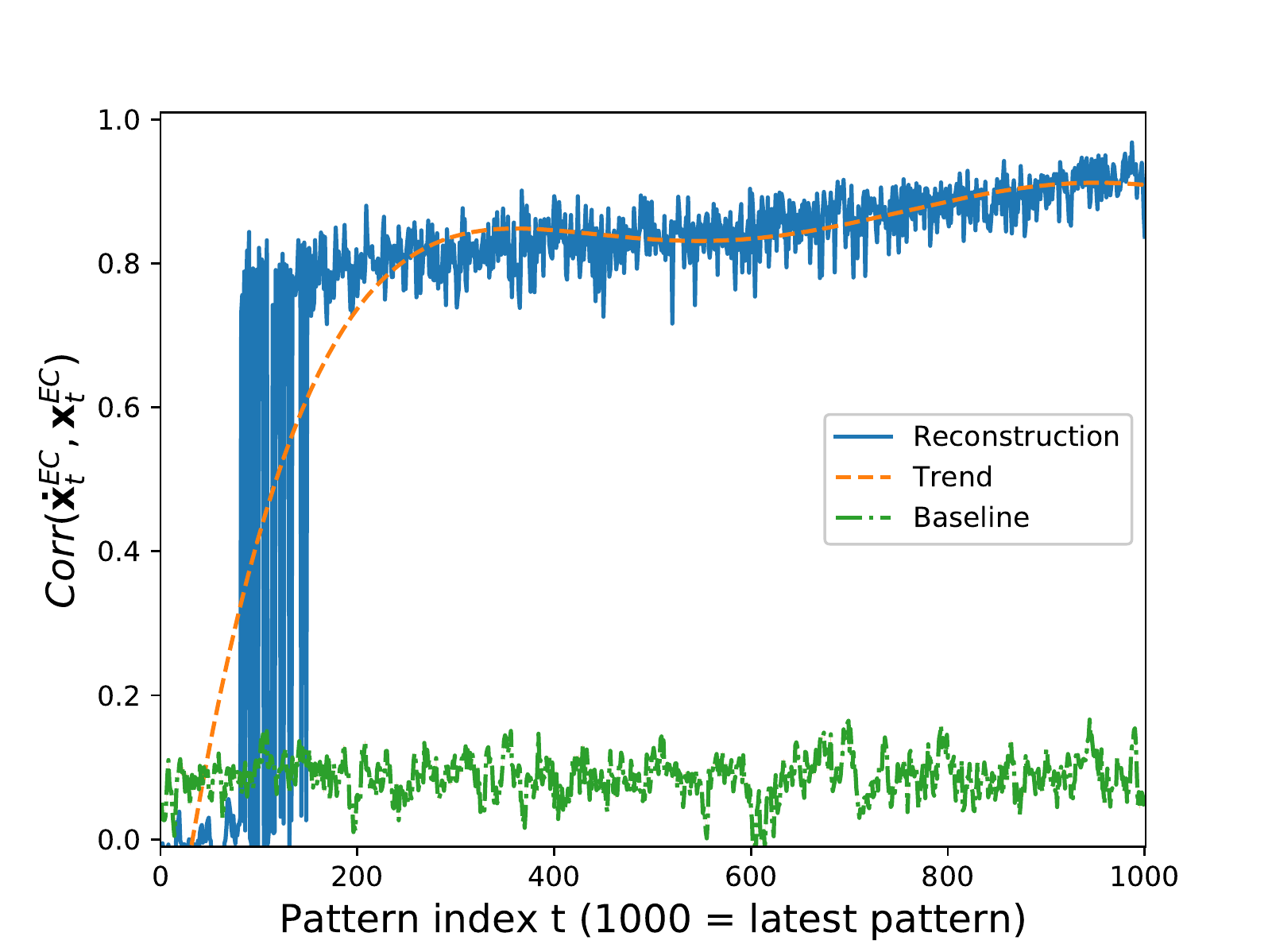}\label{fig:performance_intrinsic_steps_1000_crosscorr_activity_0032}}
\caption{Full intrinsic performance of \emph{Model-B} on (a) \emph{RAND} with N=200 and CA3 activity of 10\%, (b) \emph{RAND} with N=1000 and CA3 activity of 3.2\%, (c) \emph{RAND-CORR} with N=200 and CA3 activity of 10\%, (d) \emph{RAND-CORR} with N=1000 and CA3 activity of 3.2\%, (e) \emph{MNIST} with N=200 and CA3 activity of 10\%, and (f) \emph{MNIST} with N=1000 and CA3 activity of 3.2\%. Apart from the CA3 activity and $N$ the same setup has been used for all networks.
Also compare with the performance for $N=1000$ and 20\% activity shown in Figure~\ref{fig:model_A_uncorr_intrinsic_all}, Figure~\ref{fig:model_B_1000_mnist_intrinsic_all}, and Figure~\ref{fig:model_B_crosscorr_intrinsic_all}.
}
\label{fig:decreasing_ca3_activity}
\end{center}
\end{figure}

The experiments shown above use an average activity of 20\%. The best performance could actually be achieved with 40\% activity (data not shown), but since the activity in the hippocampus (3.2\%) is much lower we have chosen 20\% as the lowest activity that has almost the same performance as 40\% for both network sizes.  
Figure~\ref{fig:decreasing_ca3_activity} (right column) shows the performance when a physiological average activity of 3.2\% is used for a model of size $N=1000$, which shows still a rather good performance on all three datasets, except for early patterns in \emph{RAND-CORR}.
For $N=200$, however, an activity of 3.2\% does not sufice (data not shown), an average activity of 10\% is required to reach a reasonable performance (Figure~\ref{fig:decreasing_ca3_activity} left column).
This gives evidence that larger networks require less activity, so that 3.2\% might actually be optimal for a network of physiological size (N=100.000).
The results also confirm that the performance degrades with increasing correlations in the dataset, which is lowest for \emph{RAND} and highest for \emph{RAND-CORR}.
The same findings can be observed for the \emph{CIFAR} dataset (data not shown).

If one adds noise to the patterns the performance of the smaller networks degrades much quicker with increasing noise level than that of the larger networks. 
This supports our assumption that the absolute number of active neurons rather than the percentage is important and one can expect that the robustness will also improve with increasing model size.

\section{Discussion and Conclusion}\label{sec:discussion_and_conclusion}


%

Several biologically motivated studies have already analyzed the storage and retrieval of pattern sequences, some of which used more realistic spiking neurons~\citep{maass1997networks, rao2000predictive, gutig2006tempotron, jahnke2015unified, hawkins2016neurons}.
In contrast to the model presented in this work most of these studies analyze CA3 as an auto-associative memory isolated from the hippocampal circuit, only use artificially generated data, and most importantly none of them considers online learning of sequences.
Most related to our model is the work by~\citet{Bayati2018}, which, according to the CRISP framework, also addresses the storage and recall of pattern sequences through hetero-association of input sequences with intrinsically generated sequences in CA3.
In contrast to our work, however, CA1 is considered instead of DG as an additional processing step and, like the other studies, only off\-line learning is used.
To the best of our knowledge the model we present in this work is the first neural model of the hippocampus that allows to continuously store and retrieve pattern sequences in a one-shot fashion with a high capacity and a plausible forgetting behavior.

The standard framework~\citep{nadel1997memory} proposes that plasticity in CA3 is responsible for the storage of episodic memories in the hippocampus, but we have shown in Section~\ref{sec:comparison_standard} that online hetero-association of patterns in recurrent CA3 synapses leads to a poor performance and can thus not be considered as a reasonable alternative to our proposed architecture.
The first key result of our work is thus that intrinsic pattern sequences in CA3, as proposed by CRISP, is a particularly suitable requisite
for successful one-shot storage of pattern sequences. 
The second key result is that a simple generic decorrelation mechanism implemented in DG through a sparsification of patterns is sufficient to overcome the problem of correlations in the input (as shown in Section~\ref{sec:storing_temporally_correlated_patterns_in_modelb} for example). 
Our model thus supports the common assumption that DG serves as a pattern separator, which is consistent with the standard framework but differs from the role of sequence separation assigned to DG in the CRISP framework. 

It has been demonstrated in various memory tests that the functional form of forgetting is best characterized by a power function~\citep{wixted1990analyzing}.
The degeneration of patterns stored in our model follows a flat power law distribution (see Figure~\ref{fig:model_A_uncorr_all} and Figure~\ref{fig:model_B_all_crosscorr} for examples), showing that our model has a biologically plausible forgetting behavior.
The usage of Hebbian-descent as a learning rule with centering (mean free input statistics) is thereby crucial as it allows to continuously store new patterns while early stored patterns are forgotten gradually over time.
Nevertheless, the model has still a rather high capacity of 40\% with respect to the number of CA3 neurons for arbitrary input data, which is even higher for uncorrelated data (see Figure~\ref{fig:model_B_crosscorr_intrinsic_all} for example). 
Moreover, the recall of patterns still works reasonably well in the presence of noise in EC and is more robust for recently stored patterns (see Figure~\ref{fig:mnist_reconstruction_examples1} for example).

Compared to the average activity known for the rat hippocampus we have chosen much higher average activity, but at the same time also much fewer neurons per subregion.
We have argued that, independent of the model size, a certain number of active neurons are crucial for a good performance of the system, so that small models require a higher average activity than large models.
We have shown in Section~\ref{sec:decreasing_ca3_activity} that the required average activity in CA3 indeed decreases with increasing model size, so that large networks may indeed be used to store sequences at average activity known for the rat hippocampus.

Our model has the notable property to be able to `bootstrap' (improve) itself based on a process we call `dreaming' that does not require external input (Section~\ref{sec:experiments_dreaming}).
Instead in dreaming sequences are recalled intrinsically, which matches the observed hippocampal phenomenon of replay~\citep{skaggs1996replay}
and might therefore give evidence that the role of replay is memory consolidation. 
Dreaming can be used to enhance the encoding pathway EC~$\rightarrow$~CA3 such that the model can deal with correlated patterns even without subregion DG. 
While we analyze the pathways EC~$\rightarrow$~CA3 and EC~$\rightarrow$~DG~$\rightarrow$~CA3 through \emph{Model-A} and \emph{Model-B} separately, they exist simultaneously in the hippocampus.
Our results suggest that the two pathways can act as a fast online (EC~$\rightarrow$~DG~$\rightarrow$~CA3) and a slow off\-line (EC~$\rightarrow$~CA3) way of hetero-associating pattern sequences, where the latter is way more efficient in terms of the number of neurons and connections.
The process of dreaming could therefore be used to integrate all or some associations stored in pathway EC~$\rightarrow$~DG~$\rightarrow$~CA3 into EC~$\rightarrow$~CA3.
Furthermore, one could think of dreaming as a way to transfer memories from hippocampus to the cortex via systems consolidation~\citep{buzsaki1989two}, which is an interesting future research direction.

We have shown that our model can identify new unseen patterns as novel as long as they are different enough to all stored patterns (see Figure~\ref{fig:mnist_reconstruction_test_examples} for example).
This could be used to implement novelty detection, which in agreement with other studies~\citep{lisman2001storage, lee2005role,duncan2012evidence} we believe is performed by CA1.
Although the integration of subregion CA1 is postponed to future work, we want to discuss here how to train and integrate CA1 as a novelty/familiarity detector in our model.
Instead of the pathway CA3~$\rightarrow$~EC we would have the pathway CA3~$\rightarrow$CA1~$\rightarrow$~EC.
One could now imagine that the subregion CA1 could be trained to identify whether the activity received from CA3 does or does not belong to the intrinsic sequence and therefore perform novelty/familiarity detection on CA3/CA1 patterns.
Thus when a test pattern such as the fourth pattern in Figure~\ref{fig:mnist_reconstruction_test_examples} triggers a spurious intrinsic sequence,
CA1 would identify that the received activity from CA3 does not belong to a valid intrinsic pattern and send out a novelty signal.
When the test pattern such as the first pattern in Figure~\ref{fig:mnist_reconstruction_test_examples} recalls the correct intrinsic sequence or at least very similar patterns,
CA1 would identify this and send out a familiarity signal.
Similar to CA3, CA1 can be pre-trained using Hebbian-Descent on the intrinsic sequence of CA3 and does thus not affect the online learning ability of the model.

Whereas most studies have focused only on artificially generated data, we have exemplified that our model is also capable of storing and retrieving patterns of real world datasets such as handwritten digits and natural images (Section~\ref{sec:experiments_hippo_natural_images} and Section~\ref{sec:exp_hippo_mnist}).
This is of importance as the neural activities will not be random, but correlated in some way that is characteristic for the individual's environment. 

Our work illustrates that it is possible to implement a model of the hippocampus for online sequence storage and that its success highly depends on the use of an intrinsic sequence in CA3, pattern separation in DG (or a `dreaming process instead'), and the use of Hebbian-descent or a learning rule with a similar one-shot learning performance (but not Hebb's rule) for achieving a plausible forgetting behavior. 
However, we are not aware of any biologically plausible learning rule that has a similar online learning performance similar to that of Hebbian-descent.
The implementation of our models is based on the Python library PyDeep~\citep{melchior2018pydeep} publicly available at \url{https://github.com/MelJan/PyDeep} and source code for model training, usage, and evaluation is made available under the same address.

\bibliography{dissertation_references}
\bibliographystyle{apalike}

\begin{appendices}

\section{Detailed Calculation Flow of \emph{Model-A} and \emph{Model-B}.}\label{appendix:calculation_flow_models}

For notational simplicity we use matrix-notation, so that the activity of all neurons within one subregion are calculated according to Equation~\ref{eqn:neuron_element} by
\begin{eqnarray}
\vect h = \vect \phi\left(\vect{W}^T\left(\vect{x}-\vect\mu \right)+\vect{b}\right)\label{eqn:neuron_matrix}.
\end{eqnarray} 
The calculation flow for \emph{Model-A} for retrieving the next pattern is then given by
\begin{eqnarray}
\vect{x}^{_{EC}}(t) &=& s\left(\vect{W}^{_{SI\rightarrow EC\hspace*{3.6mm}}}\left(\vect{x}^{_{SI}}(t)\hspace*{8.2mm}-\vect\mu^{_{SI}}\hspace*{2.4mm}\right)+\vect{b}^{_{SI\rightarrow EC}}\hspace*{3.25mm}\right),\\
\vect{x}^{_{CA3}}(t) &=& \sigma\left(\vect{W}^{_{EC\rightarrow CA3\hspace*{1.2mm}}}\left(\vect{x}^{_{EC}}(t)\hspace*{7.2mm}-\vect\mu^{_{EC}}\hspace*{1.5mm}\right)+\vect{b}^{_{EC\rightarrow CA3}}\hspace*{1.1mm}\right),\\
\vect{x}^{_{CA3}}(t+1) &=& \sigma\left(\vect{W}^{_{CA3\rightarrow CA3}}\left(\vect{x}^{_{CA3}}(t)\hspace*{6.1mm}-\vect\mu^{_{CA3}}\hspace*{0.3mm}\right)+\vect{b}^{_{CA3\rightarrow CA3}}\right),\\
\vect{x}^{_{EC}}(t+1) &=& \sigma\left(\vect{W}^{_{CA3\rightarrow EC\hspace*{1.2mm}}}\left(\vect{x}^{_{CA3}}(t+1) -\vect\mu^{_{CA3}}\hspace*{0.3mm}\right)+\vect{b}^{_{CA3\rightarrow EC}}\hspace*{1.2mm}\right),\\
\vect{x}^{_{SI}}(t+1) &=& \sigma\left(\vect{W}^{_{EC\rightarrow SI\hspace*{3.4mm}}}\left(\vect{x}^{_{EC}}(t+1)\hspace*{1.2mm}-\vect\mu^{_{EC}}\hspace*{1.3mm}\right)+\vect{b}^{_{EC\rightarrow SI}}\hspace*{3.2mm}\right),
\end{eqnarray} 
where $s(\cdot)$ denotes the Step function used to transfer the input $\vect{x}^{_{SI}}(t)$ to a binary representation $\vect{x}^{_{EC}}(t)$. 
For \emph{Model-B} the calculations are given by
\begin{eqnarray}
\vect{x}^{_{EC}}(t) &=& s\left(\vect{W}^{_{SI\rightarrow EC}\hspace*{3.6mm}}\left(\vect{x}^{_{SI}}(t)\hspace*{8.2mm}-\vect\mu^{_{SI}}\hspace*{2.4mm}\right)+\vect{b}^{_{SI\rightarrow EC}}\hspace*{3.3mm}\right),\\
\vect{x}^{_{DG}}(t) &=& \sigma\left(\vect{W}^{_{EC\rightarrow DG}\hspace*{2.2mm}}\left(\vect{x}^{_{EC}}(t)\hspace*{7.2mm}-\vect\mu^{_{EC}}\hspace*{1.5mm}\right)+\vect{b}^{_{EC\rightarrow DG}}\hspace*{2.2mm}\right),\\
\vect{x}^{_{CA3}}(t) &=& \sigma\left(\vect{W}^{_{DG\rightarrow CA3}\hspace*{1.05mm}}\left(\vect{x}^{_{DG}}(t)\hspace*{7.0mm}-\vect\mu^{_{DG}}\hspace*{1.3mm}\right)+\vect{b}^{_{DG\rightarrow CA3}}\hspace*{1.1mm}\right),\\
\vect{x}^{_{CA3}}(t+1) &=& \sigma\left(\vect{W}^{_{CA3\rightarrow CA3}}\left(\vect{x}^{_{CA3}}(t)\hspace*{6.1mm}-\vect\mu^{_{CA3}}\hspace*{0.3mm}\right)+\vect{b}^{_{CA3\rightarrow CA3}}\right),\\
\vect{x}^{_{EC}}(t+1) &=& \sigma\left(\vect{W}^{_{CA3\rightarrow EC}\hspace*{1.2mm}}\left(\vect{x}^{_{CA3}}(t+1) -\vect\mu^{_{CA3}}\hspace*{0.3mm}\right)+\vect{b}^{_{CA3\rightarrow EC}}\hspace*{1.2mm}\right),\\
\vect{x}^{_{SI}}(t+1) &=& \sigma\left(\vect{W}^{_{EC\rightarrow SI}\hspace*{3.4mm}}\left(\vect{x}^{_{EC}}(t+1)\hspace*{1.2mm}-\vect\mu^{_{EC}}\hspace*{1.3mm}\right)+\vect{b}^{_{EC\rightarrow SI}}\hspace*{3.2mm}\right),
\end{eqnarray} 
Notice, that the input can also be a retrieved pattern or a corrupted input denoted by $\dot{\vect{x}}$ and $\tilde{\vect{x}}$, respectively.

\end{appendices}

\end{document}